%% file: template.tex
\documentclass{article}

\usepackage{PRIMEarxiv}

\usepackage[utf8]{inputenc} 
\usepackage[T1]{fontenc}    
\usepackage{hyperref}       
\usepackage{url}            
\usepackage{booktabs}       
\usepackage{amsfonts}       
\usepackage{nicefrac}       
\usepackage{microtype}      
\usepackage{lipsum}		
\usepackage{graphicx}
\usepackage{doi}
\usepackage{soul}
\usepackage{xcolor}
\usepackage{amsmath}
\usepackage{amssymb}
\usepackage{mparhack}
\usepackage{marginnote}
\usepackage{comment}
\usepackage{pifont}
\usepackage{fontawesome}
\usepackage{longtable}
\usepackage{amsthm}
\usepackage{bm}
\usepackage{array}
\usepackage{url}
\usepackage{wrapfig}
\usepackage{amsfonts}       
\usepackage{nicefrac} 
\usepackage{cancel}
\usepackage{mathrsfs}

\usepackage[acronym]{glossaries}
\input{acronyms}

\newcolumntype{L}{>{\centering\arraybackslash}m{3cm}}
\newcolumntype{M}{>{\centering\arraybackslash}m{2cm}}
\theoremstyle{definition}
\newtheorem{defi}{Definition}

\title{Unsupervised Representation Learning in Deep Reinforcement Learning: A Review}

\author{Nicolò Botteghi \\
	Department of Applied Mathematics\\
	Mathematics of Imaging and AI group \\
	University of Twente\\
	Enschede, Netherlands \\
	\texttt{n.botteghi@utwente.nl} \\
	\And
	Mannes Poel \\
	Department of Computer Science\\
	Datamanagement and Biometrics group\\
	University of Twente \\
	Enschede, Netherlands \\
	\texttt{m.poel@utwente.nl} \\
	\And
	Christoph Brune \\
	Department of Applied Mathematics\\
	Mathematics of Imaging and AI group \\
	University of Twente\\
	Enschede, Netherlands \\
	\texttt{c.brune@utwente.nl} \\
}

\hypersetup{
pdftitle={A template for the arxiv style},
pdfsubject={q-bio.NC, q-bio.QM},
pdfauthor={Nicolò Botteghi}
pdfkeywords={Reinforcement Learning - Unsupervised Representation Learning - Dynamical Systems},
}

\begin{document}
\maketitle

\begin{abstract}

This review addresses the problem of learning abstract representations of the measurement data in the context of Deep Reinforcement Learning (\acrshort{drl}). While the data are often ambiguous, high-dimensional, and complex to interpret, many dynamical systems can be effectively described by a low-dimensional set of state variables. Discovering these state variables from the data is a crucial aspect for (i) improving the data efficiency, robustness, and generalization of \acrshort{drl} methods, (ii) tackling the \textit{curse of dimensionality}, and (iii) bringing interpretability and insights into black-box \acrshort{drl}. This review provides a comprehensive and complete overview of unsupervised representation learning in \acrshort{drl} by describing the main Deep Learning tools used for learning representations of the world, providing a systematic view of the method and principles, summarizing applications, benchmarks and evaluation strategies, and discussing open challenges and future directions.
	
\end{abstract}
\keywords{Reinforcement Learning \and Unsupervised Representation Learning \and Dynamical Systems}

\onecolumn
\tableofcontents

\newpage

\section{Introduction}
\input{StateRepresentationLearning/1-Introduction}

\section{Background}\label{sec:background}
\input{StateRepresentationLearning/2-Background}

\section{Deep Learning for Learning State Representations}\label{sec:OverviewMethods}
\input{StateRepresentationLearning/3-DeepLearningforLearningStateRepresentations}

\section{Unsupervised Representation Learning in Deep Reinforcement Learning}\label{sec:unsupervisedlearningDRL}
\input{StateRepresentationLearning/4-UnsupervisedRepresentationLearning}

\section{Advanced Methods}\label{sec:advancedmethods}
\input{StateRepresentationLearning/5-AdvancedMethods}

\section{Applications}\label{sec:application}
\input{StateRepresentationLearning/6-Applications}

\section{Evaluation and Benchmark}\label{sec:evaluationandbenchmark}
\input{StateRepresentationLearning/7-EvaluationandBenchmark}

\section{Discussion}\label{sec:discussion}
\input{StateRepresentationLearning/8-Discussion}

\section{Conclusion}\label{sec:conclusion}
\input{StateRepresentationLearning/9-Conclusion}

\bibliographystyle{ieeetr}
\bibliography{references} 

\clearpage

\section*{List of Acronyms}\label{glossary}
\printglossary[type=\acronymtype, style=tree, title=]


\clearpage
\newpage
\appendix
\section{Implementation Details}\label{sec:appendixA}
In this section, we provide the implementation details for reproducing the results presented in Section \nameref{sec:OverviewMethods}. The code used for generating the results and the dataset used is publicly available at \url{https://github.com/nicob15/State_Representation_Learning_Methods}. The code is developed using the PyTorch \cite{paszke2019pytorch}.
\input{StateRepresentationLearning/10-Appendix}

\end{document}

%% file: acronyms.tex
\makeglossaries 

\newacronym{ml}{ML}{Machine Learning}

\newacronym{rl}{RL}{Reinforcement Learning}

\newacronym{mdp}{MDP}{Markov Decision Process}

\newacronym{dl}{DL}{Deep Learning}

\newacronym{drl}{DRL}{Deep Reinforcement Learning}

\newacronym{srl}{SRL}{State Representation Learning}

\newacronym{nips}{NeurIPS}{International Conference on Neural Information Processing Systems}

\newacronym{iclr}{ICLR}{International Conference on Learning Representations}

\newacronym{icml}{ICML}{International Conference on Machine Learning}

\newacronym{icra}{ICRA}{International Conference on Robotics and Automation}

\newacronym{iros}{IROS}{Internation Conference on Intelligent Robots and Systems}

\newacronym{pomdp}{POMDP}{Partially Observable Markov Decision Process}

\newacronym{bmdp}{BMDP}{Block Markov Decision Process}

\newacronym{oc}{OC}{Optimal Control}

\newacronym{dp}{DP}{Dynamic Programming}

\newacronym{fcnn}{FCNN}{Fully-Connected Neural Network}

\newacronym{cnn}{CNN}{Convolutional Neural Network}

\newacronym{rnn}{RNN}{Recurrent Neural Network}

\newacronym{lidar}{LIDAR}{Light Detection and Ranging}

\newacronym{pca}{PCA}{Principal Components Analysis}

\newacronym{ae}{AE}{Autoencoder}

\newacronym{vae}{VAE}{Variational Autoencoder}

\newacronym{kl}{KL}{Kullbach-Leibler}

\newacronym{rgb}{RGB}{Red Green Blue}

\newacronym{mse}{MSE}{Mean Squared Error}

\newacronym{cl}{CL}{Contrastive Learning}

\newacronym{cpc}{CPC}{Contrasitve Predicting Coding}

\newacronym{sac}{SAC}{Soft Actor Critic}

\newacronym{ppo}{PPO}{Proximal Policy Optimization}

\newacronym{dqn}{DQN}{Deep Q-Network}

\newacronym{a3c}{A3C}{Asynchronous Advantage Actor Critic}

\newacronym{mpc}{MPC}{Model Predictive Control}

\newacronym{e2c}{E2C}{Embed to Control}

\newacronym{ssm}{SSM}{State Space Model}

\newacronym{knn}{KNN}{K-Nearest Neighbours}

\newacronym{exai}{ExAI}{Explainable Artificial Intelligence}

\newacronym{tsne}{t-SNE}{t-distributed Stochastic Neighbour Embedding}

\newacronym{gtc}{GTC}{Ground Truth Correlation}

\newacronym{nieqa}{NIEQA}{Normalization Independent Embedding Quality Assessment}

\newacronym{lre}{LRE}{Linear Reconstruction Error}

\newacronym{vd}{VD}{Validation Decoder}

\newacronym{mor}{MOR}{Model Order Reduction}

%% file: StateRepresentationLearning/1-Introduction.tex
Dynamical systems pervade our daily life. A pendulum swinging, a robot picking up objects, the flow of water in a river, the trend of economic costs, the spread of a disease and the prediction of climate change are all examples of dynamical systems. Understanding how these systems evolve by discovering the governing equations, how they react to inputs, and how they can be controlled are crucial aspects of science and engineering \cite{brunton2016discovering}.

Governing equations are traditionally derived from physical principles, such as conservation laws and symmetries. However, many complex real-world dynamical systems exhibit strongly nonlinear and complex behaviors, making the discovery of the governing principles, describing the evolution of the systems, extremely challenging. Additionally, in many domains, such as physical systems, robotics, economics, or health, we are interested not only in discovering the governing principles but also in controlling the dynamical systems via additional sets of input variables, called interchangeably control variables or actions, such that the systems manifest the desired behaviors.  The problem of finding optimal control laws minimizing task-dependent cost functionals has been extensively studied in Optimal Control  (\acrshort{oc}) \cite{bryson1975applied,  stengel1994optimal, kirk2004optimal, lewis2012optimal, athans2013optimal} under the assumption of complete knowledge of the system dynamics.

In the last decade, Artificial Intelligence \cite{winston1984artificial}, Machine Learning  (\acrshort{ml}) \cite{mitchell1997machine}, and data-driven methods have paved the road for new approaches to studying, analyzing, understanding and controlling dynamical systems. While data, in the forms of measurements of the systems, are often abundant, physical laws are not always available. Even when present, physical laws often cannot fully capture the actual governing principles due to unknowns and uncertainties \cite{brunton2022data}. Traditional control techniques based on accurate models of the world are often unsuitable for such dynamical systems, making data-driven control methods, such as Reinforcement Learning (\acrshort{rl}) \cite{Sutton1998}, even more appealing. 

\acrshort{rl} is the branch of \acrshort{ml} studying a computational approach for learning optimal control laws from the interaction with the world. \acrshort{rl} owes its success in the Markov Decision Processes (\acrshort{mdp}s) theory \cite{puterman1990markov}, and on the Nature-inspired paradigm of an intelligent \textit{agent}, learning to act by interacting with an unknown \textit{environment} 
and improving its behavior based on the consequences of the actions taken. Consequences of the actions are quantified by a scalar signal called \textit{reward}. Similarly to \acrshort{oc},  \acrshort{rl} 
faces the so-called \textit{curse of dimensionality} \cite{Sutton1998} deriving from the continuous nature of state and action spaces and from the high-dimensionality of the observable variables. These two features are common to many dynamical systems, making \acrshort{rl} algorithms computationally inefficient.

Inspired by the outstanding successes of Deep Learning (\acrshort{dl}) \cite{lecun2015deep}, \acrshort{rl} has turned towards deep function approximators, e.g. neural networks, for representing the control strategies and tackling the curse of dimensionality. The use of deep neural networks in \acrshort{rl} gave birth to Deep Reinforcement Learning (\acrshort{drl}). \acrshort{drl} has achieved outstanding successes in learning optimal control policies directly from high-dimensional observations \cite{mnih2013playing, mnih2015human, arulkumaran2017deep}. However, neural networks' drawbacks have also been inherited, namely the low sample efficiency, the consequent need for huge amounts of data, and the instabilities of learning.

The sample inefficiency of the methods is one of the significant drawbacks of \acrshort{drl}. This aspect becomes even more severe when controlling dynamical systems requiring real-world interaction -- we cannot afford robots to collide with obstacles multiple times before learning how to navigate around them properly. The sample inefficiency of \acrshort{drl} derives from disregarding any prior knowledge of the world and from the sole use of the reward signal to optimize the neural network models. The reward signal assesses only the quality of the decision but does not directly indicate the best course of action in the long term. It is worth highlighting that \acrshort{drl} does not rely on any labelled data as in the case of standard supervised learning. Therefore, \acrshort{drl}, to improve sample efficiency, has recently shifted its focus toward unsupervised learning of data representations \cite{Bengio2013}.

\begin{figure*}[h!]
    \centering
        \includegraphics[width=1.0\textwidth]{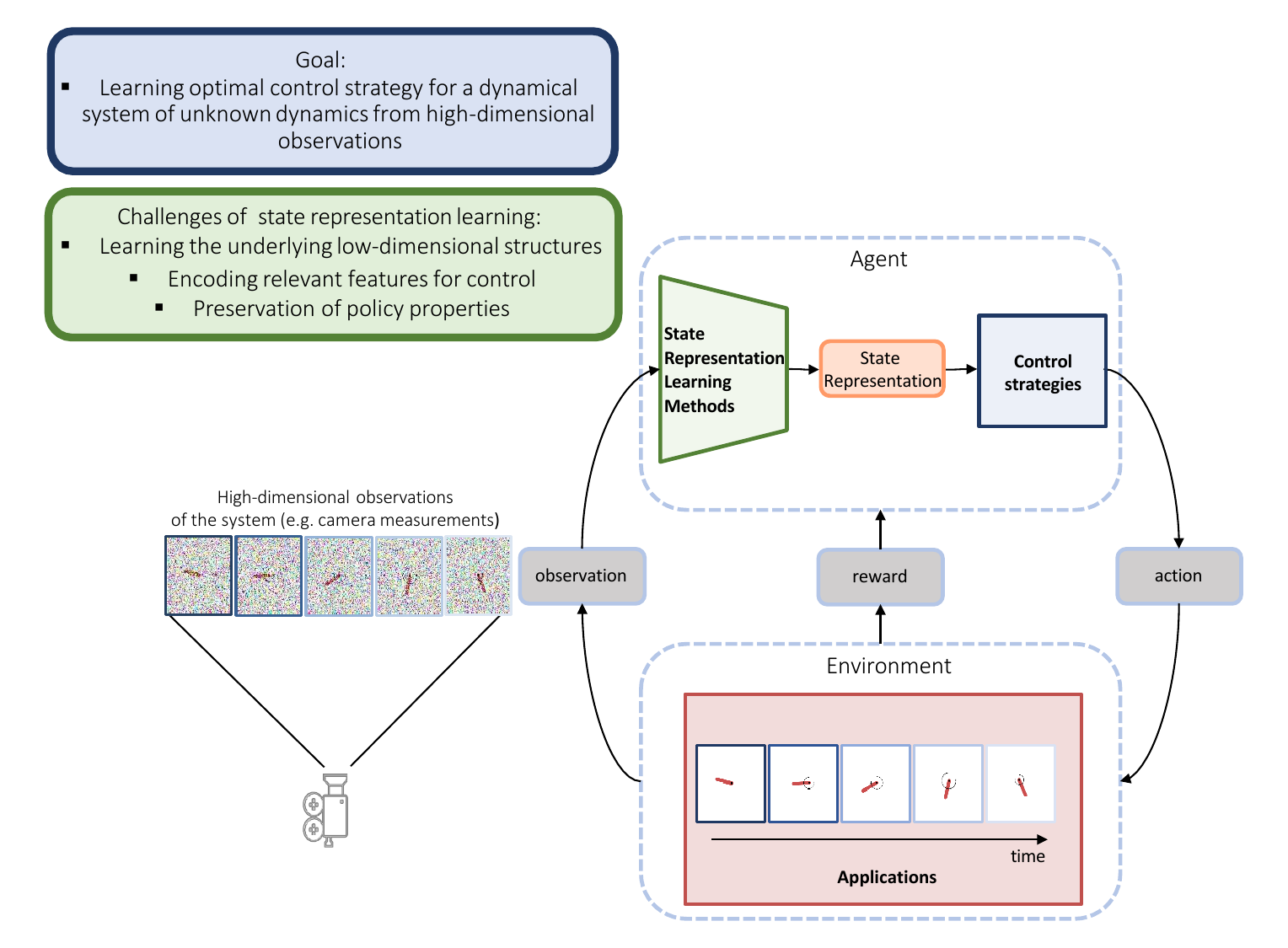}
        \caption{State Representation Learning for Deep Reinforcement Learning.}
        \label{fig:DRLSRL}
\end{figure*}

In real-world control problems, we often do not have access to the state variables unequivocally describing the evolution of the systems. However, we only have access to indirect, high-dimensional, and noisy measurements -- observations in \acrshort{drl} terminology -- e.g. noisy RGB images or, in general, any given sensory data. While data are often high-dimensional, many dynamical systems exhibit low-dimensional behaviors that can be effectively described by a limited number of latent variables that capture the systems' essential properties for control and identification purposes. Given the observation stream, we want to (i) learn low-dimensional state representations that preserve the relevant properties of the world and then (ii)  use such representations to learn the optimal policies. When \acrshort{drl} methods rely on compact representations instead of high-dimensional observations, the algorithms gain higher sample efficiency, generalization and robustness. Figure \ref{fig:DRLSRL} depicts how to exploit unsupervised representation learning in \acrshort{drl} in the context of control of dynamical systems, e.g. a simple pendulum. Again, we do not usually have access to the labelled data, i.e. the actual state values. Learning meaningful data representations for control without supervision is a significant and open challenge of \acrshort{drl} research, and it is often studied under the name of State Representation Learning (\acrshort{srl}) \cite{Bohmer2015, Lesort2018, Achille2017}. 

\begin{figure}
    \centering
    \includegraphics[width=1.0\textwidth]{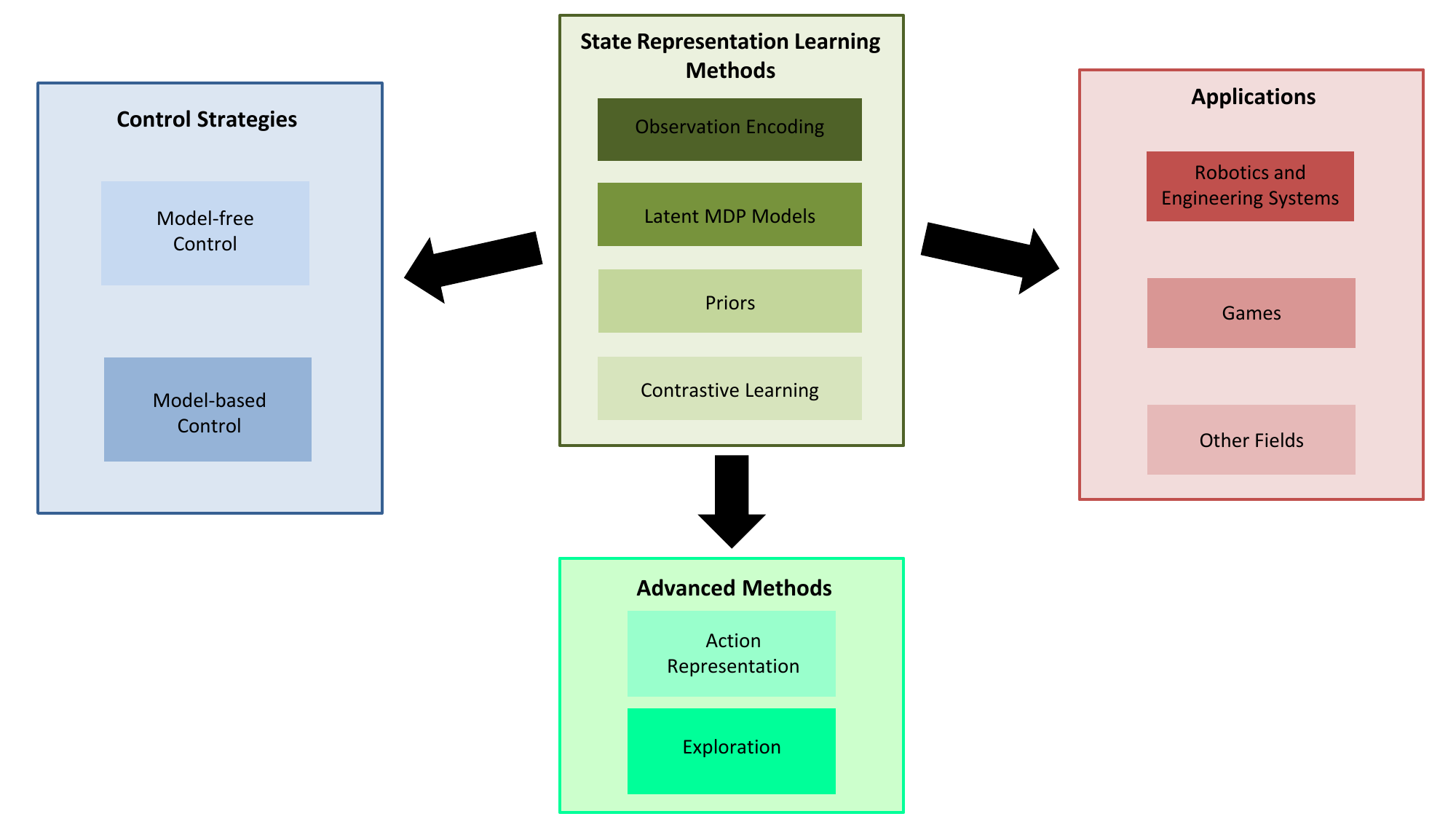}
    \caption{Structure of the review. }
    \label{fig:mainblocks}
\end{figure}

Our paper aims to provide a comprehensive and complete review of the most relevant, influential, and newest trends of unsupervised \acrshort{srl} in \acrshort{drl}. Other surveys from the literature, e.g. \cite{Bohmer2015, Lesort2018, Achille2017}, although they cover a considerable amount of work, they are outdated considering the rapid evolution of the research in this field and do not cover the latest discoveries. Our review:
\begin{itemize}
    \item describes the main \acrshort{dl} tools used in \acrshort{drl} for learning meaningful representations and their desired properties,
    \item classifies the important trends by analyzing the most relevant contributions in the field,
    \item summarizes the applications, benchmarks, and evaluation strategies crucial for the advancement of the research, and
    \item provides an elaborate discussion on open challenges and future research directions.
\end{itemize}
Figure \ref{fig:mainblocks} provides a visual overview of these key elements of the review.

We review over 100 publications on a wide variety of \acrshort{srl} approaches for \acrshort{drl}. We include papers from relevant journals and conferences in the field (e.g. \acrshort{nips}, \acrshort{iclr}, \acrshort{icml}, \acrshort{icra}, and \acrshort{iros}) and ArXiv, and we checked each contribution's references and related work. We identified the relevant contributions using keywords such as: "State Representation Learning", "Deep Reinforcement Learning", and "Learning Abstract Representations". The latest work included in the review is dated January 2022.

The review is organized as follows: Section \ref{sec:background} provides the background information related to \acrshort{mdp}s and the notation. Section \ref{sec:OverviewMethods} provides a gentle introduction to the Deep Learning methods constituting the building blocks for learning meaningful representations. Section \ref{sec:unsupervisedlearningDRL} describes the \acrshort{srl} methods from the literature,  Section \ref{sec:advancedmethods} describes advanced combinations of the previously-introduced methods, and Section \ref{sec:application} the most prominent applications of \acrshort{srl}.  Section \ref{sec:evaluationandbenchmark} discusses the evaluation metrics and the benchmarks. Section \ref{sec:discussion} uses our experience in the field to discuss the open challenges and the most exciting research directions, and Section \ref{sec:conclusion} concludes the review.

%% file: StateRepresentationLearning/2-Background.tex
In Section \ref{subsec:rl-oc}, we provide a brief introduction on the connection between  \acrshort{oc} and \acrshort{rl}, in Section \ref{subsec:MDP}, we introduce the Markov \acrshort{mdp}, in Section \ref{subsec:POMDP} the Partially Observable Markov Decision Processes (\acrshort{pomdp}), and in Section \ref{subsec:BMDP} the Block Markov Decision Process (\acrshort{bmdp}) framework. Eventually, in Section \ref{subsec:Notation}, we introduce the notation used in \acrshort{srl} and this review.

\subsection{Optimal Control and Reinforcement Learning}\label{subsec:rl-oc}
\subsubsection{Terminology}
The fields of \acrshort{oc} and \acrshort{rl}  use similar concepts and techniques to solve decision-making and control problems. While the terminology and notation used in these two domains may vary, there exists a strong correspondence between them \cite{lewis2009reinforcement, bertsekas2019reinforcement, sutton1992reinforcement, kiumarsi2017optimal, lewis2012reinforcement, lewis2013reinforcement}. Table \ref{tab:RL-OC-notation} aims to summarize this relation, providing a clear understanding of how the main concepts align across these disciplines. Additionally, in \nameref{glossary} we introduce the acronyms used throughout the review.
\begin{table}[h!]
\caption{Correspondence between \acrshort{rl} and \acrshort{oc} notation.}
\centering
\small
\begin{tabular}{|p{0.23\linewidth} | p{0.23\linewidth} |  p{0.33\linewidth}|} 
\hline
 \hline
 Optimal Control & Reinforcement Learning & Explanation\\ [0.5ex] 
 \hline\hline
 state variables $(\mathbf{x})$ & state variables $(\mathbf{s})$ & Represent the system's current \hbox{configuration}.  \\
 state estimates $(\hat{\mathbf{x}})$ &latent state variables $(\mathbf{z})$ & Estimate of the state variable when not directly observable. \\
 outputs $(\mathbf{y})$ & observations $(\mathbf{o})$ & Represent what can be measured of the system.  \\
output estimates $(\hat{\mathbf{y}})$ & observation \hbox{reconstructions $(\hat{\mathbf{o}})$} &  Estimate of the system's observable quantities. \\
control inputs $(\mathbf{u})$ & actions $(\mathbf{a})$ & The decision or control applied to the system.  \\
 control law $(\mu)$ & policy $(\pi)$ &Specifies the strategy for selecting \hbox{actions} based on states. \\
 cost $(c)$ & reward $(r)$ & Instantaneous scalar value defining the desirability of the current state and action.  \\
 cost function $(J)$ & reward function $(R)$ & User-defined function defining the goal of the agent, i.e. desirability of each state-action pair.  \\
vector field $(f)$ / \hbox{state transition map $(F)$} & transition function $(T)$  &  Describes how the system evolves over time. The time variable can be either discrete (e.g. in the case of $T$ and $F$) or continuous (e.g. in the case of $f$). \\
controller & agent &  The decision-making entity. \\
system & environment &  The environment is considered to be everything outside the agent. The system, in this context, is a part of the environment. \\ 
 \hline
 \hline
\end{tabular}
 \label{tab:RL-OC-notation}
\end{table}

\subsubsection{Problem Settings}

In this review, we specifically focus on the control of discrete-time and nonlinear dynamical systems:
\begin{equation}
    \bm{s}_{t+1} = T(\bm{s}_t, \bm{a}_t)
    \label{eq:dyn_system}
\end{equation}
 where $\bm{s}_{t}$ and $\bm{s}_{t+1}$ indicate the state variables at timestep $t$ and $t+1$ respectively, $\bm{a}_t$ is the action at timestep $t$, and $T$ is the transition function determining the evolution over time of the system given the current state and action.
Given the system described in Equation \eqref{eq:dyn_system}, the agent aims to find the optimal policy $\pi$ that maximizes the reward collected in a particular time horizon $H$ (either finite or infinite).
These problem settings strongly resemble a discrete-time version of the \acrshort{oc} problem, often referred to as Dynamic Programming (\acrshort{dp}) \cite{Sutton1998, bertsekas2012dynamic}, where a sufficient condition for optimality is provided by solving the Hamilton-Jacobi-Bellman equation \cite{mceneaney2006max, fleming2000max}. However, differently from \acrshort{dp}, in \acrshort{rl}, the agent does not know the transition function $T$ and the reward function $R$ and has to learn the optimal behavior by interacting with the environment, i.e. from the observations and rewards samples \cite{Sutton1998, lewis2009reinforcement}. In Section \ref{subsec:MDP}, we will introduce the mathematical framework to study the \acrshort{rl} (and \acrshort{dp}) problem, i.e. the \acrshort{mdp}s.

We consider control scenarios where the state of the environment $\bm{s}$ is not directly observable by the agent. The agent can only utilize high-dimensional, indirect, and noisy observations of the state variables. In this setting, we review the methods learning from observations, i.e., raw data, low-dimensional and informative representations of the state variables without observing them directly. We make use of Unsupervised Learning \cite{goodfellow2016deep} to learn state representations without supervision, i.e., without access to the actual state variables. This problem closely relates to State Estimation using observers \cite{chernousko1993state, thrun2002probabilistic, barfoot2017state, kwakernaak1972linear}, well-studied topic in control systems and robotics. However, we assume no knowledge of the function that maps states to observations, yet we are trying to infer it by utilizing \acrshort{ml} methods.

Learning meaningful state representations for control relies on extracting features from observation that the agent can effectively control with its action. For this reason, many of the methods from the literature try to estimate the environment's unknown (latent) dynamics from samples. This process strongly relates to System Identification \cite{eykhoff1974system, ljung1998system}, where a mathematical model is derived from observation data. 

In some applications, we may know the systems' dynamics. However, due to its cost, the computation model is prohibitively computationally expensive and cannot be used for real-time or control applications. To overcome this challenge, one can learn from data reduced-order and computationally-efficient models by relying on dimensionality reduction techniques \cite{lassila2014model, schilders2008model, lucia2004reduced, noack2011reduced, fresca2021comprehensive}. \acrshort{srl} principles closely relate to Model-order Reduction (\acrshort{mor}) ones, with the main difference that \acrshort{srl} focuses on learning low-dimensional state representations and latent models for the sake of optimally controlling the systems. In contrast, \acrshort{mor} methods focus on prediction accuracy and reliability of the reduced-order models.

\subsection{Markov Decision Processes}\label{subsec:MDP}

The agent-environment interaction scheme of \acrshort{rl} can be studied by means of the \acrshort{mdp}. 
\begin{defi}
    \acrshort{mdp} (adapted from \cite{Sutton1998})\textbf{:} A \acrshort{mdp} is a tuple $\langle \mathcal{S}, \mathcal{A}, T, R \rangle$ where $\mathcal{S} \subset \mathbb{R}^n$ is the set of states, $\mathcal{A}\subset \mathbb{R}^m$ is the set of actions, $T: \mathcal{S} \times \mathcal{S} \times \mathcal{A} \longrightarrow [0,1]; \ (\bm{s}',\bm{s},\bm{a}) \longmapsto T(\bm{s}',\bm{s},\bm{a})=p(\bm{s}'|\bm{s},\bm{a})$ is the transition function, describing the probability $p(\bm{s}'|\bm{s},\bm{a})$ of reaching state $\bm{s}'$ from state $\bm{s}$ while taking action $\bm{a}$, and $R: \mathcal{S} \times \mathcal{A} \longrightarrow \mathbb{R}; \ (\bm{s},\bm{a})\longmapsto R(\bm{s},\bm{a})$ is the reward function.
 \end{defi}
The goal of the agent is to find the optimal stochastic $\pi^*(\bm{a}|\bm{s})$,  or deterministic $\bm{a}=\pi^*(\bm{s})$, policy, i.e. the policy maximizing the expected cumulative reward in the long run. 

\subsubsection{Partially Observable Markov Decision Processes}\label{subsec:POMDP}

In many real-world problems, the state of the environment is not directly accessible by the agent, and the world is only observable via partial and often high-dimensional observations. Examples of partial observable environments are: the Atari games \cite{mnih2013playing} when played using pixel information, i.e., raw images, or a robot learning to navigate by relying only on sensory readings \cite{morik2019state}. The \acrshort{mdp} framework is insufficient for properly modeling these problems as a single observation may not contain sufficient information to distinguish between two or more states unequivocally. Therefore, one must rely on the framework of \acrshort{pomdp}.
\begin{defi}
    \acrshort{pomdp} (adapted from \cite{Sutton1998})\textbf{:} A \acrshort{pomdp} is a tuple $\langle \mathcal{S}, \mathcal{A}, \mathcal{O}, T, \Omega, \mathcal{R} \rangle$ where $\mathcal{S} \subset \mathbb{R}^n$ is the set of unobservable states, $\mathcal{A}\subset \mathbb{R}^m$ is the set of actions, $\mathcal{O}\subset \mathbb{R}^o$ is the set of observations, $T: \mathcal{S} \times \mathcal{S} \times \mathcal{A} \longrightarrow [0,1]; \ (\bm{s}',\bm{s},\bm{a}) \longmapsto T(\bm{s}',\bm{s},\bm{a})=p(\bm{s}'|\bm{s},\bm{a})$ is the transition function, $\Omega: \mathcal{O} \times \mathcal{S}\times \mathcal{A} \longrightarrow [0,1]; \ (\bm{o},\bm{s},\bm{a}) \longmapsto \Omega(\bm{o},\bm{s},\bm{a})=p(\bm{o}|\bm{s},\bm{a})$ is the observation function and $R: \mathcal{S} \times \mathcal{A} \longrightarrow \mathbb{R}; \ (\bm{s},\bm{a})\longmapsto R(\bm{s},\bm{a})$ is the reward function. 
\end{defi}
To learn the optimal policy, the agent must first infer its current state $\bm{s}_t$. Due to the partial observability of the environment, a single observation is not enough for discriminating between states and the state estimation problem \hbox{ $\bm{s}_t \sim p(\bm{s}_t|\bm{o}_t, \bm{o}_{t-1}, \dots, \bm{o}_{t-n}, \dots, \bm{o}_0,  \bm{a}_{t-1}, \dots, \bm{a}_{t-n}, \dots, \bm{a}_0)$ }requires the history of observations and actions. We often only need $n$ history steps to discriminate between two states unequivocally \cite{liu2022partially}. In the most general settings, the problem of learning state representations should be modeled using the \acrshort{pomdp} framework. However, we can often rely on a simplified version of such framework: the \acrshort{bmdp} \cite{sanders2020clustering}. We refer to \cite{liu2022partially} for a formal description of the difference between \acrshort{pomdp} and \acrshort{bmdp}.

\subsubsection{Block MDP}\label{subsec:BMDP}

In a \acrshort{bmdp} \cite{du2019provably}, each observation has only one corresponding environment state and can be considered \textit{Markovian}, i.e., a single observation contains sufficient information for determining the state. 
\begin{defi}
    \acrshort{bmdp} (adapted from \cite{du2019provably})\textbf{:} A \acrshort{bmdp} is a tuple $\langle \mathcal{S}, \mathcal{A}, \mathcal{O}, T, \Omega, \mathcal{R} \rangle$ where $\mathcal{S}$ is the set of unobservable states, $\mathcal{A}$ is the set of actions, $\mathcal{O}$ is the observation space assumed to be Markovian, $T: \mathcal{S} \times \mathcal{S} \times \mathcal{A} \longrightarrow [0,1]; \ (\bm{s}',\bm{s},\bm{a}) \longmapsto T(\bm{s}',\bm{s},\bm{a})=p(\bm{s}'|\bm{s},\bm{a})$ is the transition function, $R: \mathcal{S} \times \mathcal{A} \longrightarrow \mathbb{R}; \ (\bm{s},\bm{a})\longmapsto R(\bm{s},\bm{a})$ is the reward function, and $\Omega: \mathcal{O} \times \mathcal{S}\times \mathcal{A} \longrightarrow [0,1]; \ (\bm{o},\bm{s},\bm{a}) \longmapsto \Omega(\bm{o},\bm{s},\bm{a})=p(\bm{o}|\bm{s},\bm{a})$ is the observation function.
\end{defi}
The advantage of modelling problems via \acrshort{bmdp}s rather than \acrshort{pomdp}s is that we do not need any memory structure for learning the state representation, as single observations already contain all the information for inferring the states, i.e., $p(\bm{s}_t|\bm{o}_t)$. However, even when the observations are Markovian, encoding the relevant information from the high-dimensional and potentially noisy measurements into the state representation still remains a challenge.

\subsection{State Representation via Encoding History}\label{subsec:Notation}
In the most general case, a state representation $\bm{z}$ is a function of all the history of observations and actions collected by the agent while interacting with the environment:
\begin{equation}
    \bm{z}_t = \phi_{\text{enc}}(\bm{o}_t, \bm{o}_{t-1}, \dots, \bm{o}_0, \bm{a}_{t-1}, \dots, \bm{a}_0)\,, 
    \label{eq:generic_encoder}
\end{equation}
where we refer to $\phi_{\text{enc}}$ as the \textit{encoder}, $\bm{z}_t$ is the latent state at time step $t$, $\bm{o}_t, \dots, \bm{o}_0$ indicate the observations from time step $t$ to $0$ and $\bm{a}_{t-1}, \dots, \bm{a}_0$ indicate the actions from time step $t-1$ to $0$. In practice, we rarely need the complete history, and only a subset of it is sufficient for inferring the state:
\begin{equation}
    \bm{z}_t = \phi_{\text{enc}}(\bm{o}_t, \dots, \bm{o}_{t-n}, \bm{a}_{t-1}, \dots, \bm{a}_{t-n-1})\,, 
\end{equation}
with $n \ll t$.

In the case of \acrshort{bmdp}s, the latent states depend only on the information at the current time step, i.e. no history is considered:
\begin{equation}
    \bm{z}_t = \phi_{\text{enc}}(\bm{o}_t).
\end{equation}

For introducing and describing the \acrshort{srl} methods, in Section \ref{sec:unsupervisedlearningDRL}, we will use the \acrshort{bmdp}s notation for the sake of simplicity. 

\textit{Remark:} The \acrshort{bmdp} notation can be easily extended to sequences of observations and actions:
\begin{equation}
    \bm{z}_t = \phi_{\text{enc}}(\bm{h}_t)\,,
    \label{eq:phi_enc_general}
\end{equation}
with $\bm{h}_t = [\bm{o}_t, \dots, \bm{o}_{t-n}, \bm{a}_{t-1}, \dots, \bm{a}_{t-n-1}]$ denoting the history of observations and actions.

In this review, we specifically focus on the methods relying on \acrshort{dl} and neural networks for approximating the unknown function $\phi_{\text{enc}}$.

%% file: StateRepresentationLearning/3-DeepLearningforLearningStateRepresentations.tex
This section is composed of three parts: (i) we introduce the basic neural network architectures in Section \ref{subsec:neural_networks_architectures} that can be used to approximate the mapping $\phi_{\text{enc}}$  (see Equation \eqref{eq:phi_enc_general}), (ii) we show different unsupervised learning methods for learning the optimal mapping $\phi_{\text{enc}}$ in Section \ref{subsec:SRL_methods}, and (iii) we summarize and discuss the methods introduced in Section \ref{sec3:discussion}.

\subsection{Neural Network Architectures}\label{subsec:neural_networks_architectures}

\subsubsection{Fully-Connected Neural Networks}

\begin{figure}[h!]
    \centering
    \includegraphics[width=0.5\textwidth]{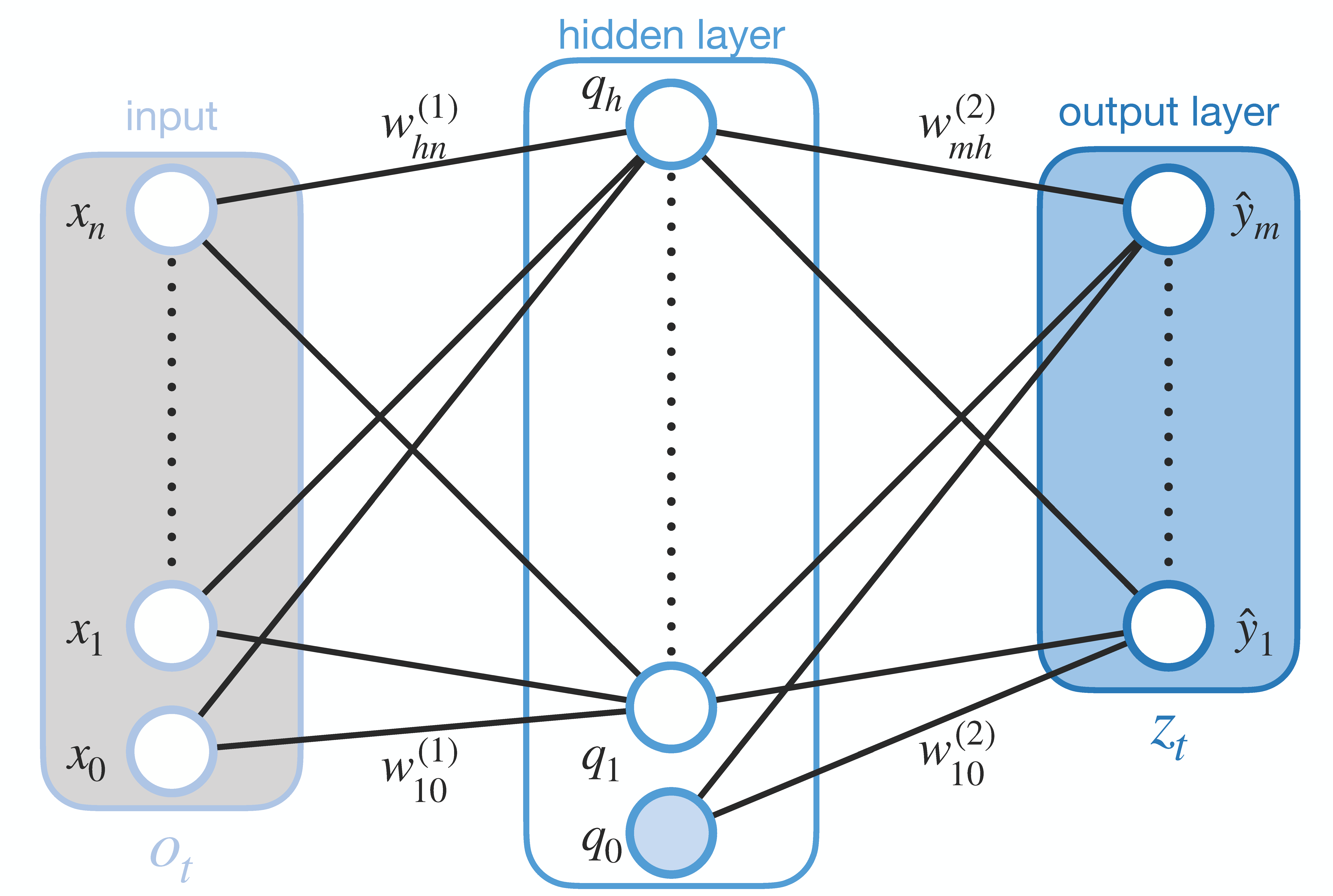}
    \caption{Two-layers \acrshort{fcnn} with $n$ inputs, $h$ hidden neurons and $m$ outputs. The input and output are indicated with $\bm{x}$, and $\bm{y}$, the weights of the network are indicated with $w$ (e.g. $w^{(1)}_{hn}$ corresponds to the weight of the first layer connecting the $n^{th}$-input to the $h^{th}$-hidden node) and the biases with the subscript $0$. In the context of \acrshort{srl}, the input the observation $\bm{o}_t$, while the output is the latent state $\bm{z}_t$. Figure reproduced and adapted from \cite{botteghi2021robotics}.}
    \label{fig:ANN}
\end{figure}

Fully-Connected Neural Networks (\acrshort{fcnn}s), sometimes called Multi-Layer Perceptrons or Feedforward Neural Networks, \cite{goodfellow2016deep} lay the foundations of \acrshort{dl}. \acrshort{fcnn}s are a family of parametric models used to approximate any unknown function $f: \mathcal{X} \longrightarrow \mathcal{Y}; \bm{x} \longmapsto f(\bm{x})$ by means of a parametric function $\phi$ such that $\bm{y}=\phi(\bm{x}; \bm{\theta})$. The neural network parameters $\bm{\theta}$ are optimized, via a training set of examples and gradient descent, to find the best approximation of $f$. An example of a two-layer \acrshort{fcnn} is shown in Figure \ref{fig:ANN}.

Assuming the \acrshort{fcnn} depicted in Figure \ref{fig:ANN} with $n$ inputs, $h$ \textit{hidden neurons}, or \textit{nodes}, $m$ outputs, a set \textit{weights}  and \textit{biases} $\bm{\theta}$, i.e. the parameters, and an \textit{activation function} $\xi(\cdot)$, we can define the contribution of the $i^{th}$-component of the input $x_i$ to the $k^{th}$-component of the output $y_k$ as:
\begin{equation}
\label{IO_feedforward_nn}
\begin{split}
    q_j &= \xi(\sum^n_{i=1} w^{(1)}_{ji}x_i+w^{(1)}_{j0}) \\
    y_k &= \xi(\sum^h_{j=1} w^{(2)}_{kj}q_j+w^{(2)}_{k0}) \\
\end{split}
\end{equation}

If the activation function $\xi(\cdot)$ is chosen linear, the output is a linear combination of the inputs. However, in most cases, the activation function is chosen nonlinear. \acrshort{fcnn}s with nonlinear activations are universal function approximators \cite{hornik1989multilayer}. Common activation functions are sigmoid, tanh, rectified linear unit (or ReLU), and softmax. 

Independently of the structure of the network, we need a way to learn the parameters $\bm{\theta}$, i.e. weights $w_i$ and biases $b_i$, of the network to best approximate the unknown target function $f$, given a set of training samples. We can do this by means of gradient descent and back-propagation by choosing a suitable loss, or cost function $\mathscr{L}(\cdot)$. From now on, we indicate the loss function as $\mathscr{L}(\bm{\theta})$ to explicitly highlight the dependency of the loss on the parameter vector $\bm{\theta}$.

\subsubsection{Convolutional Neural Networks}

Convolutional neural networks (\acrshort{cnn}s) \cite{lecun1995convolutional} are neural networks specialised in efficiently handling data with grid-like structure, e.g. images (2D pixel grids) or \acrshort{lidar} readings (1D data grids). \acrshort{cnn}s inherit their name from the mathematical operator they employ in at least one of their layers, i.e. the convolution.

Unlike \acrshort{fcnn}s that use full connectivity of each input and output neuron, \acrshort{cnn}s are an example of sparse connectivity when the kernel size is smaller than the input data size. For example, images can contain thousands or millions of pixels, while the kernels have way smaller dimensions. Kernels are used because we are interested in learning to extract meaningful features in the images, such as edges or geometrical shapes, which generally occupy multiple pixels. Convolution allows reducing the number of network parameters, the physical memory required to store it and the number of computations for computing the outputs.

\begin{figure}[h!]
    \centering
   \includegraphics[width=0.75\textwidth]{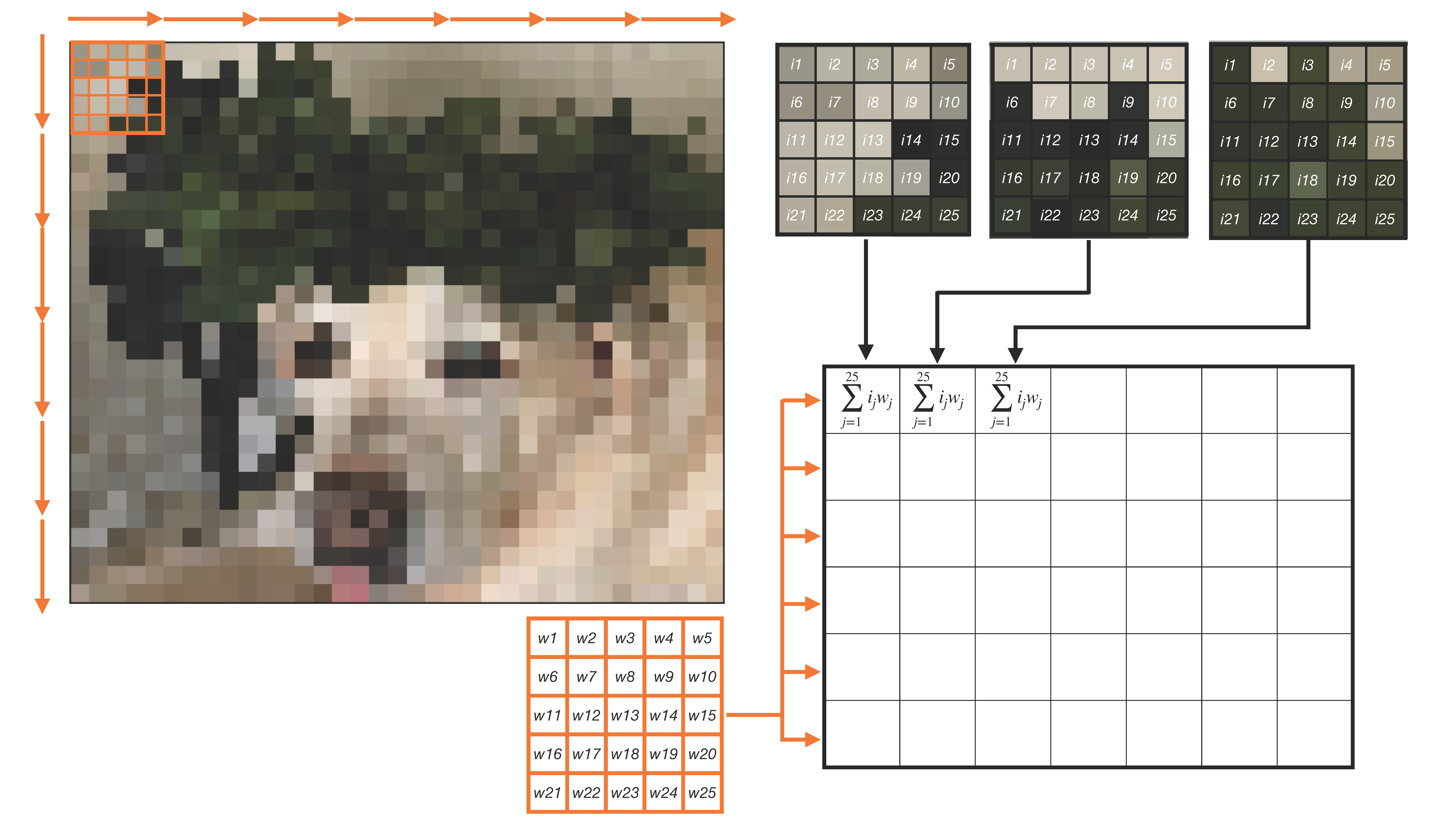}
    \caption{Example of 2D-convolution with a kernel with size $5 \times 5$ and stride $5$. Figure reproduced from \cite{botteghi2021robotics}.} 
    \label{fig:2D-convolution}
\end{figure}

The trainable parameters of \acrshort{cnn}s are the elements of the kernels. By learning the weights of the kernels, it is possible to identify different features of the input data, for example, edges and shapes in the case of 2D images. \acrshort{cnn}s employ parameter sharing to reduce the number of parameters with respect to \acrshort{fcnn}s. Multiple kernels are slid with a certain stride length (the stride is the parameter determining how much the kernel is shifted) across the whole data grid to extract different features. Figure \ref{fig:2D-convolution} shows an example of convolution applied to an image. It is worth highlighting that, because of parameter sharing, \acrshort{cnn} are shift equivariant -- applying a shift transformation to the image before applying the convolution operator is equivalent to applying the convolution and then the shift transformation. As with \acrshort{fcnn}s, multiple convolutional layers can be stacked together. The more the convolutional layer is close to the network's output, the higher the level of abstraction of the learned features (e.g., edges and colours to more complex shapes). 

In the context of \acrshort{srl}, we often deal with image observation. Therefore \acrshort{cnn}s are fundamental architectures for extracting the relevant features, lightening the computational burden of \acrshort{fcnn}s, and effectively reducing the dimensionality.

\subsubsection{Recurrent Neural Networks}

Recurrent neural networks (\acrshort{rnn}s) \cite{pineda1987generalization} are the class of neural networks specialized in handling sequences of data with either fixed or variable length. \acrshort{rnn}s can be used in different contexts, for example, to map sequences to sequences, e.g. in natural language processing, or sequences to a single outputs, e.g. for solving \acrshort{pomdp} in \acrshort{drl} and \acrshort{srl}.

Similar to the case of \acrshort{fcnn}s, we can analytically define the relation between the input vector $\bm{x}$, the hidden state vector $\bm{h}$ and the predicted output $\hat{\bm{y}}$ as:
\begin{equation}
\label{IO_RNN}
\begin{split}
    \bm{h}_t &= \xi(W\bm{h}_{t-1} + U\bm{x}_t + \bm{b})\\
    \hat{\bm{y}}_t &= V\bm{y}_t + \bm{c}\,,\\
\end{split}
\end{equation}
where $W, U, V$ are the weight matrices and $\bm{b}, \bm{c}$ are the bias vectors.

\begin{figure}[h!]
    \centering
   \includegraphics[width=0.5\textwidth]{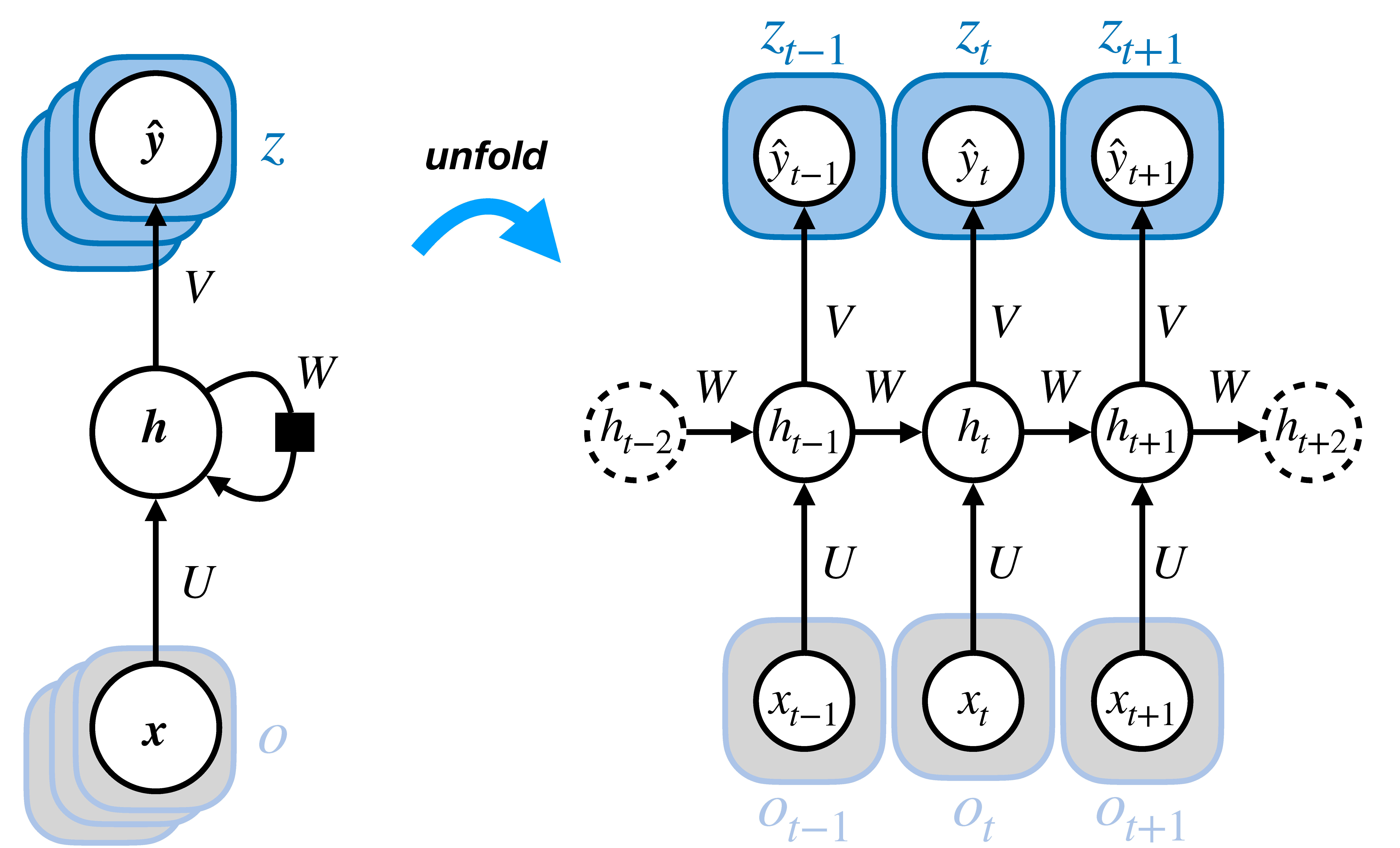}
    \caption{Computational graph of a recurrent neural network. Figure reproduced and adapted from \cite{botteghi2021robotics}.} 
    \label{fig:RNN_computational_graph}
\end{figure}
Similarly to \acrshort{fcnn}s and \acrshort{cnn}s, \acrshort{rnn}s are trained by defining a loss function and by applying back-propagation. A widely-used \acrshort{rnn} architecture is the Long-Short Term Memory \cite{graves2012supervised} using a sophisticated gating mechanism for preventing the vanishing of the gradient problem that affects the simple \acrshort{rnn} architecture.

\subsubsection{Neural Networks in State Representation Learning}

\acrshort{fcnn}s, \acrshort{cnn}s, and \acrshort{rnn} form the building blocks for parameterizing the unknown mapping $\phi_{\text{enc}}$, which encodes observations from the world into a low-dimensional latent state space that can be used to learn the optimal behavior. In (unsupervised) \acrshort{srl}, the set of input data to $\phi_{\text{enc}}$ is the observation stream $\bm{o}_t,\dots,\bm{o}_0$, as we do not have access to the actual output states $\bm{s}_t,\dots,\bm{s}_0$. Therefore, the output predictions of the encoder $\phi_{\text{enc}}$, i.e., the latent states, are indicated by $\bm{z}_t,\dots,\bm{z}_0$. Unfortunately, in the context of \acrshort{srl}, we cannot rely on the true states $\bm{s}$ and directly perform regression. However, we can use auxiliary models to learn the encoder's parameters. We will discuss this aspect in Section \ref{subsec:observation_reconstruction}-\ref{subsec:contrastive_losses}.

\begin{figure*}[h!]
    \centering
        \includegraphics[width=1.0\textwidth]{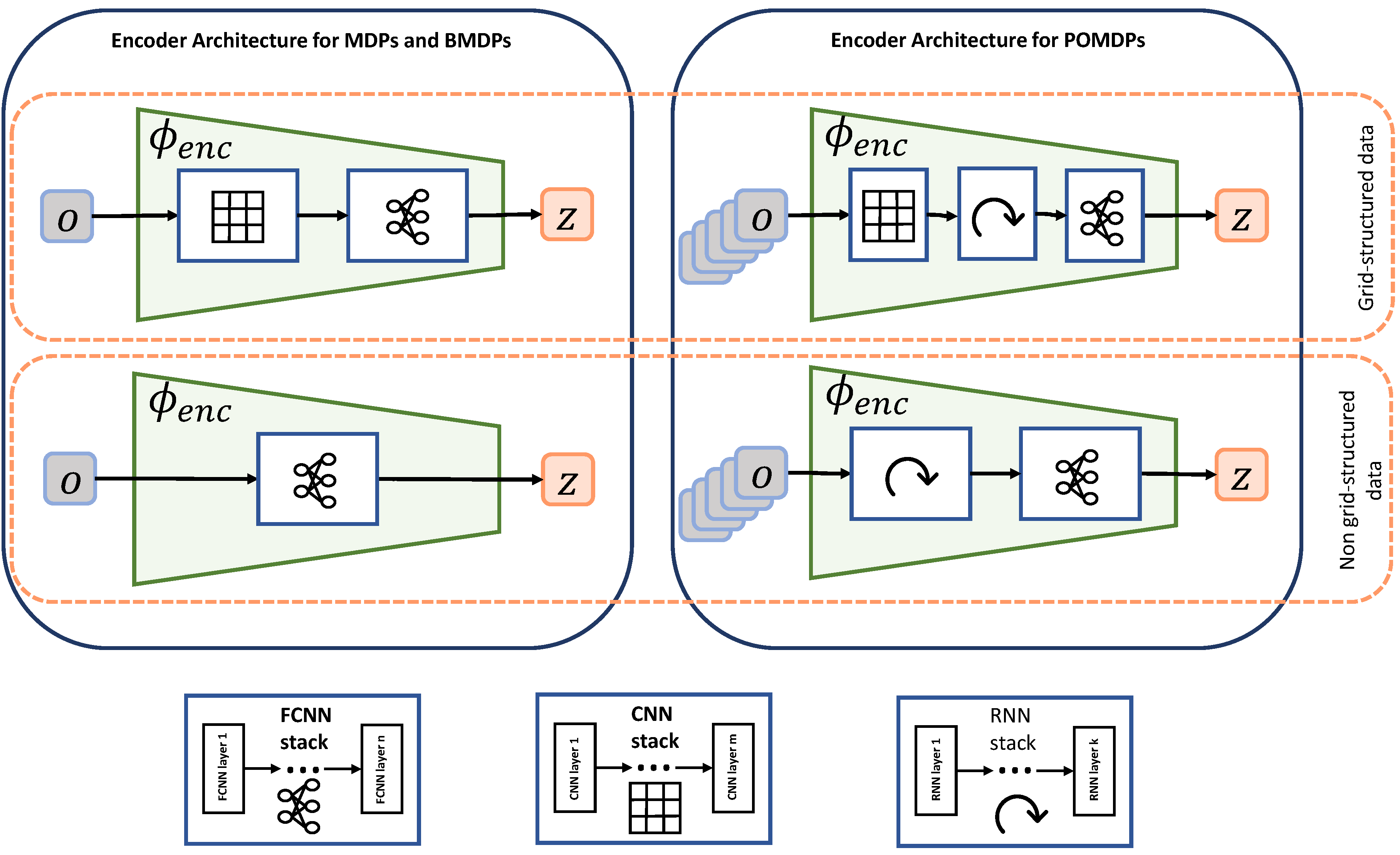}
        \caption{Neural network architectures of the encoder $\phi_{\text{enc}}$ in \acrshort{mdp}s, \acrshort{bmdp}s, and \acrshort{pomdp}s in the presence of grid-structured and non-grid-structured data. \acrshort{fcnn}s are indicated by the graph symbol, \acrshort{cnn}s by the matrix grid, and \acrshort{rnn}s by the looping arrow. The encoder $\phi_{\text{enc}}$ can be either deterministic, i.e. $\bm{z}=\phi_{\text{enc}}(\bm{o})$, or stochastic, i.e.$\bm{z} \sim p(\bm{z}|\bm{o})=\phi_{\text{enc}}(\bm{o})$.}
        \label{fig:architecture_state_encoder}
\end{figure*}

The neural network architecture of $\phi_{\text{enc}}$ has to be chosen depending on the \acrshort{mdp} type and the data structure. In Figure \ref{fig:architecture_state_encoder}, we depict the four most used neural network architectures from the literature categorized based (1) on the \acrshort{mdp} structure, i.e., in the case of \acrshort{mdp}s, \acrshort{bmdp}s or \acrshort{pomdp}s (in \acrshort{pomdp}s, time matters and to reconstruct the latent state space and capture its dynamics we need to rely on sequences of observations) and (2) on the availability of grid-structured data.

\subsection{State Representation Learning Methods}\label{subsec:SRL_methods}

Before diving into the methods, we introduce and discuss the properties of the state representations we aim at learning in Section \ref{subsec:propertieslearningstaterepresentation}. The rest section is articulated in four main blocks:
\begin{itemize}
    \item Section \ref{subsec:observation_reconstruction} describes the approaches for \acrshort{srl} based on  Principal Components Analysis (\acrshort{pca}) \cite{wold1987principal} and observation reconstruction \cite{lecun1987phd, kingma2014stochastic}, 
    \item Section \ref{subsec:MDP_models} presents the \acrshort{mdp} models and describes their usage as auxiliary models for encoding observations into a low-dimensional latent state space,
    \item Section \ref{subsec:robotics_priors} discusses the so-called Robotics Priors \cite{jonschkowski2014state}, and
    \item Section \ref{subsec:contrastive_losses} focuses on contrastive losses \cite{chopra2005learning}.
\end{itemize}
These four blocks will be a recurring theme in the rest of the review.

\subsubsection{Properties of the Learned State Representations}\label{subsec:propertieslearningstaterepresentation}

Learning compact and meaningful state representations is crucial for the performance of the \acrshort{drl} algorithms. In \acrshort{drl}, the benefits are several:
\begin{itemize}
    \item Sample efficiency. Fast learning from limited data samples is fundamental to \acrshort{drl}'s progress. For example, in robotics, we have limited access to real-world data while we need to learn complicated behaviors. 
    \item Robustness. Due to the reduction of dimensionality and the latent states being in a lower-dimensional space, we are less prone to overfitting and noise.
    \item Generalization. By encoding and disentangling the critical factor for explaining the world from the irrelevant ones, good representations are more prone to generalize to new environments with similar features.
    \item Interpretability. Latent representations are often composed of interpretable features and can often be visualized in 2D or 3D plots. 
\end{itemize}

In the most general case of a \acrshort{pomdp}, the goal is to encode each sequence of high-dimensional observations into a latent state space $\mathcal{Z}$ containing all the relevant information for optimally and efficiently learning to solve the control task. In the case of a \acrshort{bmdp}, we encode single observations to infer the latent states. Here, we discuss which are the desired properties for $\mathcal{Z}$ (the first three properties are general properties from representation learning \cite{Bengio2013}, while the remaining three  properties are specific to \acrshort{srl}), according to \cite{Bengio2013, Bohmer2015, Lesort2018}:
\begin{enumerate}
    \item $\mathcal{Z}$ should be \textit{smooth}. If $\bm{o}_{t_1} \approx \bm{o}_{t_2}$ then $\phi_{\text{enc}}(\bm{o}_{t_1}) \approx \phi_{\text{enc}}(\bm{o}_{t_2})$, i.e., the representation can generalize to similar observations. The same reasoning can be applied in the case of observation sequences $\bm{h}_{t_1} \approx \bm{h}_{t_2}$, as shown by the notation in Equation \eqref{eq:phi_enc_general}.
    
    \item $\mathcal{Z}$ should be a \textit{ low-dimensional manifold}. The latent states are concentrated in a region of much smaller dimensionality than the observation space.
    
    \item elements in $\mathcal{Z}$ should represent the \textit{complex dependencies} arising from the study of complex systems, possibly in a simple manner. For example, the transition function between latent states should be a relatively simple (quasi-linear) transformation.
    
    \item $\mathcal{Z}$ should be \textit{Markovian}. An \acrshort{drl} algorithm without memory should be able to solve the problem using the state representation optimally.
    
    \item $\mathcal{Z}$ should be \textit{temporally coherent}. Consecutive observations should be mapped closer in the latent space than non-consecutive ones.
    
    \item $\mathcal{Z}$ should be \textit{sufficiently expressive} and contain enough information to allow the agent to learn the optimal policy.
\end{enumerate}
We will link these properties to the methods in Section \ref{sec3:discussion}, and to the evaluation metrics in Section \ref{subsec:EvalMetrics}.

\subsubsection{Observation Encoding}\label{subsec:observation_reconstruction}

Assuming \marginpar{\scriptsize Principal \hbox{Components} \hbox{Analysis}} a high-dimensional observation space, one can project the observations into a lower-dimensional space. The projection must maintain all (most of) the relevant information while discarding the irrelevant ones. This process can be done by considering the covariance matrix of the data, computing the eigenvalues and projecting the dataset in the lower-dimensional space defined by the $n$ eigenvector directions with the highest variance, i.e. the eigenvectors best explaining the data. This procedure is called \acrshort{pca} \cite{wold1987principal}. \acrshort{pca} allows reconstructing observations by projecting back (inverse mapping) to the original observation space the eigenvalues that mostly explaining the variance of the data. One of the drawbacks of \acrshort{pca} is that the data are linearly projected into a low-dimensional space. The use of a linear projection hinders the application of \acrshort{pca} to cases where a linear map is not sufficiently expressive to learn a meaningful state representation. While in principle kernel \acrshort{pca} \cite{scholkopf1997kernel} could be used to find a nonlinear projection mapping, we found no evidence of its use in \acrshort{srl}. 

\begin{figure}[h!]
\centering
\includegraphics[width=0.6\textwidth]{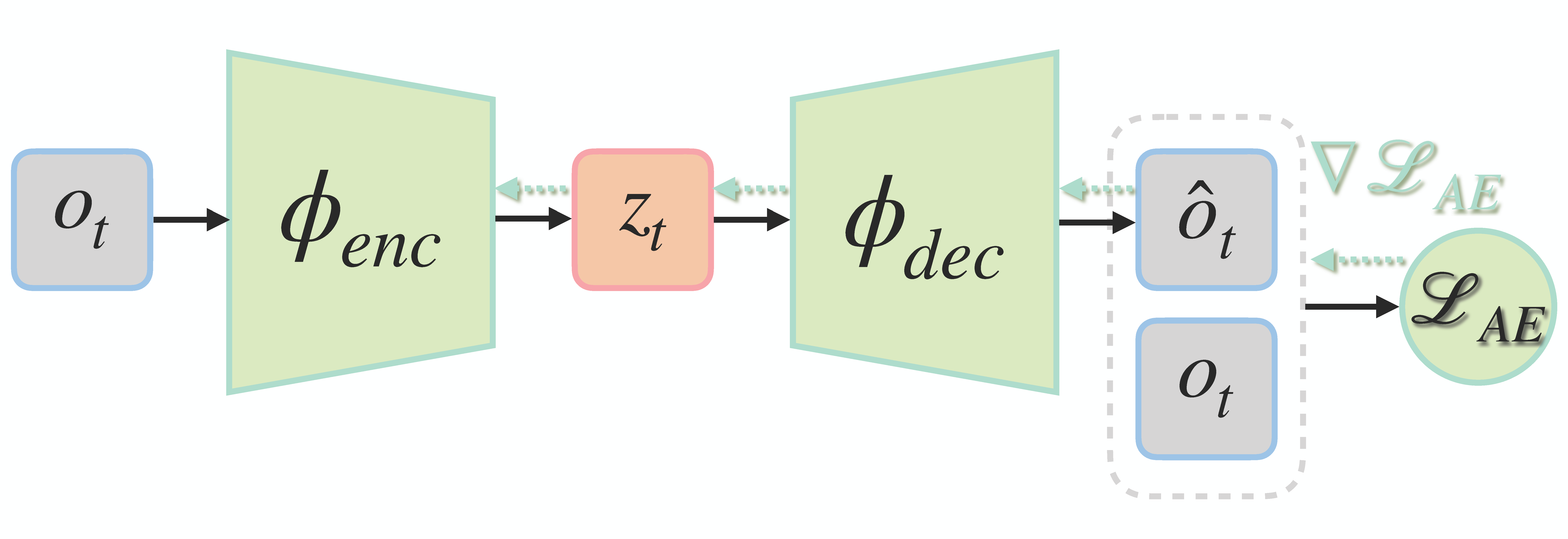}
\caption{Autoencoder architecture. The encoder $\phi_{\text{enc}}$ projects observations $\bm{o}$ to lower-dimensional latent state variables $\bm{z}$, while the decoder  $\phi_{\text{dec}}$ takes the latent states and reconstructs the observations. \acrshort{pca} can be seen as a linear encoder, projecting observations to latent states. The solid line represents the forward pass of the data through the networks, while the dashed line represents the gradient flow.}
\label{fig:AE_architecture}
\end{figure}

Autoencoders\marginpar{\scriptsize Autoencoders} (\acrshort{ae}s) \cite{lecun1987phd} are neural networks composed of an encoder $\phi_{\text{enc}}$, with parameters $\bm{\theta}_{\text{enc}}$, and a decoder $\phi_{\text{dec}}$, with parameters $\bm{\theta}_{\text{dec}}$. The encoder maps the input data $\bm{o}_t$ to a low-dimensional latent representation or abstract code $\bm{z}_t=\phi_{\text{enc}}(\bm{o}_t;\bm{\theta}_{\text{enc}})$, while the decoder reconstructs the input signal $\hat{\bm{o}}_t = \phi_{\text{dec}}(\bm{z}_t; \bm{\theta}_{\text{dec}})$ given the latent representation $\bm{z}_t$, see Figure \ref{fig:AE_architecture}. 
The low-dimensionality of the latent representations is a form of regularisation for preventing the learning of the trivial identity mapping. For more information, we refer the reader to \cite{Bengio2013}. The \acrshort{ae} is trained by minimizing the so-called \textit{reconstruction loss}, i.e. the Mean Squared Error (\acrshort{mse}) loss between the original input and the reconstructed one, as shown in Equation \eqref{eq:reconstruction loss}).
\begin{equation}
\begin{split}
   \min_{\bm{\theta}_{\text{enc}},\bm{\theta}_{\text{dec}}} &\ \ \ \ \mathscr{L}_{AE}(\bm{\theta}_{\text{enc}}, \bm{\theta}_{\text{dec}})\\ 
    \mathscr{L}_{AE}(\bm{\theta}_{\text{enc}}, \bm{\theta}_{\text{dec}})&=\mathbb{E}_{\bm{o}_t\sim \mathcal{D}}[\mid \mid \bm{o}_t - \hat{\bm{o}}_t\mid \mid^2]\\ &=\mathbb{E}_{\bm{o}_t\sim \mathcal{D}}[\mid \mid \bm{o}_t - \phi_{\text{dec}}(\bm{z}_t;\bm{\theta}_{\text{dec}})\mid \mid^2]\\
    &= \mathbb{E}_{\bm{o}_t\sim \mathcal{D}}[\mid \mid \bm{o}_t - \phi_{\text{dec}}(\phi_{\text{enc}}(\bm{o}_t;\bm{\theta}_{\text{enc}});\bm{\theta}_{\text{dec}})\mid \mid^2]\,,
\end{split}
    \label{eq:reconstruction loss}
\end{equation}
where $\bm{o}_t$ is the original input observation, $\hat{\bm{o}}_t$ is its reconstruction using the \acrshort{ae}, and $\mathcal{D}$ is the memory buffer (as in standard \acrshort{drl}, the memory buffer collects the samples obtained by the agent during the interaction with the environment and allows their re-use for training the neural networks) containing the collected samples. In \acrshort{drl}, the encoder of the \acrshort{ae} is used to learn state representations from high-dimensional observations. Inputs can be either single observations, e.g. in \acrshort{bmdp}s, or sequences of observations, e.g. in \acrshort{pomdp}s. The latent state variables are then used to learn policies and value functions. \acrshort{ae}s tend to learn clusters of the latent state variables based on the similarities of the observations. However, these clusters may be disjoint and far from each other due to the lack of regularization of the latent state space (\acrshort{ae}s use no regularization loss). A non-regularized latent state space does not allow a good generation of observations through the decoder given a randomly sampled latent state. Being able to generate observations may be necessary in scenarios where limited data is available, for example, in real-world robotics learning. Additionally, data may be noisy, and we would like to find a way to quantify the uncertainties on our latent state variables.

\begin{figure}[h!]
        \centering
        \includegraphics[width=0.6\textwidth]{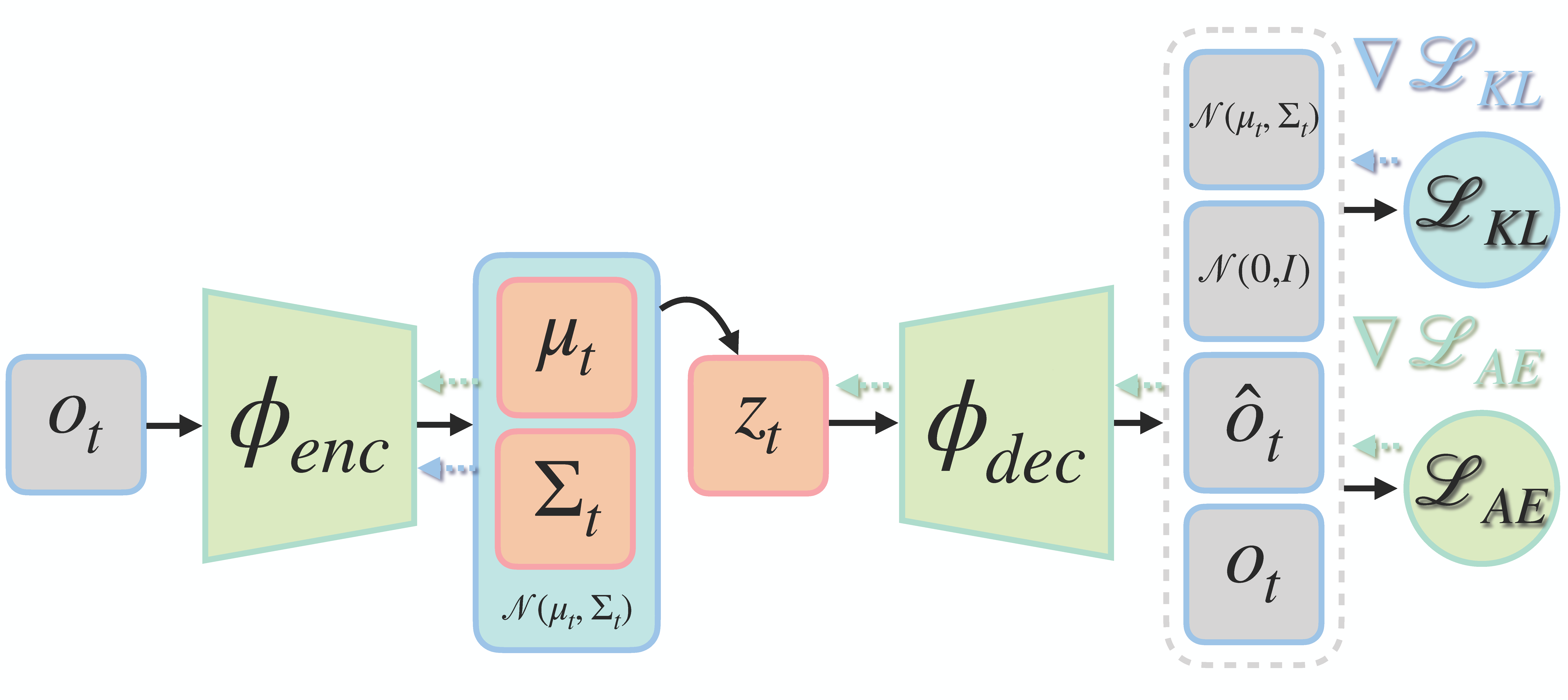}
        \caption{Variational autoencoder architecture. The solid line represents the forward pass of the data through the networks, while the dashed line represents the gradient flow.}
        \label{fig:VAE_architecture}
\end{figure}

In\marginpar{\scriptsize Variational \hbox{Autoencoders}} these scenarios, \acrshort{ae}s may be replaced with Variational Autoencoders (\acrshort{vae}s) \cite{kingma2014stochastic}. Similarly to \acrshort{ae}s, \acrshort{vae}s are composed of an encoder and a decoder. However, differently from \acrshort{ae}s, the latent space is modelled as a Normal distribution $\mathcal{N}(\bm{\mu}_t,\Sigma_t;\bm{\theta}_{\text{enc}})$ with the encoder learning its statistics $\bm{\mu}_t$ and $\Sigma_t$ given an observation $\bm{o}_t$. The latent state $\bm{z}_t$ is sampled from $\mathcal{N}(\bm{\mu}_t,\Sigma_t;\bm{\theta}_{\text{enc}})$, often using the so-called reparameterization trick \cite{kingma2014stochastic}. Analogously to the case of an \acrshort{ae}, $\bm{z}_t$ is then decoded via $\phi_{\text{dec}}$ to reconstruct the input observation $\hat{\bm{o}}_t$. The \acrshort{vae} architecture is depicted in Figure \ref{fig:VAE_architecture}.
The \acrshort{vae} loss function, shown in Equation \eqref{eq:VAE_loss}, is composed of two terms: a reconstruction loss, similar to the \acrshort{ae}, e.g.  maximum likelihood, and  a Kullbach-Leibler divergence (\acrshort{kl}) regularization term to enforce that the latent state distribution $\mathcal{N}(\bm{\mu}_t,\Sigma_t;\bm{\theta}_{\text{enc}})$ is close to the Normal distribution $\mathcal{N}(\bm{0},I)$.
\begin{equation}
\begin{split}
   \min_{\bm{\theta}_{\text{enc}},\bm{\theta}_{\text{dec}}} &\ \ \ \ \mathscr{L}_{AE}(\bm{\theta}_{\text{enc}}, \bm{\theta}_{\text{dec}}) + \mathscr{L}_{KL}(\bm{\theta}_{\text{enc}})\\ 
    \mathscr{L}_{AE}(\bm{\theta}_{\text{enc}}, \bm{\theta}_{\text{dec}})&= \mathbb{E}_{\bm{o}_t\sim \mathcal{D}}[-\log \phi_{\text{dec}}(\bm{o}_t|\bm{z}_t;\bm{\theta}_{\text{enc}},\bm{\theta}_{\text{dec}})]\\
    \mathscr{L}_{KL}(\bm{\theta}_{\text{enc}}) &=\mathbb{E}_{\bm{o}_t\sim \mathcal{D}}[ \text{KL}(\mathcal{N}(\bm{\mu}_t,\Sigma_t;\bm{\theta}_{\text{enc}})||\mathcal{N}(\bm{0},I))] \\
\end{split}
    \label{eq:VAE_loss}
\end{equation}

The relative weight between reconstruction and regularization loss  is a critical parameter for the performance of the \acrshort{vae}. A commonly used variation of the \acrshort{vae} framework is the so-called $\beta$-\acrshort{vae} \cite{higgins2016beta}. $\beta$-\acrshort{vae}s are special \acrshort{vae}s aiming to learn representations that disentangle the underlying factors of variation. This can be done by introducing a parameter $\beta$ in the loss function in Equation \eqref{eq:VAE_loss} balancing reconstruction and regularization loss as follows:
\begin{equation}
\begin{split}
   \min_{\bm{\theta}_{\text{enc}},\bm{\theta}_{\text{dec}}} &\ \ \ \ \mathscr{L}_{AE}(\bm{\theta}_{\text{enc}}, \bm{\theta}_{\text{dec}}) + \beta \mathscr{L}_{KL}(\bm{\theta}_{\text{enc}})\,,\\ 
\end{split}
    \label{eq:betaVAE_loss}
\end{equation}
where $\beta > 1$  enforces a stronger regularization of the latent state space, limits the capacity of the encoder in using the latent variables to encode features, and consequently promotes the learning of disentangled representations. Notice that for $\beta=1$ we obtain the \acrshort{vae} formulation of \cite{kingma2014stochastic}. 

The major flaw of the observation-reconstruction methods is intrinsic in the reconstruction loss itself, see Equation \eqref{eq:reconstruction loss}-\eqref{eq:betaVAE_loss}. The loss does not enforce any discrimination among features relevant for the control and features that are irrelevant. For example, images may contain many features, such as background textures, that are not relevant for learning the optimal policy, yet crucial for minimizing the reconstruction loss.

\subsubsection{Latent MDP models}\label{subsec:MDP_models}

The state representations we seek to learn in \acrshort{drl} must include all the relevant information for learning to make optimal decisions. Therefore, learning approaches for \acrshort{srl} have to be able to discriminate between task-relevant and task-irrelevant features. 

Instead of relying only on observation reconstruction, we can focus on encoding the relevant features for predicting the evolution of the world in the latent state space: the \acrshort{mdp} models. The \acrshort{mdp} models include:
\begin{itemize}
    \item the \textit{forward} or \textit{transition model} predicting the next state given the current one and the action taken,
    \item  the \textit{reward model} predicting the reward given the current state and action taken (note that the reward model may only be a function of the state, but for the sake of generality we introduce it as a function of state-action pairs), and
    \item  the  \textit{inverse model} predicting the action between pairs of consecutive states.
\end{itemize}

These latent models are parameterized with shallow \acrshort{fcnn} (only a few layers) because they deal with low-dimensional latent variables. In the context of \acrshort{srl}, the \acrshort{mdp} models act on latent state variables to describe the latent dynamics of the systems. Because these models utilize latent variables, we refer to them as latent \acrshort{mdp} models. However, good latent \acrshort{mdp} models are generally hard to learn, especially from high-dimensional noisy inputs. In the following paragraphs, we explicitly focus on the latent forward model, the latent reward model, and the latent inverse model. These latent models are learned using the tuples $(\bm{o}_t, \bm{a}_t, \bm{o}_{t+1}, r_t)$ from the memory buffer. Eventually, we discuss different combinations of the \acrshort{mdp} models that lay the foundation of two important theoretical frameworks in \acrshort{drl}: the \textit{\acrshort{mdp} homomorphism} and the \textit{bisimulation metric}.

 The latent \marginpar{\scriptsize Forward Model} forward model $T$ aims to describe the system's forward evolution by predicting the next latent state $\bm{z}_{t+1}$ from the current latent state $\bm{z}_t$ and action $\bm{a}_t$. Depending on the context, the forward model can be either deterministic, i.e. $\bm{z}_{t+1}=T(\bm{z}_t, \bm{a}_t)$, or stochastic. For the sake of conciseness, we present the latent \acrshort{mdp} models in deterministic settings where a deterministic encoder maps the observations to the latent variables $\bm{z}_t=\phi_{\text{enc}}(\bm{o}_t;\bm{\theta}_{\text{enc}})$ and the latent variables are fed to deterministic latent models to predict the latent dynamics. {\color{black}However, the methods presented here can be directly extended to the stochastic case by assuming $\bm{z}_t\sim p(\bm{z}_t|\bm{o}_t)=\phi_{\text{enc}}(\bm{o}_t;\bm{\theta}_{\text{enc}})$ and $\bm{z}_{t+1}\sim p(\bm{z}_{t+1}|\bm{z}_{t}, \bm{a}_{t})=\phi_{T}(\bm{z}_t, \bm{a}_t;\bm{\theta}_{T})$, where the two distributions are often assumed to be Gaussian).

In deterministic settings, see Figure \ref{fig:forward_model}, the latent forward model $T$, parameterized by a \acrshort{fcnn} with parameters $\bm{\theta}_T$, maps $\bm{z}_t$ and $\bm{a}_t$ to $\bm{z}_{t+1}=T(\bm{z}_t,\bm{a}_t;\bm{\theta}_T)$. 
\begin{figure}[h!]
    \centering
    \includegraphics[width=0.6\textwidth]{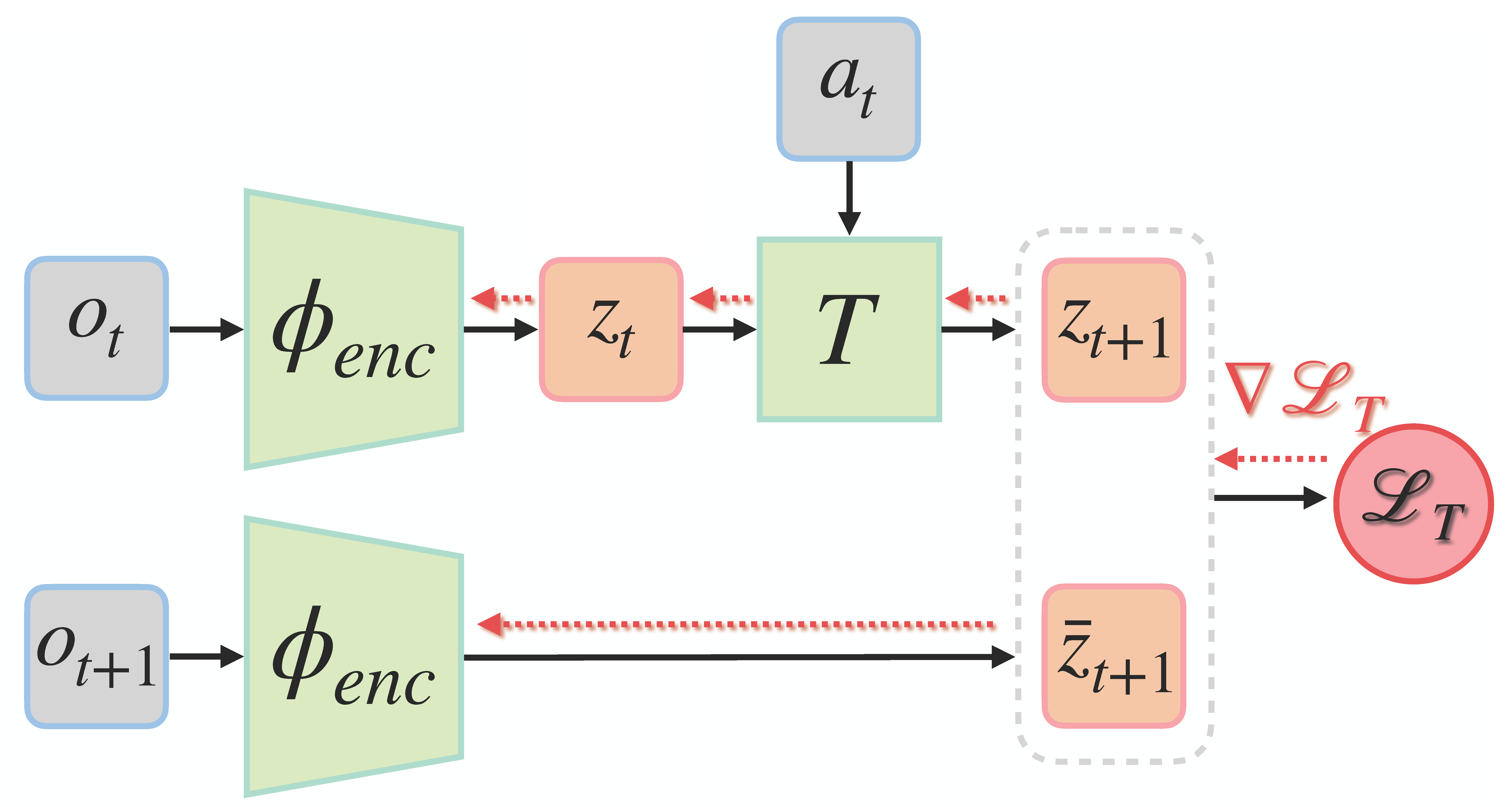}
    \caption{Latent forward model. The solid line represents the forward pass of the data through the networks, while the dashed line represents the gradient flow.}
    \label{fig:forward_model}
\end{figure}
The latent forward model can be used, and it is in practice often used, to compute an auxiliary loss whose gradients are employed to train the encoder $\phi_{\text{enc}}$. The loss, for the deterministic formulation of the models}, is shown in Equation \eqref{eq:forward_model_loss}. 
\begin{equation}
\begin{split}
   \min_{\bm{\theta}_{\text{enc}},\bm{\theta}_{T}} &\ \ \ \ \mathscr{L}_{T}(\bm{\theta}_{\text{enc}}, \bm{\theta}_T)\\ 
    \mathscr{L}_{T}(\bm{\theta}_{\text{enc}}, \bm{\theta}_{T})&=\mathbb{E}_{\bm{o}_t,\bm{a}_t,\bm{o}_{t+1}\sim \mathcal{D}}[\mid \mid \Bar{\bm{z}}_{t+1} - \bm{z}_{t+1} \mid \mid^2] \\
    &=\mathbb{E}_{\bm{o}_t,\bm{a}_t,\bm{o}_{t+1}\sim \mathcal{D}}[\mid \mid \phi_{\text{enc}}(\bm{o}_{t+1};\bm{\theta}_{\text{enc}}) - T(\bm{z}_t, \bm{a}_t;\bm{\theta}_{T})\mid \mid^2]\\
    &= \mathbb{E}_{\bm{o}_t,\bm{a}_t,\bm{o}_{t+1}\sim \mathcal{D}}[\mid \mid \phi_{\text{enc}}(\bm{o}_{t+1};\bm{\theta}_{\text{enc}}) - T(\phi_{\text{enc}}(\bm{o}_t;\bm{\theta}_{\text{enc}}), \bm{a}_t;\bm{\theta}_{T})\mid \mid^2]\,,
\end{split}
    \label{eq:forward_model_loss}
\end{equation}
where the target next latent state $\Bar{\bm{z}}_{t+1}=\phi_{\text{enc}}(\bm{o}_{t+1};\bm{\theta}_{\text{enc}})$ is generated by encoding the next observation $\bm{o}_{t+1}$, while $\bm{z}_{t+1}=T(\bm{z}_t,\bm{a}_t;\bm{\theta}_{T})$ is the predicted next latent state given the current observation $\bm{o}_t$ and the action taken $\bm{a}_t$. Remember that we are studying unsupervised representation learning and we do not have access to the true next states of the environment but only to next (indirect) observations.

Because the target next state $\Bar{\bm{z}}_{t+1}$ is not known and generated by the same encoder, the representation may collapse to trivial embeddings. To prevent this issue other auxiliary losses such as contrastive losses have to be used \cite{kipf2019contrastive, franccois2019combined, van2020plannable}. We discuss contrastive losses in Section \ref{subsec:contrastive_losses}.

\begin{figure}[h!]
    \centering
    \includegraphics[width=0.6\textwidth]{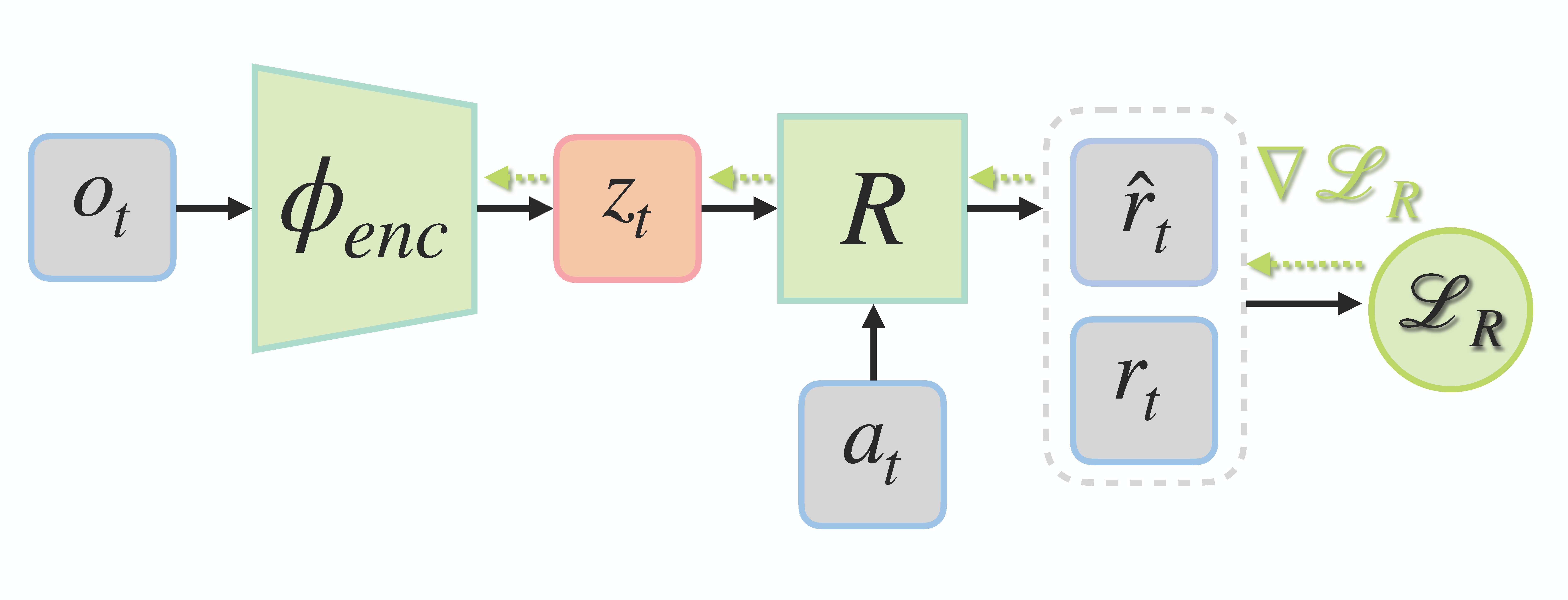}
    \caption{Latent Reward Model. The solid line represents the forward pass of the data through the networks, while the dashed line represents the gradient flow.}
    \label{fig:reward_model}
\end{figure}

To\marginpar{\scriptsize \hbox{Reward} Model} encode task-relevant information into the state representation, we can make use of a latent reward model $R$ to predict the reward $\hat{r}_t=R(\bm{z}_t, \bm{a}_t)$ of a latent state-action pair $(\bm{z}_t, \bm{a}_t)$ (see Figure \ref{fig:reward_model}).

The reward model utilizes the reward samples $r_t$ as target values for regression. In the deterministic case, an example of reward model loss is shown in Equation \eqref{eq:reward_model_loss}).
\begin{equation}
\begin{split}
   \min_{\bm{\theta}_{\text{enc}},\bm{\theta}_{R}} &\ \ \ \ \mathscr{L}_{{R}}(\bm{\theta}_{\text{enc}}, \bm{\theta}_{R})\\ 
    \mathscr{L}_{{R}}(\bm{\theta}_{\text{enc}}, \bm{\theta}_{R})&=\mathbb{E}_{\bm{o}_t,\bm{a}_t,r_t\sim \mathcal{D}}[\mid \mid r_t - \Hat{r}_{t} \mid \mid^2] \\
    &= \mathbb{E}_{\bm{o}_t,\bm{a}_t,r_t\sim \mathcal{D}}[\mid \mid r_t - R(\bm{z}_t, \bm{a}_t;\bm{\theta}_{R})\mid \mid^2]\\
    &= \mathbb{E}_{\bm{o}_t,\bm{a}_t,r_t\sim \mathcal{D}}[\mid \mid r_t - R(\phi_{\text{enc}}(\bm{o}_t;\bm{\theta}_{\text{enc}}), \bm{a}_t;\bm{\theta}_{R})\mid \mid^2]\,,
\end{split}
    \label{eq:reward_model_loss}
\end{equation}
where $r_t$ is the reward sample from the memory buffer $\mathcal{D}$, and $\hat{r}_t = R(\bm{z}_t, \bm{a}_t;\bm{\theta}_{R},\bm{\theta}_{\text{enc}})$ is the reward predicted by the reward model $R$, parameterized by a neural network with parameters $\bm{\theta}_{R}$.

However, in the case of sparse rewards, the reward model alone may not be sufficient to learn a good representation as we may not be able to generate sufficiently rich gradients for training the models. Therefore, usually the latent reward loss (Equation \eqref{eq:reward_model_loss}) is used in combination with additional models, e.g. latent forward models, and contrastive losses \cite{van2020plannable}.

\begin{figure}[h!]
    \centering
    \includegraphics[width=0.6\textwidth]{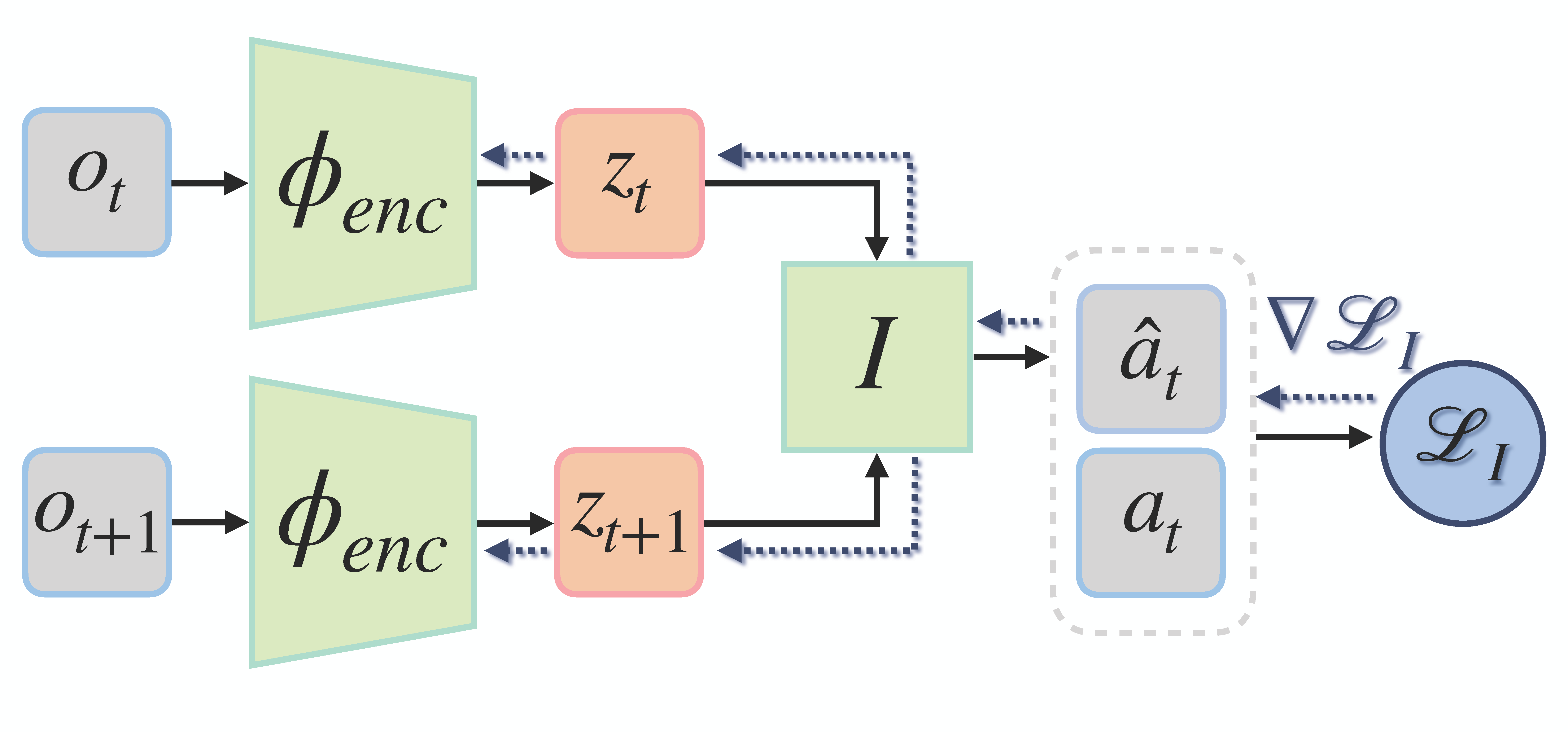}
    \caption{Latent inverse model. The solid line represents the forward pass of the data through the networks, while the dashed line represents the gradient flow.}
    \label{fig:inverse_model}
\end{figure}

Eventually\marginpar{\scriptsize Inverse Model}, we can learn a latent inverse model $I$, again usually modelled with a neural network, predicting the action $\bm{a}_t$ connecting two consecutive latent states $\bm{z}_t$ and $\bm{z}_{t+1}$. The inverse model aids the learning of the features that are needed for predicting the action, i.e. the features the agent can control and change. Figure \ref{fig:inverse_model} depicts the latent inverse model.

In \acrshort{drl}, the action space can be continuous or discrete. Therefore, the loss function for training the latent inverse model may be either the \acrshort{mse} loss (see Equation \eqref{eq:inverse_model_loss}) for continuous action spaces or, for discrete action spaces, the cross-entropy loss used for solving multi-class classification problems (see Equation \eqref{eq:inverse_model_loss_discrete}). 
\begin{equation}
\begin{split}
   \min_{\bm{\theta}_{\text{enc}},\bm{\theta}_{I}} &\ \ \ \ \mathscr{L}_{{I}}(\bm{\theta}_{\text{enc}}, \bm{\theta}_{I})\\ 
    \mathscr{L}_{{I}}(\bm{\theta}_{\text{enc}}, \bm{\theta}_{I})
    &=\mathbb{E}_{\bm{o}_t,\bm{a}_t,\bm{o}_{t+1}\sim \mathcal{D}}[\mid \mid \bm{a}_t - \Hat{\bm{a}}_{t} \mid \mid^2] \\
    &= \mathbb{E}_{\bm{o}_t,\bm{a}_t,\bm{o}_{t+1}\sim \mathcal{D}}[\mid \mid \bm{a}_t - I(\bm{z}_t,\bm{z}_{t+1};\bm{\theta}_{I})\mid \mid^2] \\
    &= \mathbb{E}_{\bm{o}_t,\bm{a}_t,\bm{o}_{t+1}\sim \mathcal{D}}[\mid \mid \bm{a}_t - I(\phi_{\text{enc}}(\bm{o}_t;\bm{\theta}_{\text{enc}}), \phi_{\text{enc}}(\bm{o}_{t+1};\bm{\theta}_{\text{enc}});\bm{\theta}_{I})\mid \mid^2]\,,
\end{split}
    \label{eq:inverse_model_loss}
\end{equation}
In the former, shown in Equation \eqref{eq:inverse_model_loss}, the \acrshort{fcnn}, representing the inverse model, directly outputs the value of the action $\bm{a}_t$, while in the latter, shown in Equation \eqref{eq:inverse_model_loss_discrete}, the network outputs a distribution over the discrete actions.
\begin{equation}
\begin{split}
   \min_{\bm{\theta}_{\text{enc}},\bm{\theta}_{I}} &\ \ \ \ \mathscr{L}_{{I}}(\bm{\theta}_{\text{enc}}, \bm{\theta}_{I})\\ 
    \mathscr{L}_{{I}}(\bm{\theta}_{\text{enc}}, \bm{\theta}_{I})
    &=\mathbb{E}_{\bm{o}_t,\bm{a}_t,\bm{o}_{t+1}\sim \mathcal{D}}[-\sum_{i \in |\mathcal{A}|} a^i_t\,\log \hat{\bm{a}}_t] \\
    &= \mathbb{E}_{\bm{o}_t,\bm{a}_t,\bm{o}_{t+1}\sim \mathcal{D}}[-\sum_{i \in |\mathcal{A}|} a^i_t\,\log I(\bm{z}_t,\bm{z}_{t+1};\bm{\theta}_{I})]\,,
\end{split}
    \label{eq:inverse_model_loss_discrete}
\end{equation}
where $\bm{a}_t$ is the action label, $a^i_t$ the $i^{th}$-component of the action $\bm{a}_t$, and $\hat{\bm{a}}_t$ is a probability value obtained by applying a softmax activation at the output of the latent inverse model $I$. The softmax activation is only used in the context of discrete action spaces and in combination with the cross-entropy loss. In the continuous action spaces, the output of the inverse model is either bounded, e.g. by a tanh, or unbounded, e.g. via a linear activation.

Latent forward\marginpar{\scriptsize Multiple \hbox{Objectives}}, inverse, and reward models are often used and optimized jointly for improving the quality of the learned state representations as shown in Equation \eqref{eq:mdp_models_lossSRL}. 
\begin{equation}
\begin{split}
    \min_{\bm{\theta}_{\text{enc}},\bm{\theta}_{T}, \bm{\theta}_{I},\bm{\theta}_{R}} \ \ \ \ &\omega_{{T}}\mathscr{L}_{{T}}(\bm{\theta}_{\text{enc}}, \bm{\theta}_{T}) + \omega_{{I}}\mathscr{L}_{{I}}(\bm{\theta}_{\text{enc}}, \bm{\theta}_{I}) + \omega_{{R}}\mathscr{L}_{{R}}(\bm{\theta}_{\text{enc}}, \bm{\theta}_{R})\,,
\end{split}
\label{eq:mdp_models_lossSRL}
\end{equation}
where $\omega_{{T}}$, $\omega_{{I}}$ and $\omega_{{R}}$ are three scalar factors for scaling the contribution of each loss. Similar loss functions for learning state representations for \acrshort{drl} are used, for example, in \cite{Goroshin2015, van2016stable, Agrawal2016, gelada2019deepmdp}.                                                                           

A\marginpar{\scriptsize MDP \hbox{Homomorphism}} fundamental concept in RL is the notion of \textit{\acrshort{mdp} homomorphism} \cite{ravindran2002model}. The idea is to exploit the symmetries of the problem to transform a high-dimensional (in states and actions) and potentially complex \acrshort{mdp} $\mathcal{M}$ into a low-dimensional \acrshort{mdp} $\bar{\mathcal{M}}$ that is easier to solve. Moreover, because we learn a homomorphism between two \acrshort{mdp}s, this framework guarantees that the optimal policy learned for the low-dimensional \acrshort{mdp} $\bar{\mathcal{M}}$ can be lifted to the original high-dimensional \acrshort{mdp} $\mathcal{M}$. This statement implies that once we learn the optimal policy for $\bar{\mathcal{M}}$, this policy is the optimal policy for $\mathcal{M}$. Below we provide the formal definitions:
\begin{defi}
\textbf{Stochastic MDP Homomorphism} (adapted from \cite{ravindran2002model}\textbf{:}  A stochastic \acrshort{mdp} homomorphism $h$ from an \acrshort{mdp} $\mathcal{M} = \langle \mathcal{S}, \mathcal{A}, T, R  \rangle$ to an \acrshort{mdp} $\mathcal{\Bar{M}} = \langle \mathcal{\Bar{S}}, \mathcal{\Bar{A}}, \Bar{T}, \Bar{R}  \rangle$ is a tuple $\langle f, g_{\bm{s}} \rangle$, with:
\begin{itemize}
    \item $f: \mathcal{S} \longrightarrow \mathcal{\Bar{S}}$
    \item $g_{\bm{s}}: \mathcal{A} \longrightarrow \mathcal{\Bar{A}}$
\end{itemize}
such that the following identities hold:
\begin{equation}
    \forall_{\bm{s}, \bm{s}' \in \mathcal{S}, \bm{a} \in \mathcal{A}} \ \ \ \Bar{T}(f(\bm{s}')|f(\bm{s}), g_{\bm{s}}(\bm{a})) = \sum_{\bm{s}^{\dagger} \in {[\bm{s}']_f}} T(\bm{s}^{\dagger}|\bm{s},a)
    \label{transition_mdpHomo_st}
\end{equation}
\begin{equation}
    \forall_{\bm{s}, \bm{a} \in \mathcal{A}} \ \ \ \Bar{R}(f(\bm{s}), g_{\bm{s}}(\bm{a})) = R(\bm{s}, \bm{a})\,,
    \label{reward_mdpHomo_st}
\end{equation}
where $[\bm{s}']_f$  is the equivalence class of $\bm{s}'$ under $f$.
\end{defi}

\begin{defi}
\textbf{Deterministic MDP Homomorphism} (adapted from \cite{van2020plannable})\textbf{:} A deterministic \acrshort{mdp} homomorphism $h$ from an \acrshort{mdp} $\mathcal{M} = \langle \mathcal{S}, \mathcal{A}, T, R  \rangle$ to an \acrshort{mdp} $\mathcal{\Bar{M}} = \langle \mathcal{\Bar{S}}, \mathcal{\Bar{A}}, \Bar{T}, \Bar{R}  \rangle$ is a tuple $\langle f, g_{\bm{s}} \rangle$, with:
\begin{itemize}
    \item $f: \mathcal{S} \longrightarrow \mathcal{\Bar{S}}$
    \item $g_{\bm{s}}: \mathcal{A} \longrightarrow \mathcal{\Bar{A}}$\,,
\end{itemize}
such that the following identities hold:
\begin{equation}T(\bm{s},\bm{a}) = \bm{s}' \Longrightarrow \Bar{T}(f(\bm{s}), g_{\bm{s}}(\bm{a})) = f(\bm{s}')
    \label{transition_mdpHomo}
\end{equation}
\begin{equation}
    \forall_{\bm{s} \in \mathcal{S}, \bm{a} \in \mathcal{A}} \ \ \ \Bar{R}(f(\bm{s}), g_{\bm{s}}(\bm{a})) = R(\bm{s}, \bm{a}).
    \label{reward_mdpHomo}
\end{equation}
\end{defi}

If Equation \eqref{transition_mdpHomo_st} and \eqref{reward_mdpHomo_st} are satisfied (the same holds for Equation \eqref{transition_mdpHomo} and \eqref{reward_mdpHomo} in deterministic settings), the (stochastic) policy $\bar{\pi}_{\Bar{\mathcal{M}}}$ of the homomorphic image $\bar{\mathcal{M}}$ can be lifted to the original \acrshort{mdp} $\mathcal{M}$, and we denote it as $\pi_{\mathcal{M}}$, such that for any $\bm{a} \in g_{\bm{s}}^{-1}$:
\begin{equation}
    \pi_{\mathcal{M}}(\bm{a}|\bm{s}) = \frac{\bar{\pi}_{\Bar{\mathcal{M}}}(\bar{\bm{a}}|f(\bm{s}))}{\mid g_{\bm{s}}^{-1}(\bar{\bm{a}})\mid}\,,
\end{equation}
where for any $\bm{s}$, $g_{\bm{s}}^{-1}(\bar{\bm{a}})$ denotes the set of actions that have the same image $\bar{\bm{a}} \in \mathcal{\bar{A}}$ under $g_{\bm{s}}$, i.e., $\bar{\bm{a}}=g_{\bm{s}}(\bar{a})$.

\begin{figure}[h!]
    \centering
    \includegraphics[width=0.45\textwidth]{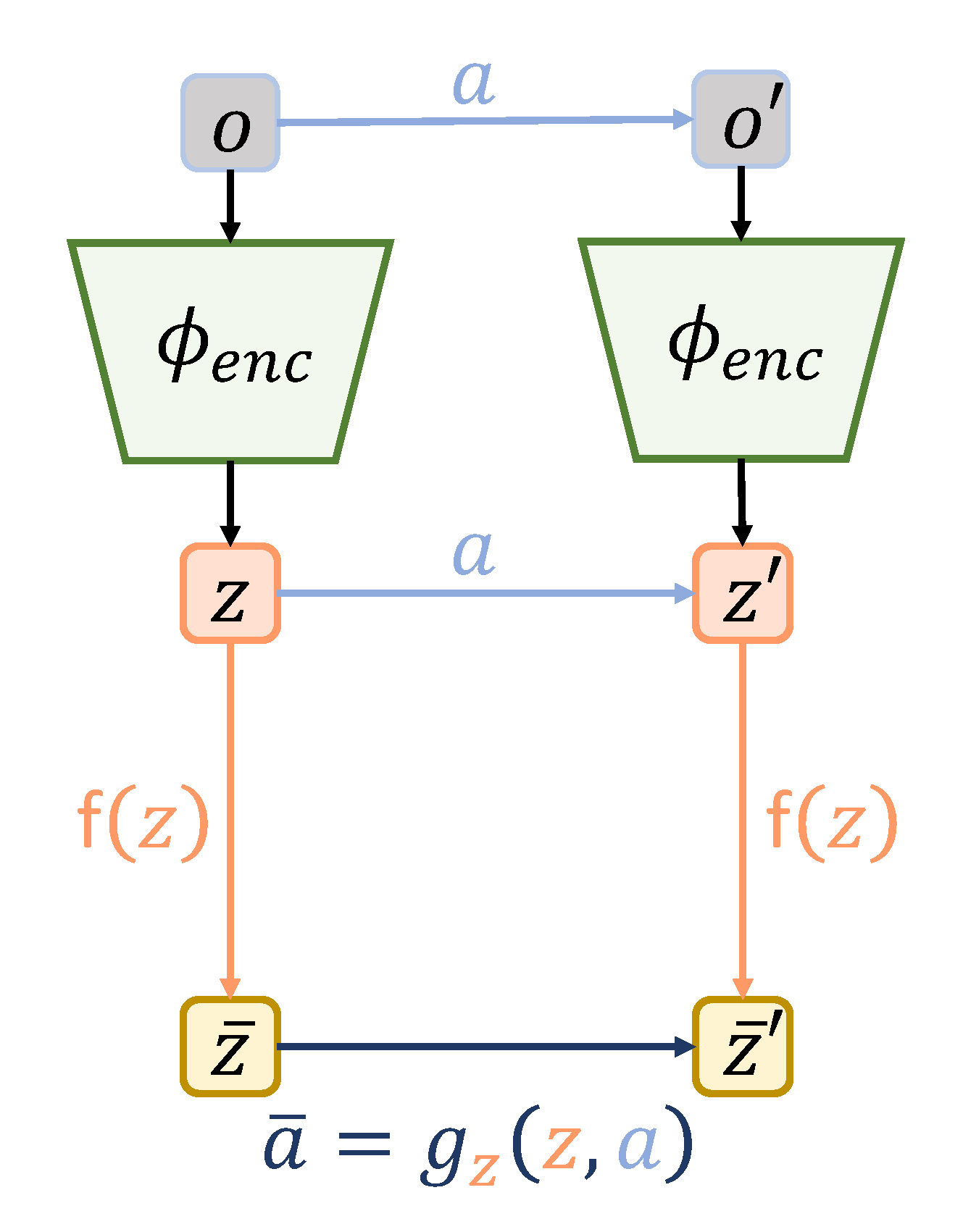}
    \caption{Commutative diagram. Encoding the observation $\bm{o}$ in $\bm{z}$ via the encoder $\phi_{\text{enc}}$ and applying action $a$ is equivalent to abstracting $\bm{z}$ via $f$ and transitioning via $\bar{\bm{a}}$.}
    \label{fig:commutativediagram}
\end{figure}

In the context of \acrshort{srl}, it is possible to leverage the \acrshort{mdp} homomorphism metric to learn the state representation. The \acrshort{mdp} homomorphism metric makes use of {\color{black} an encoder $\phi_{\text{enc}}$}, a transition $\bar{T}$, a reward model $\bar{R}$ to enforce the relation in Equation \eqref{transition_mdpHomo} and \eqref{reward_mdpHomo} (analogously,  Equation \eqref{transition_mdpHomo_st} and \eqref{reward_mdpHomo_st} are used for stochastic \acrshort{mdp}s), and  two additional models, namely $f: \mathcal{Z} \rightarrow \mathcal{\bar{Z}}$ and $g_{\bm{z}}:
\mathcal{Z}\times \mathcal{A}\rightarrow\mathcal{\bar{A}}$ learning (latent) state and state-action symmetries. The advantage of the \acrshort{mdp} homomorphism metric is that, when minimized, i.e., $\mathscr{L}_{\bar{T}} \rightarrow 0$ and $\mathscr{L}_{\bar{R}} \rightarrow 0$, we obtain a \acrshort{mdp} homomorphism exploiting the symmetries of the problem. The policy learned using this representation is guaranteed optimal for the original \acrshort{mdp} too. Moreover, the latent state space is guaranteed to be Markovian. In Figure \ref{fig:commutativediagram}, the commutative diagram representing the \acrshort{mdp} homomorphism is shown.
 However, learning forward and reward models is difficult for complex applications. The losses rarely converge globally and precisely to zero, and in most cases, we end up with an approximate \acrshort{mdp} homomorphism \cite{ravindran2004approximate}.

A\marginpar{\scriptsize Bisimulation \hbox{Metric}} closely-related concept to the \acrshort{mdp} homomorphism that can be used for learning compact representations is the \textit{bisimulation metric} \cite{givan2003equivalence, ferns2004metrics}. For large \acrshort{mdp}, we can think of partitioning the state space into sets of states with equivalent behavior. The bisimulation relation is an equivalence relation $E \subseteq \mathcal{S} \times \mathcal{S}$ that satisfies: 
\begin{equation}
    \bm{s}E\bm{s}^{\dagger} \iff \forall \bm{a} \  R(\bm{s},\bm{a})=R(\bm{s}^{\dagger},\bm{a}) \ \ \  \land  \ \ \  \forall X \in \mathcal{S}/E \ T(X|\bm{s},\bm{a})=T(X|\bm{s}^{\dagger},\bm{a})\,,
\label{eq:bisimulation}
\end{equation}
where $\mathcal{S}/E$ is the partition of $\mathcal{S}$ into $E$-equivalent subset of states. The bisimulation is the union of all the bisimulation relations. From \eqref{eq:bisimulation} follows that two states are bisimilar if they have the same transition and reward and, consequently, the same value function. However, the bisimulation metric is difficult to satisfy completely due to the strict equality in \eqref{eq:bisimulation} and is not robust to small changes in the transition probabilities and rewards.

A \textit{lax bisimulation metric} is introduced in \cite{taylor2008bounding} to exploit symmetries of the environment by relaxing the equality in the bisimulation metric in Equation \eqref{eq:bisimulation}. We can define the lax bisimulation relation $B$ as:
\begin{equation}
    \bm{s}B\bm{s}^{\ddagger} \iff \forall \bm{a} \ \exists \bm{b} \text{ such that } R(\bm{s},\bm{a})=R(\bm{s}^{\ddagger},\bm{b}) \ \ \ \land \ \ \
    \forall X \in S/B \ T(X|\bm{s},\bm{a})=T(X|\bm{s}^{\ddagger},\bm{b})\,,
\label{eq:lax_bisimulation}
\end{equation}
Again, the lax bisimulation is the union of the lax bisimulation relations. Differently from \eqref{eq:bisimulation}, the lax bisimulation metric partitions $\mathcal{S}$ by selecting subset of states with equivalent transition and reward functions but allowing different actions. It is easy to see that the lax bisimulation is equivalent to the \acrshort{mdp} homomorphism \cite{taylor2008bounding} (see Equation \eqref{transition_mdpHomo_st} and \eqref{reward_mdpHomo_st}).

An example of a (deep) bisimulation metric in \acrshort{srl} is introduced in \cite{Zhang2020}, where a state representation is learned such that the $L_1$ distances between pairs of latent states correspond to bisimulation metrics. The encoder $\phi_{\text{enc}}$ is trained by minimizing:
\begin{equation}
\begin{split}
    \min_{\bm{\theta}_{\text{enc}}} &\ \ \ \ \mathscr{L}_{\text{bm}}(\bm{\theta}_{\text{enc}}) \\
    \mathscr{L}_{\text{bm}}(\bm{\theta}_{\text{enc}}) &=  \mathbb{E}_{\bm{o},\bm{a},r,\bm{o}'\sim \mathcal{D}}[(||\bm{z}_{t_1}-\bm{z}_{t_2}||_1 - ||\hat{r}_{t_1}-\hat{r}_{t_2}||_1 
    - \gamma W_2(T(\bm{z}_{t_{1}+1}|\Tilde{\bm{z}}_{t_1}, \bm{a}_{t_1}),T(\bm{z}_{t_{2}+1}|\Tilde{\bm{z}}_{t_2}, \bm{a}_{t_2})))^2]\,,
    \label{eq:bisimulation_loss}
\end{split}
\end{equation}
where $\bm{z}_{t_{1}}=\phi_{\text{enc}}(\bm{o}_{t_{1}})$,  $\bm{z}_{t_{2}}=\phi_{\text{enc}}(\bm{o}_{t_{2}})$, $\Tilde{\bm{z}}$ indicates the stop of the gradients such that they do no propagate through the encoder, $\gamma$ is a scaling factor, and $W_2$ is the 2-Wasserstein metric. {\color{black} It is worth noticing that transition and reward models are used to compute the bisimulation loss in Equation \eqref{eq:bisimulation_loss}. However, these models are not updated jointly with the encoder, but in alternating fashion accordingly to Equation $\eqref{eq:forward_model_loss}$ and \eqref{eq:reward_model_loss}.}

\subsubsection{Priors}\label{subsec:robotics_priors}

When the underlying \acrshort{mdp} model is a dynamical system fully or partially governed by the laws of physics, such as in robotics, we can encode such prior physical knowledge in the form of loss functions and use them to train of the encoder $\phi_{\text{enc}}$ such that the latent states evolve according to these laws. For example, consider a mobile robot navigating based on onboard sensory readings; the underlying actual state space has specific properties dictated by the laws of physics. Therefore, we can shape the loss functions using the physical laws to learn smooth and coherent state representations.

\begin{figure}[h!]
    \centering
    \includegraphics[width=0.6\textwidth]{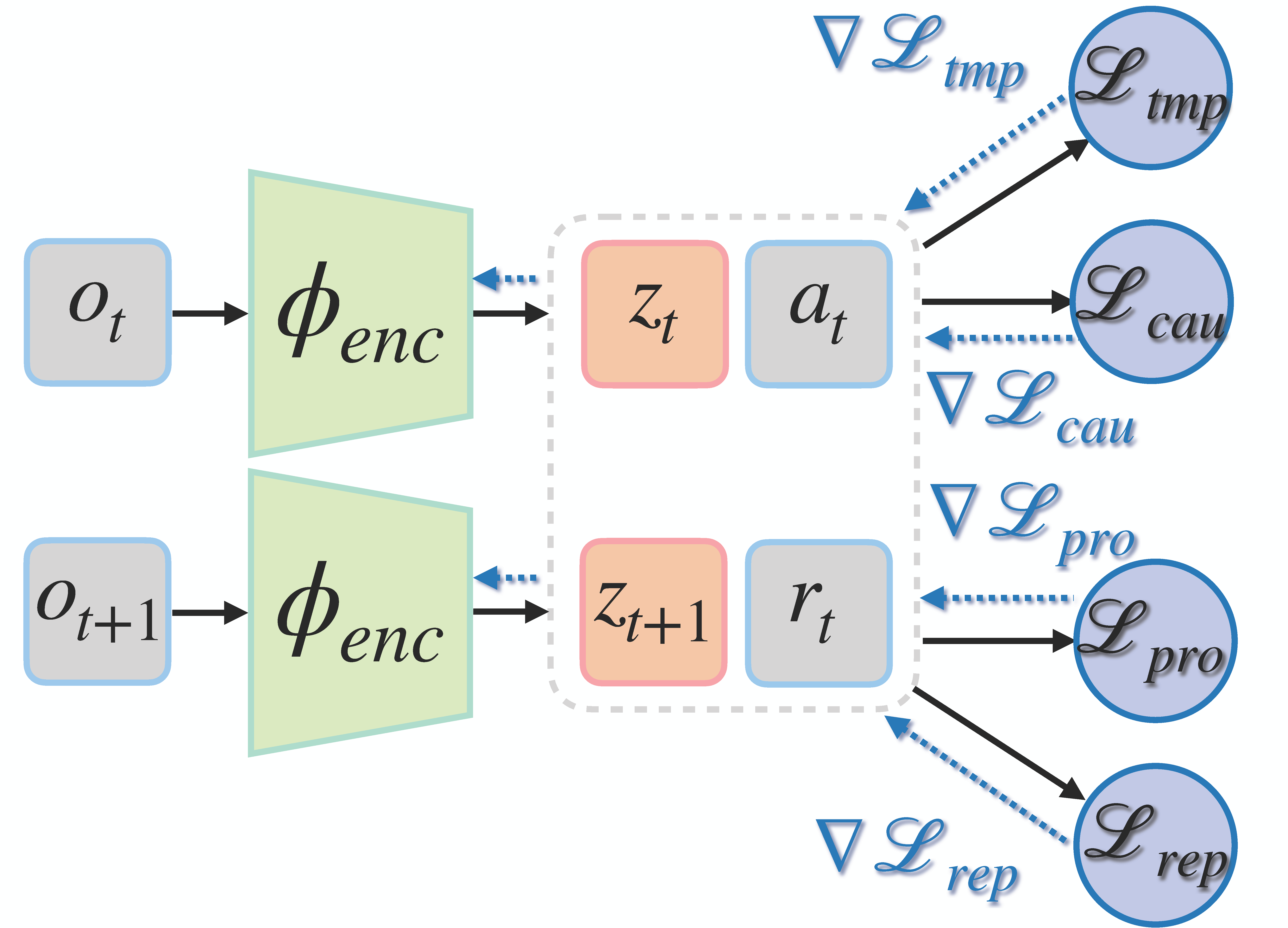}
    \caption{Robotics Priors. Different robotics priors can be used to train the encoder and shape the state representation. The solid line represents the forward pass of the data through the networks, while the dash line represents the gradient flow.}
    \label{fig:robotics_priors}
\end{figure}

These loss functions, called \textit{robotics priors} \cite{jonschkowski2014state}, do not rely on any auxiliary model, e.g. forward or reward model, and can be used to directly train the encoder, as shown in Figure \ref{fig:robotics_priors}.
Robotics priors share similarities with Physics Informed Neural Networks (PINNs) \cite{raissi2019physics} as both methods combine data and physical knowledge to improve the learning process. PINNs are a class of neural networks specialized in solving partial differential equations by jointly minimizing a data fidelity term (e.g. \acrshort{mse} between the network prediction and the measured value of the solution) and the residual of the differential equation.

An example of a robotics prior is the \textit{temporal coherence} loss introduced in \cite{jonschkowski2014state} and shown in Equation \eqref{eq:tempcohexample}.
\begin{equation}
\begin{split}
    \min_{\bm{\theta}_{\text{enc}}} &\ \ \ \ \mathscr{L}_{\text{tmp}}(\bm{\theta}_{\text{enc}}) \\
    \mathscr{L}_{\text{tmp}}(\bm{\theta}_{\text{enc}}) &= \mathbb{E}_{\bm{o},\bm{o}'\sim \mathcal{D}}[||\Delta \bm{z}_t||^2] \\
    &= \mathbb{E}_{\bm{o},\bm{o}'\sim \mathcal{D}}[||\bm{z}_{t+1}-\bm{z}_t||^2].
    \label{eq:tempcohexample}
\end{split}
\end{equation}
The temporal coherence loss enforces smooth and small changes in the learned state space between pairs of consecutive latent states. The prior knowledge enforced by this loss function is the assumption that the actual state space is smooth and continuous. This property of the state space is valid for environments obeying physical laws, e.g., in robotics. 

The temporal coherence prior alone is insufficient for learning an informative state representation. The global minimum of such loss function is reached when mapping each latent state to the zero vector. Therefore, to prevent the learning of the trivial mapping, the \textit{causality} loss is introduced. The causality loss is an example of a contrastive loss, and it is discussed in detail in Section \ref{subsec:contrastive_losses}. This prior, see Equation \eqref{eq:causexample}, utilizes pairs of latent states, collected at different time instants $t_1$ and $t_2$ when the same actions $\bm{a}_{t_1} = \bm{a}_{t_2}$ is taken from such states, but the rewards received  $r_{t_1}, r_{t_2}$ are different. The loss function is minimized when the two states predictions are pushed apart in the latent  space. In continuous space, we can replace the equality in the conditioning of Equation \eqref{eq:causexample} with an equality up to a small $\epsilon$, e.g. $| \bm{a}_{t_1} - \bm{a}_{t_2} | \leq \epsilon$.
\begin{equation}
\begin{split}
    \min_{\bm{\theta}_{\text{enc}}} &\ \ \ \ \mathscr{L}_{\text{cau}}(\bm{\theta}_{\text{enc}}) \\
    \mathscr{L}_{\text{cau}}(\bm{\theta}_{\text{enc}}) &=  \mathbb{E}_{\bm{o},\bm{a},r,\bm{o}'\sim \mathcal{D}}[e^{-||\bm{z}_{t_1}-\bm{z}_{t_2}||^2}|\bm{a}_{t_1}=\bm{a}_{t_2}, r_{t_1} \neq r_{t_2}]. 
    \label{eq:causexample}
\end{split}
\end{equation}

When the same actions $\bm{a}_{t_1}, \bm{a}_{t_2}$ are applied in different states, we want our state variations $\Delta \bm{z}_{t_1}=\bm{z}_{t_1+1}-\bm{z}_{t_1}$ and $\Delta \bm{z}_{t_2}=\bm{z}_{t_2+1}-\bm{z}_{t_2}$ to be similar in magnitude.  This prior is encoded by the \textit{proportionality} loss and shown in Equation \eqref{eq:propexample}.
\begin{equation}
\begin{split}
    \min_{\bm{\theta}_{\text{enc}}} &\ \ \ \ \mathscr{L}_{\text{pro}}(\bm{\theta}_{\text{enc}}) \\
    \mathscr{L}_{\text{pro}}(\bm{\theta}_{\text{enc}}) &=  \mathbb{E}_{\bm{o},\bm{a},\bm{o}'\sim \mathcal{D}}[(||\Delta \bm{z}_{t_1}||-||\Delta \bm{z}_{t_2}|)^2|\bm{a}_{t_1}=\bm{a}_{t_2}]. 
    \label{eq:propexample}
\end{split}
\end{equation}

Another example is the \textit{repeatability} loss, in Equation \eqref{eq:repeatexample}, enforcing that state changes produced by the same action to be equal in magnitude and direction.
\begin{equation}
\begin{split}
    \min_{\bm{\theta}_{\text{enc}}} &\ \ \ \ \mathscr{L}_{\text{rep}}(\bm{\theta}_{\text{enc}}) \\
    \mathscr{L}_{\text{rep}}(\bm{\theta}_{\text{enc}}) &=  \mathbb{E}_{\bm{o},\bm{a},r,\bm{o}'\sim \mathcal{D}}[e^{-||\bm{z}_{t_1}-\bm{z}_{t_2}||^2}||\Delta \bm{z}_{t_2} - \Delta \bm{z}_{t_1}||^2 |\bm{a}_{t_1}=\bm{a}_{t_2}].
    \label{eq:repeatexample}
\end{split}
\end{equation}

Eventually, to obtain an informative and coherent state representation, the encoder $\phi_{\text{enc}}$ is then usually trained by jointly minimizing the sum of multiple robotics prior losses \cite{jonschkowski2014state}, as shown in Equation \eqref{eq:prior_lossSRL}.
\begin{equation}
\begin{split}
 \min_{\bm{\theta}_{\text{enc}}} \ \ \ \ &\omega_{\text{t}}\mathscr{L}_{\text{tem}}(\bm{\theta}_{\text{enc}}) + \omega_{\text{c}}\mathscr{L}_{\text{cau}}(\bm{\theta}_{\text{enc}}) + \omega_{\text{p}}\mathscr{L}_{\text{pro}}(\bm{\theta}_{\text{enc}}) + \omega_{\text{r}}\mathscr{L}_{\text{rep}}(\bm{\theta}_{\text{enc}})\,,  
\end{split}
\label{eq:prior_lossSRL}
\end{equation}
where $\omega_{\text{t}}$, $\omega_{\text{c}}$, $\omega_{\text{p}}$, and $\omega_{\text{r}}$ are scaling factors for scaling the contribution of each loss.

While allowing efficient learning of state representations for simple robotics tasks, the robotics priors struggle in generalizing, e.g., in the presence of domain randomization, and robustness when the observations include visual distractors \cite{lesort2019deep}.

\subsubsection{Contrastive Learning}\label{subsec:contrastive_losses}

Contrastive learning (\acrshort{cl}) \cite{chopra2005learning} is an unsupervised approach (sometimes referenced to as self-supervised) for learning data representations by only relying on similarities and dissimilarities of data pairs. Contrastive learning has achieved impressive results in \acrshort{dl}, and in the most recent years, this paradigm has been used in unsupervised state representation learning more and more often. Nowadays, contrastive learning is a crucial element for sample-efficient \acrshort{drl} \cite{laskin2020curl}.

The critical ingredient of \acrshort{cl} is the contrastive loss. Contrastive losses act on pairs of latent variables (states in the context of \acrshort{drl}). The minimization of a contrastive loss pushes apart \textit{negative} pairs, i.e., data not belonging to the same class or non-consecutive observations in the context of \acrshort{drl}. Additionally, contrastive losses prevent the collapsing of the representations to trivial embeddings when no labelled data are available. However, contrastive losses must always be combined with another loss function to effectively improve \acrshort{drl} algorithms' performance (only pushing apart the latent states is not enough for efficient policy learning in complex problems). Many contrastive losses have been proposed in the \acrshort{dl} literature, but here we focus on the most widely used in \acrshort{drl}: (i) InfoNCE, (ii) hinge, and (iii) causality loss.

Contrastive\marginpar{\scriptsize InfoNCE Loss} Predictive Coding (\acrshort{cpc}) was introduced in \cite{oord2018representation} for unsupervised learning of representations of future observations given the past ones. \acrshort{cpc} has been employed in unsupervised learning  of visual representations, supervised learning \cite{chen2020simple}, data-efficient supervised learning \cite{henaff2020data}, and \acrshort{drl} for sample-efficient learning of state representations \cite{laskin2020curl, stooke2021decoupling, you2022integrating}. \acrshort{cpc} employs the so-called InfoNCE loss (Equation \eqref{eq:infoNCEloss}). Optimizing InfoNCE allows the neural networks to learn (future) representations from negative samples.

{\color{black}
Given a set $X$ of $N$ random samples containing one positive sample $\bm{x}_{t+k} \sim p(\bm{x}_{t+k}|\bm{c}_t)$ and $N-1$ negative samples from the proposal distribution $p(\bm{x}_{t+k})$, we optimize:
\begin{equation}
    \mathscr{L}_{\text{InfoNCE}} = -\mathbb{E}_{X}\Big[\log\frac{f_k(\bm{x}_{t+k},\bm{c}_t)}{\sum_{x_j\in X}f_k(\bm{x}_j,\bm{c}_t)}\Big]\,,
    \label{eq:infoNCEloss}
\end{equation}
where $\bm{c}_t=h(\bm{x}_t, \bm{x}_{t-1}, \cdots)$ is the context vector summarizing time information from the latent states and $h$ is a nonlinear transformation, $f_k(\bm{x}_{t+k},\bm{c}_t) \propto \frac{p(\bm{x}_{t+k}|\bm{c}_t)}{p(\bm{x}_{t+k})}$ is the density ratio preserving the mutual information between $\bm{x}_{t+k}$ and $\bm{c}_t$. In practice, $f_k(\bm{x}_{t+k},\bm{c}_t) = \exp(\bm{z}_{t+k}^T W_k \bm{c}_t)$ with $W_k$ a learnable linear transformation. The InfoNCE loss is categorical cross-entropy of classifying the positive samples correctly.

An adaptation of the InfoNCE loss for \acrshort{srl} is proposed by \cite{laskin2020curl}, where the positive samples and context vectors are obtained by encoding the same observations $\bm{o}$ but with different augmentation techniques (e.g., different cropping) applied to them. In contrast, the negative examples are a set of encoded randomly-sampled observations.}

Another\marginpar{\scriptsize Hinge Loss} example of a contrastive loss for unsupervised learning of representations is the hinge loss \cite{kipf2019contrastive, van2020plannable, botteghi2021low}.
\begin{equation}
    \mathscr{L}_{\text{hinge}}(\phi_{\text{enc}}) = \mathbb{E}_{\bm{o},\bm{o}^{\neg} \sim \mathcal{D}}[\max(0, \epsilon -  d(\bm{z},\bm{z}^{\neg}))]\, ,
\end{equation}
where $d(,)$ is a distance between two non-consecutive latent states $\bm{z}$ and $\bm{z}^{\neg}$ and $\epsilon$ is a scaling parameter. The operator $\max(\cdot,\cdot)$ prevents the distance from growing indefinitely. The minimization of  $\mathscr{L}_{\text{hinge}}$ pushes apart non-consecutive state pairs.

Another\marginpar{\scriptsize Causality Loss} example is the causality loss introduced by \cite{jonschkowski2014state} pushing apart pairs of latent states when the same action is applied, but the reward received is different:
\begin{equation}
\mathscr{L}_{\text{cau}} =  \mathbb{E}_{\bm{o},\bm{a},r,\bm{o}^{\neg} \sim \mathcal{D}}[e^{-||\bm{z}_{t_1}-\bm{z}_{t_2}||^2}|\bm{a}_{t_1}=\bm{a}_{t_2}, r_{t_1} \neq r_{t_2}]. 
\end{equation}
The causality loss can be generalized by removing the conditioning on action and rewards such that  the loss pushes apart non-consecutive latent states. This loss function is used in \cite{franccois2019combined}.
\begin{equation}
\mathscr{L}_{\text{cau}} =  \mathbb{E}_{\bm{o},\bm{o}^{\neg} \sim \mathcal{D}}[e^{-||\bm{z}_{t_1}-\bm{z}_{t_2}||^2}] 
\end{equation}

\subsection{Academic Example: Learning State Representation for a Pendulum from High-Dimensional Observations}\label{subsec:tutorial} 

In this section, we introduce an academic example to show (i) the state representations learned by 16 different \acrshort{srl} methods, introduced in the previous sections, and (ii) how the methods relate to the properties of the state representation introduced in Section \ref{subsec:propertieslearningstaterepresentation}. We consider one of the most popular baselines in control and \acrshort{srl}: the pendulum. 
\begin{figure}[h!]
    \centering
    \includegraphics[width=1.0\textwidth]{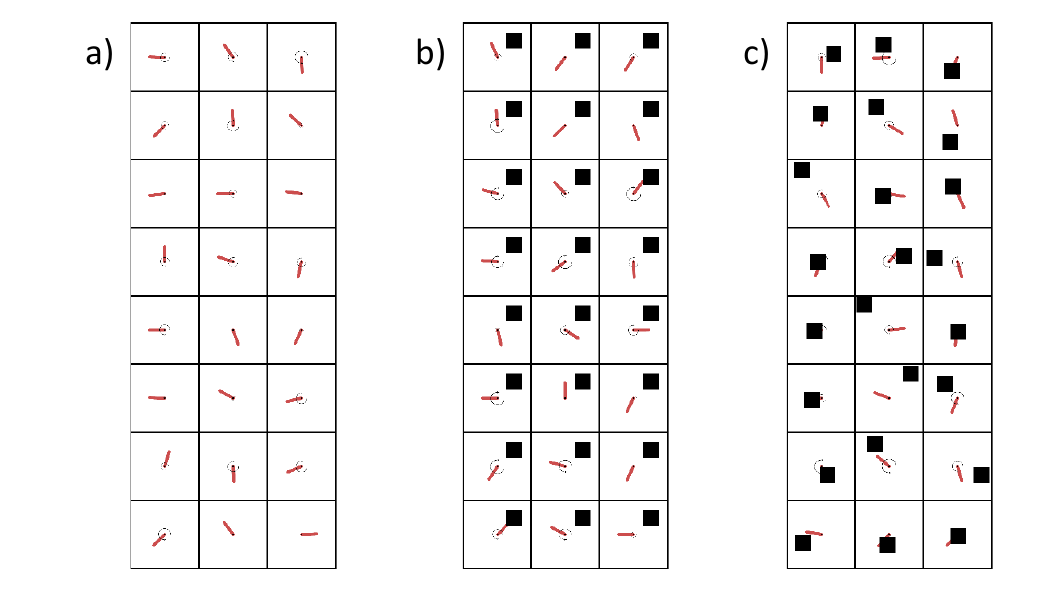}
    \caption{Different types of observations: (a) High-dimensional observations, (b) high-dimensional observations with distractors, i.e., black squares, and (c) high-dimensional observations with randomly-moving irrelevant features.}
    \label{fig:diff_obs}
\end{figure}
Despite its simplicity, the pendulum has a nonlinear dynamics and has an intrinsic low-dimensionality of its state variables, offering the possibility for a good interpretation of the results. Our experiments aim to learn a compact representation for a pendulum from high-dimensional observations (i.e. \acrshort{rgb} images of size $84\times84\times3$). Additionally, to make the task more challenging, we add irrelevant features (black square) to the inputs, that we called distractors, either fixed or randomly-moving. Examples of observations are shown in Figure \ref{fig:diff_obs}.  Their size and shape were empirically chosen to make the task sufficiently challenging while keeping the task of encoding the relevant features feasible. Distractors allow us to test the robustness of our methods by comparing whether the encoder can learn the relevant features for control or not. 
Details of the experiments are provided in \ref{sec:appendixA}, and the code is available at: \url{https://github.com/nicob15/State_Representation_Learning_Methods}.

\begin{figure}[h!]
    \centering
    \includegraphics[width=1.0\textwidth]{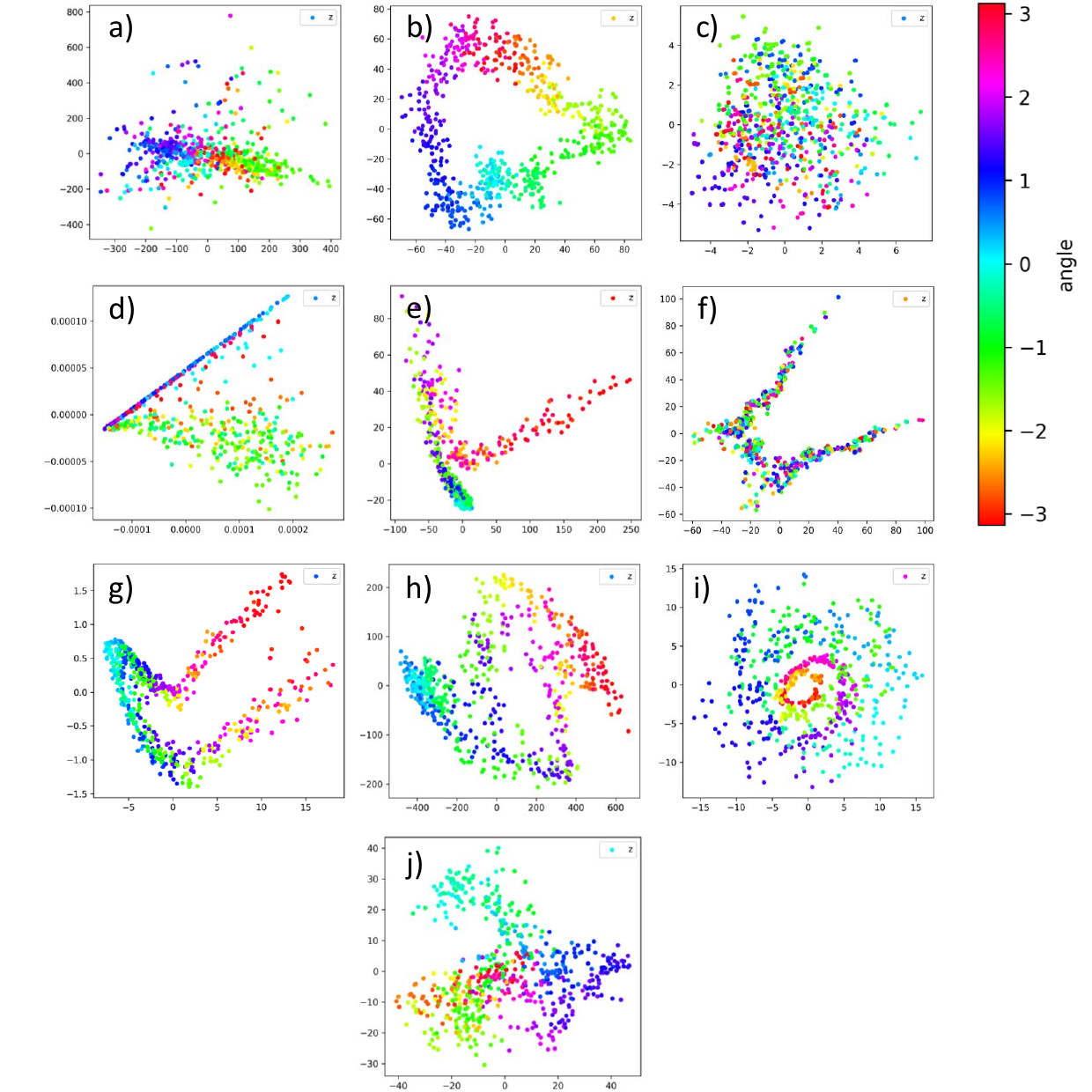}
        \caption{Learned state representations without distractors using the different methods: Principal Components Analysis (a), Autoencoder (b), Variational Autoencoder (c), Forward Model (d), Reward Model (e), Inverse Model (f), MDP Homomorphism (g), Bisimulation Metric (h), Priors (i), and Contrastive Learning (j). The colorbar represents the true angle of the pendulum. }
        \label{fig:learned_representations_no_dist}
\end{figure}

\begin{figure}[h!]
    \centering
    \includegraphics[width=1.0\textwidth]{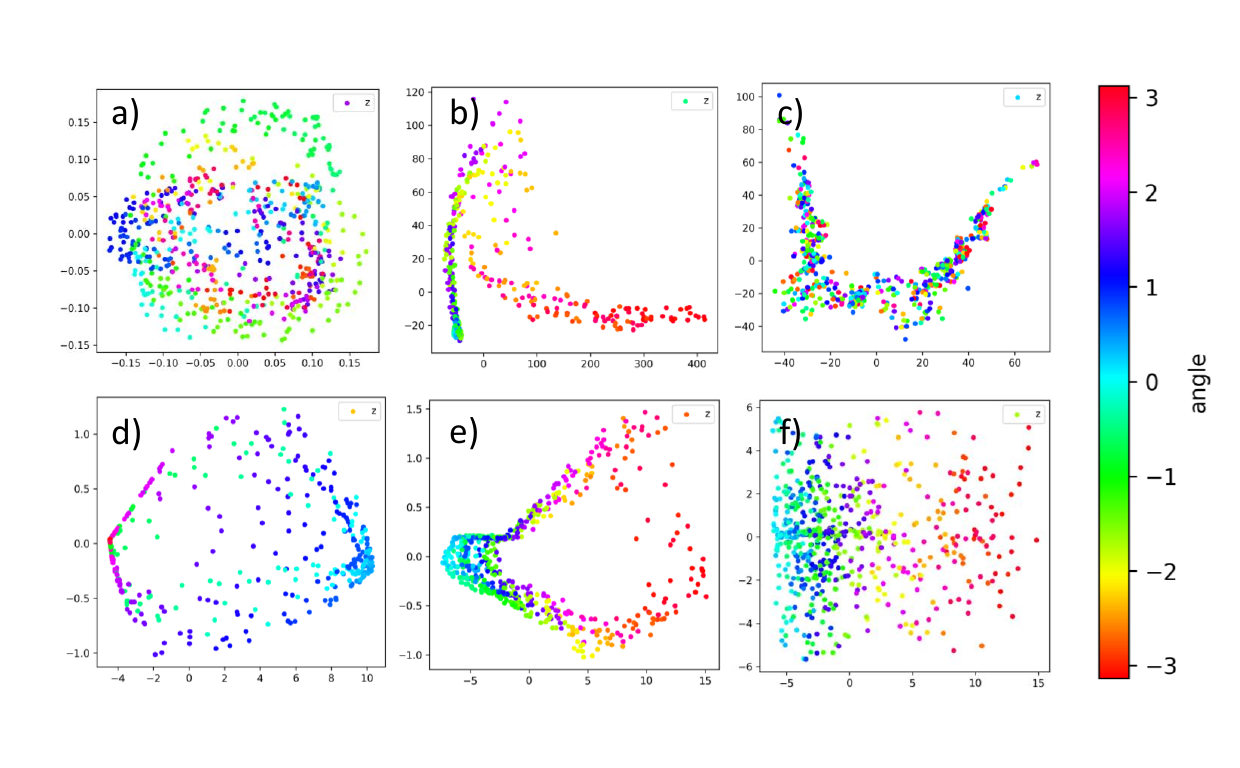}
        \caption{Learned state representations without distractors using the different methods: Autoencoder with Forward Model (a), Autoencoder with Reward Model (b), Autoencoder with Inverse Model (c), Forward Model with Contrastive Learning (d), Forward Model and Reward Model (e), Forward Model, Reward Model, and Inverse Model (f) The colorbar represents the true angle of the pendulum. }
       \label{fig:learned_representations_2_no_dist}
\end{figure}

\begin{figure}[h!]
    \centering
    \includegraphics[width=1.0\textwidth]{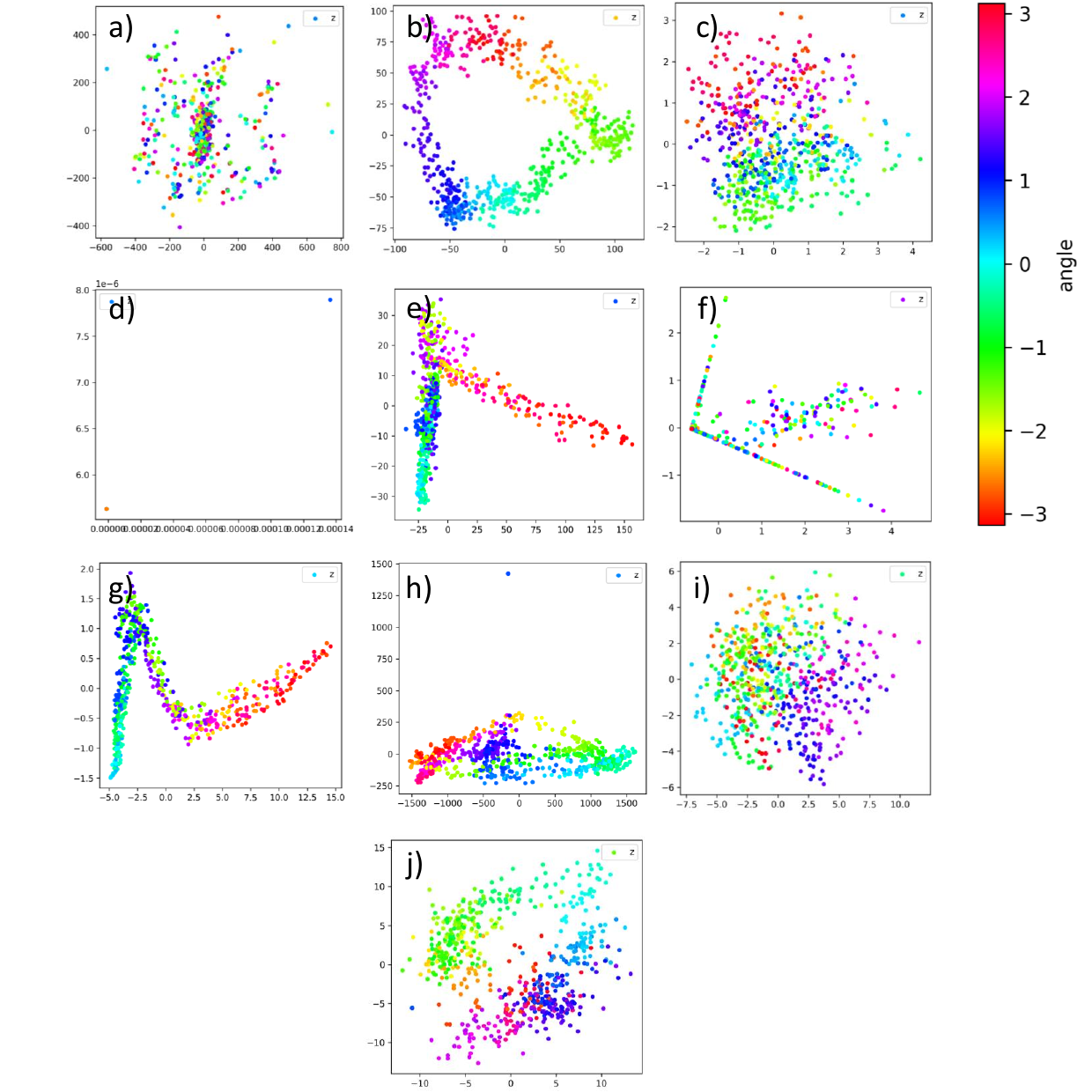}
        \caption{Learned state representations with fixed distractors using the different methods: Principal Components Analysis (a), Autoencoder (b), Variational Autoencoder (c), Forward Model (d), Reward Model (e), Inverse Model (f), MDP Homomorphism (g), Bisimulation Metric (h), Priors (i), and Contrastive Learning (j). The colorbar represents the true angle of the pendulum. }
        \label{fig:learned_representations_fixed_dist}
\end{figure}

\begin{figure}[h!]
    \centering
    \includegraphics[width=1.0\textwidth]{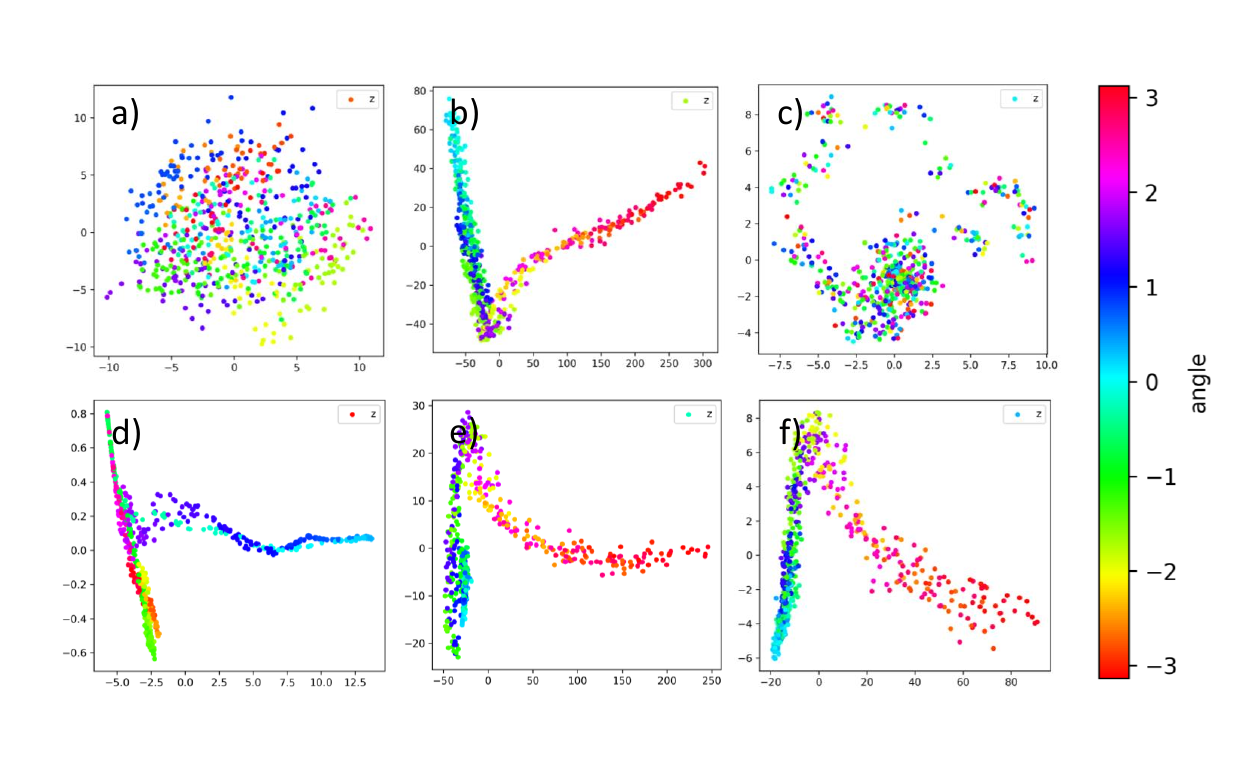}
        \caption{Learned state representations with fixed distractors using the different methods: Autoencoder with Forward Model (a), Autoencoder with Reward Model (b), Autoencoder with Inverse Model (c), Forward Model with Contrastive Learning (d), Forward Model and Reward Model (e), Forward Model, Reward Model, and Inverse Model (f) The colorbar represents the true angle of the pendulum. }
       \label{fig:learned_representations_2_fixed_dist}
\end{figure}

\begin{figure}[h!]
    \centering
    \includegraphics[width=1.0\textwidth]{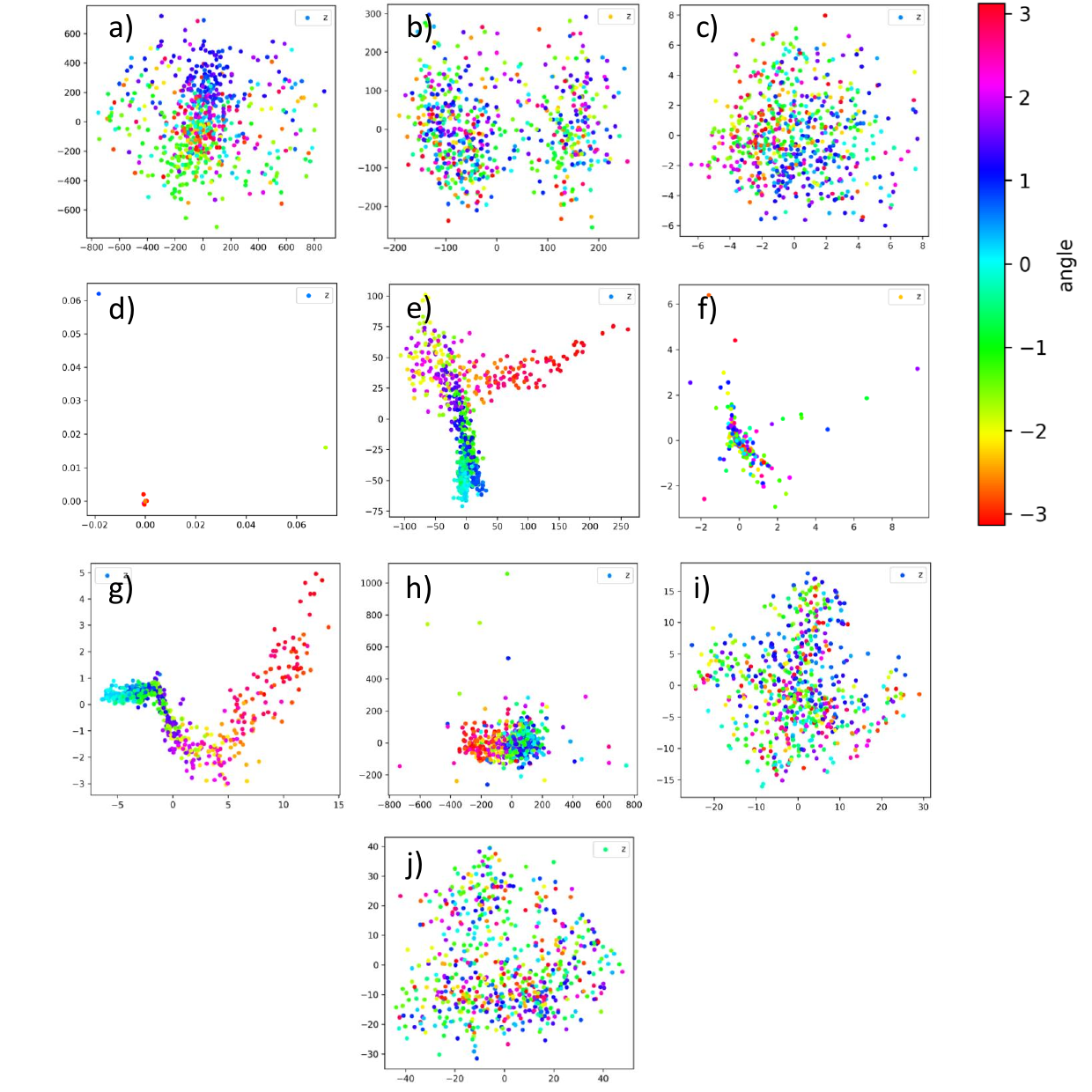}
        \caption{Learned state representations with randomly-positioned distractors using the different methods: Principal Components Analysis (a), Autoencoder (b), Variational Autoencoder (c), Forward Model (d), Reward Model (e), Inverse Model (f), MDP Homomorphism (g), Bisimulation Metric (h), Priors (i), and Contrastive Learning (j). The colorbar represents the true angle of the pendulum. }
        \label{fig:learned_representations_random_dist}
\end{figure}

\begin{figure}[h!]
    \centering
    \includegraphics[width=1.0\textwidth]{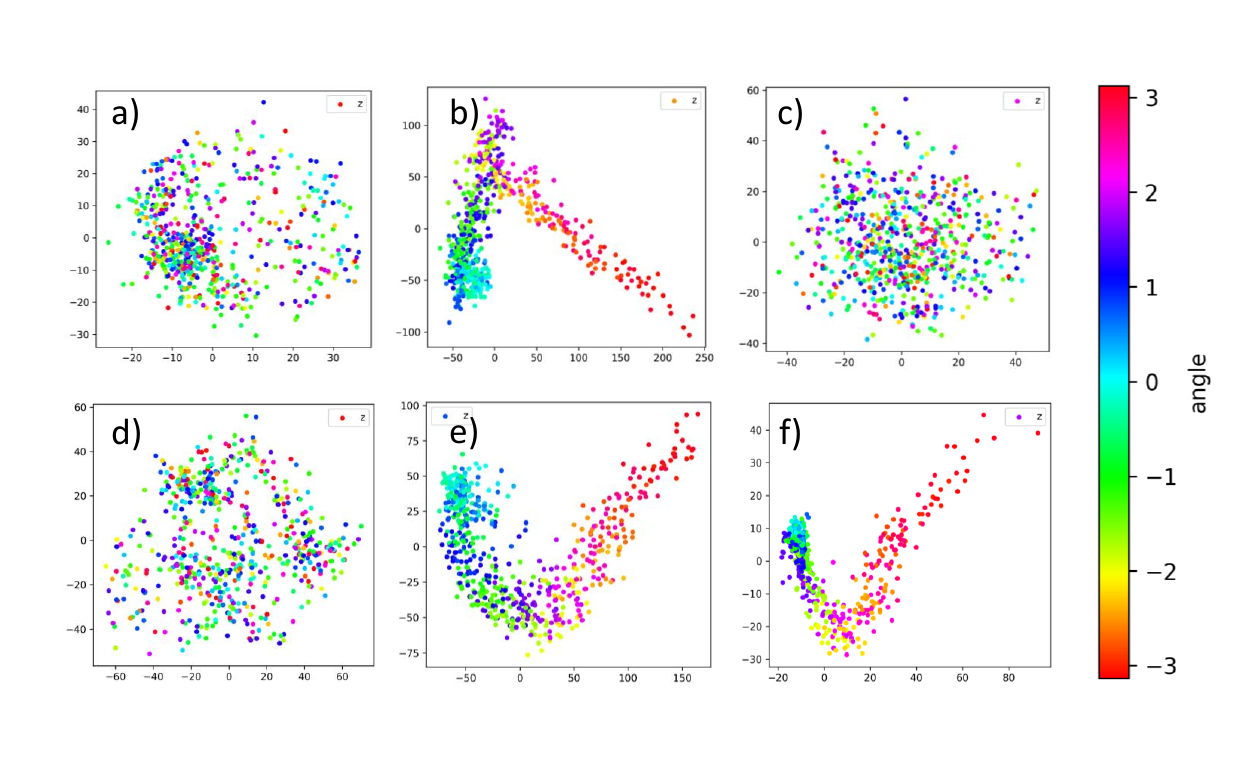}
        \caption{Learned state representations with randomly-positioned distractors using the different methods: Autoencoder with Forward Model (a), Autoencoder with Reward Model (b), Autoencoder with Inverse Model (c), Forward Model with Contrastive Learning (d), Forward Model and Reward Model (e), Forward Model, Reward Model, and Inverse Model (f) The colorbar represents the true angle of the pendulum. }
       \label{fig:learned_representations_2_random_dist}
\end{figure}

 We show the learned state representations in the case of distractor-free observations in Figure \ref{fig:learned_representations_no_dist} and \ref{fig:learned_representations_2_no_dist}, the representations in the case of fixed distractor in Figure \ref{fig:learned_representations_fixed_dist} and \ref{fig:learned_representations_2_fixed_dist}, and  the representations in presence of randomly-moving distractors in Figure \ref{fig:learned_representations_random_dist} and \ref{fig:learned_representations_2_random_dist}. The color bar represents the true angle of the pendulum and it is used for assessing the smoothness of the latent state space. A smooth representation should be able to encode similar observations, i.e. observations captured with similar angles of the pendulum, close together in the latent space. In our experiments, we set the latent state space dimension to 50 for all the methods. Thus, to visualize and plot the results, we use \acrshort{pca} to project the latent state variables to a 2-dimensional space linearly. This procedure is commonly done in the \acrshort{srl} literature. The interpretability of such figures is strongly related to the ability of the method to compress the measurements, i.e., the ability to reduce dimensionality, extract the relevant features, i.e., smoothness and expressiveness, and disentangle the factors of variation, i.e., complexity of the dependencies.

In Table \ref{tab:method-property}, we relate the methods introduced in the previous sections to the properties of the learned state space (see Section \ref{subsec:propertieslearningstaterepresentation}). While the table presents a qualitative analysis of the pendulum example, we believe that most properties and results can be generalized and extended to more advanced applications.
\begin{table*}[h!]
    \caption{Relation between the \acrshort{srl} methods and the desired properties of the latent space $\mathcal{Z}$ (described in Section \ref{subsec:propertieslearningstaterepresentation}) in the pendulum example. Each method is graded with yes/no (\faCheck/\faTimes), low (\faPlus), medium (\faPlus \faPlus), or high (\faPlus \faPlus \faPlus) based on a qualitative analysis of the results obtained, i.e., the quality of the representations with the three types of observations, and the results from the literature.}
    \label{tab:method-property}
    \centering
    \tiny
    \begin{tabular}{| l | p{0.1\linewidth}|p{0.1\linewidth}|p{0.1\linewidth}|p{0.07\linewidth}|p{0.07\linewidth}|p{0.1\linewidth}|}
     \hline
    \hline
     Method ($\downarrow$) - Property ($\rightarrow$)    &   Latent state space smoothness &   Ability to reduce \hbox{dimensionality} &   Complexity of the \hbox{dependencies} &   Markovian &   Temporal \hbox{coherence} &   \hbox{Expressiveness}  \\
     \hline
    \hline
     Principal Component Analysis  &   \faPlus\faPlus &   \faPlus &   \faPlus & \faTimes & \faTimes &  \faPlus \\
    \hline
     Autoencoders  &   \faPlus\faPlus &   \faPlus\faPlus & \faPlus\faPlus & \faTimes & \faTimes &   \faPlus\faPlus \\
    \hline
      Variational Autoencoders &   \faPlus\faPlus &   \faPlus\faPlus &  \faPlus\faPlus & \faTimes & \faTimes &  \faPlus\faPlus \\
    \hline 
      Forward Model  &   \faPlus &   \faPlus\faPlus\faPlus &  \faPlus & \faTimes & \faCheck &   \faPlus \\
    \hline
      Reward Model &   \faPlus\faPlus\faPlus &   \faPlus\faPlus\faPlus & \faPlus\faPlus & \faTimes & \faTimes &   \faPlus\faPlus \\
    \hline
      Inverse Model  &   \faPlus &   \faPlus\faPlus\faPlus & \faPlus & \faTimes & \faTimes &   \faPlus \\
    \hline
      MDP Homomorphism \cite{van2020plannable}  &  \faPlus\faPlus\faPlus &   \faPlus\faPlus\faPlus &   \faPlus\faPlus\faPlus &  \faCheck &   \faCheck &  \faPlus\faPlus\faPlus \\
    \hline
      Bisimulation Metric  \cite{Zhang2020} &   \faPlus\faPlus\faPlus &   \faPlus\faPlus\faPlus &   \faPlus\faPlus\faPlus &   \faCheck &   \faCheck &   \faPlus\faPlus\faPlus \\
    \hline
      Priors \cite{botteghi2021prior}  &   \faPlus\faPlus &   \faPlus\faPlus\faPlus &   \faPlus\faPlus & \faTimes &  \faCheck &   \faPlus\faPlus \\
    \hline
      Contrastive Learning  &   \faPlus\faPlus &  \faPlus\faPlus & \faPlus &  \faTimes &  \faCheck  &   \faPlus \\
    \hline
    \hline
    Autoencoders with Forward Model & \faPlus\faPlus &  \faPlus\faPlus  & \faPlus\faPlus &  \faTimes & \faCheck & \faPlus\faPlus    \\
    \hline
    Autoencoders with Reward Model  & \faPlus\faPlus\faPlus  & \faPlus\faPlus   & \faPlus\faPlus & \faTimes & \faTimes  & \faPlus\faPlus\faPlus    \\
    \hline
    Autoencoders with Inverse Model   & \faPlus &  \faPlus\faPlus   & \faPlus\faPlus & \faTimes &  \faTimes &  \faPlus\faPlus  \\
    \hline
    Forward Model with Contrastive Learning & \faPlus\faPlus\faPlus & \faPlus\faPlus\faPlus  & \faPlus\faPlus\faPlus &  \faCheck &   \faCheck &   \faPlus\faPlus\faPlus \\
    \hline
    Forward and Reward Model  & \faPlus\faPlus\faPlus & \faPlus\faPlus\faPlus & \faPlus\faPlus\faPlus  &  \faCheck &   \faCheck &   \faPlus\faPlus\faPlus \\
    \hline
    Forward, Reward, and Inverse Model  & \faPlus\faPlus\faPlus & \faPlus\faPlus\faPlus & \faPlus\faPlus\faPlus  &  \faCheck &   \faCheck &   \faPlus\faPlus\faPlus \\
    \hline
    \hline
    \end{tabular}
\end{table*}

Eventually, we quantitatively evaluate the quality of the state representations learned using the different \acrshort{srl} methods. In particular, after the training of the different \acrshort{srl} encoders, we train an additional model, that we call state decoder, to reconstruct the true states of the pendulum from the latent states. After the training of the state decoder, we record the reconstruction error over the test set. The additional model is a $\acrshort{fcnn}$ with input dimension equal to the latent state dimension and output dimension equal to the true state dimension. Details can be found in  \ref{sec:appendixA}. In Table \ref{tab:quantitative-eval-methods}, we show the reconstruction errors using a linear and a nonlinear decoder for the three different observation types, namely the case with no distractors (Case 1), the case with fixed distractors (Case 2), and the case with randomly-moving distractors (Case 3).
\begin{table*}[h!]
    \caption{Quantitative evaluation of the \acrshort{srl} methods in the pendulum example.  We compare the different methods for the three types of observation groups: no distractors (Case 1), fixed distractors (Case 2), and randomly-moving distractors (Case 3). We train, on the training set, a linear and a nonlinear decoder to reconstruct the true states from the latent states, learned by the different methods. We report the mean square error of the reconstructions on the test set of the linear (left) and nonlinear (right) decoders.}
    \label{tab:quantitative-eval-methods}
    \centering
    \tiny
    \begin{tabular}{| l | p{0.1\linewidth}|p{0.1\linewidth}|p{0.1\linewidth}|}
     \hline
    \hline
     Method ($\downarrow$)  &   Case 1 &   Case 2  & Case 3  \\
     \hline
    \hline
     Principal Component Analysis  & 8.127 - 0.715   & 7.879 - 0.556   & 9.679 - 9.711   \\
    \hline
     Autoencoders  & 6.932 - 0.207    & 7.141 - 0.184    & 7.808 -  2.893  \\
    \hline
      Variational Autoencoders & 6.810 - 0.189 & 6.866 -  0.207   & 6.947 - 3.455  \\
    \hline 
      Forward Model  & 7.211- 7.211   & 7.211 -  7.211   & 7.211 - 7.211   \\
    \hline
      Reward Model & 6.181 - 2.026   & 6.369 - 3.229    &  6.640 - 3.570  \\
    \hline
      Inverse Model  &  7.163 - 7.123  & 7.210 - 7.211   & 7.190 - 7.182   \\
    \hline
      MDP Homomorphism \cite{van2020plannable}  & 6.455 - 6.252   & 7.145 - 7.078   & 7.026 - 7.020  \\
    \hline
      Bisimulation Metric  \cite{Zhang2020} & 3.045 - 0.400  & 1.945 - 0.515    &  7.178 - 7.362 \\
    \hline
      Priors \cite{botteghi2021prior}  & 6.970 - 6.929   & 7.063 - 7.078   & 7.192 - 7.179  \\
    \hline
      Contrastive Learning  & 6.595 - 0.733 & 6.848 - 3.847  & 7.126 - 7.128  \\
    \hline
    \hline
    Autoencoders with Forward Model & 4.263 - 3.038   & 6.864 - 0.163    & 6.978 -  2.705 \\
    \hline
    Autoencoders with Reward Model  & 6.246 -  0.554   & 6.613 - 0.547    & 6.985 -  3.650  \\
    \hline
    Autoencoders with Inverse Model   & 6.368 - 0.458    & 6.915 - 4.039   & 6.839 - 2.771   \\
    \hline
    Forward Model with Contrastive Learning & 6.400 - 6.374    &  7.063 - 7.024   & 7.087 - 6.983   \\
    \hline
    Forward and Reward Model  & 1.260 - 0.609    & 6.047 - 3.341    & 6.514 - 3.236 \\
    \hline
    Forward, Reward, and Inverse Model  & 6.979 - 6.948 & 6.601 - 4.996    & 6.630 - 4.865  \\
    \hline
    \hline
    \end{tabular}
\end{table*}

\subsection{Summary}\label{sec3:discussion}

To conclude, we aim to summarize the results of the qualitative and the quantitative analysis and highlight the key strength and weaknesses of each category of methods. From the qualitative analysis (Figure \ref{fig:learned_representations_no_dist}-\ref{fig:learned_representations_2_random_dist} and Table \ref{tab:method-property}), we can conclude that:
\begin{itemize}
\item Observation reconstruction-based methods tend to focus on encoding features useful for minimizing the reconstruction loss, e.g., backgrounds or distractors (see Figure \ref{fig:learned_representations_random_dist} and \ref{fig:learned_representations_2_random_dist}). However, this class of methods tends to have a more stable training procedure and learn representations that hardly collapse.
\item Approaches relying on \acrshort{mdp} models naturally encode environment dynamics. However, they tend to be harder to train, and they often need to be coupled with contrastive losses, e.g., in the case of the forward model, to prevent representation collapse.
\item Priors allow higher sample efficiency than other methods but tend to suffer from visual distractors and non-stationary environments.
\item The reward model is crucial for smoothness, temporal coherence, and expressivity of the methods when randomly-positioned distractors are present.  The reward model is a good discriminator of task-relevant and task-irrelevant information, especially in the case of a nonsparse reward function, as in the considered pendulum example. With very sparse reward functions, the ability of the reward model to separate features is drastically reduced, and, in this context, discriminating task-relevant and task-irrelevant information is an open challenge in the field. Additional mechanisms for discriminating between latent states have to be used often, e.g., value function estimation.
\item Contrastive losses need to be combined with other losses, e.g., forward model loss \cite{van2020plannable} or temporal-difference loss \cite{laskin2020curl}, to improve sample efficiency and prevent representation collapse.
\end{itemize}
From the quantitative analysis (see Table \ref{tab:quantitative-eval-methods}), we can conclude that:
\begin{itemize}
    \item Independently of the method and observation type used, linearly reconstructing the true states from the latent states is not possible with good accuracy in many cases (as indicated by reconstruction errors around 7). This can be seen by the higher state-reconstruction loss of the linear decoder (left) compared to the state-reconstruction loss of the nonlinear decoder (right).
    \item \acrshort{ae}-based architectures, with the exception of the linear \acrshort{pca}, and the reward model tend to improve the reconstruction of the nonlinear state decoders. Reconstruction loss and reward model loss (for the case treated case of a dense reward function) allow good discrimination of the latent states. Conversely, adding a forward model tend to deteriorate the reconstruction capabilities of the state decoder. This is due to the fact that the forward model loss is zero when the distance between two consecutive latent states is zero (see Equation \eqref{eq:forward_model_loss}).
    \item Combinations of \acrshort{ae} with latent models tend to achieve the best overall performance in the thee treated scenarios. 
    \item Because MDP homomorphism and priors-based methods aim to learn the most compact state representation by exploiting the symmetries of the problem, they naturally make the task of reconstructing the true states more challenging. This aspect can be noticed, for example, from Figure \ref{fig:learned_representations_no_dist} g) and \ref{fig:learned_representations_no_dist} i), where latent states obtained by encoding different measurements by with same transition and reward function are mapped close to each other in the latent space. While these methods make it difficult to recover the true state variables from the latent state variables, it is worth to keep in mind that the estimation of the value function may become simpler for such state representations due to the exploitation of the problem symmetries. 
    \item The bisimulation metric is the only method that achieves good linear and nonlinear reconstruction error in the case of no distractors and in the case of fixed distractors. This method shows promising results and the potential for further research in the case of randomly-moving distractors.
\end{itemize}

%% file: StateRepresentationLearning/4-UnsupervisedRepresentationLearning.tex
When defining a taxonomy diagram for \acrshort{drl} methods, one of the first distinctions we have to address is how the samples collected via the interaction with the environment are used. Usually, we have two options: \textit{model-free} and \textit{model-based} methods. Although we are aware of recent approaches combining model-free and model-based methods and that some of the model-free \acrshort{srl} methods presented here could be adapted to work in model-based settings and vice versa, for the sake of simplicity, we use this explicit dichotomy in our review.
In model-free algorithms, the samples are used to directly estimate the value function and policy, e.g. via temporal-difference learning or policy gradient \cite{Sutton1998}. In contrast, in model-based algorithms, the samples are used to build environment models, namely a transition model or transition and rewards model, that are then exploited to estimate the value function and policy analogously to model-free \acrshort{drl}.

\begin{figure}[h!]
    \centering
        \includegraphics[width=1.0\textwidth]{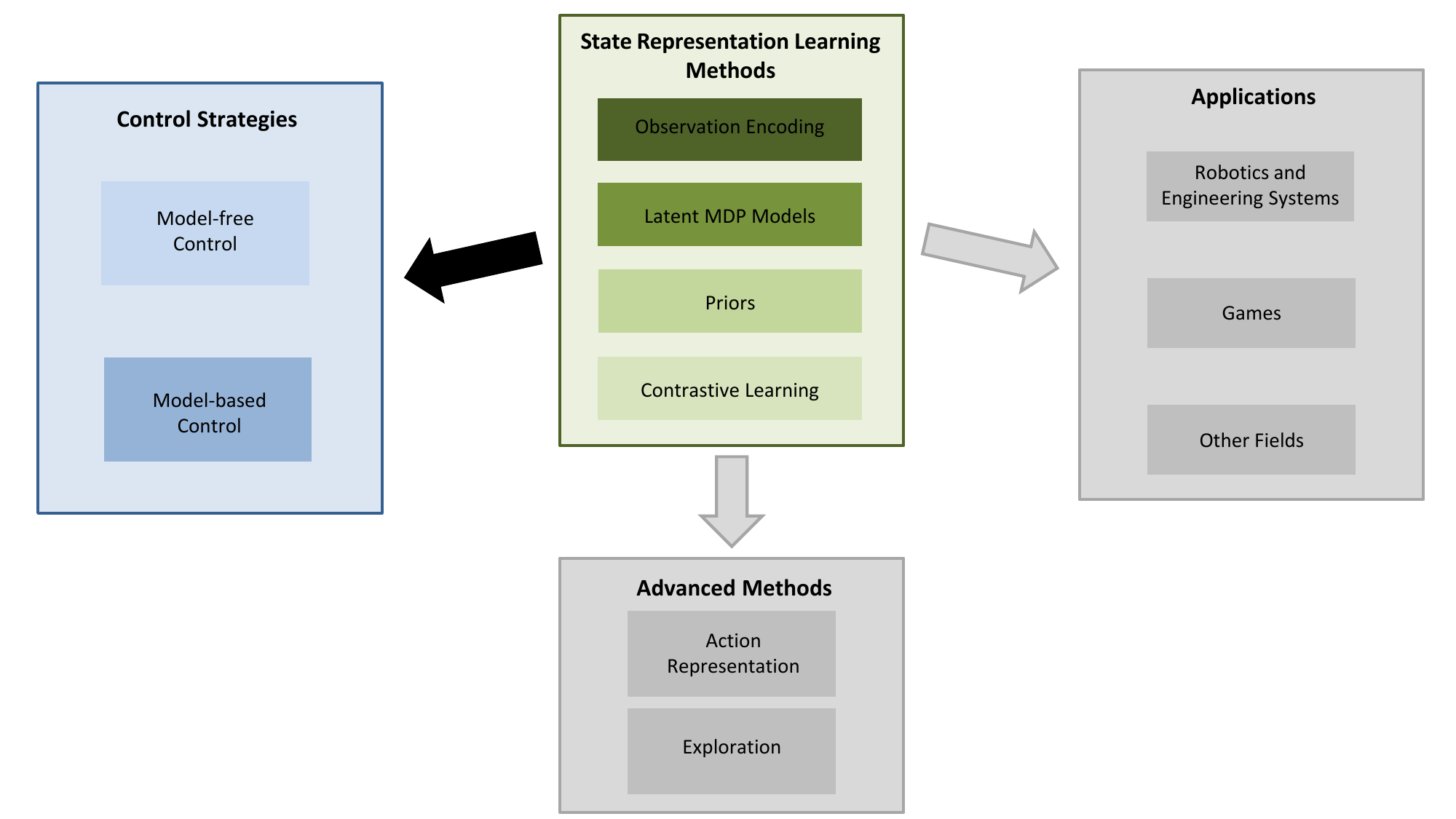}
        \caption{Overview of State Representation Learning methods in Deep Reinforcement Learning.}
        \label{fig:bigpicture2}
\end{figure}

For structuring our review (see Figure \ref{fig:bigpicture2}), we utilize these two classes of \acrshort{drl} methods, namely model-free and model-based, to define two macro objectives of the \acrshort{srl} approaches:
\begin{enumerate}
    \item Increasing the sample efficiency, robustness, and generalization abilities of model-free \acrshort{drl} by learning meaningful low-dimensional state representations.
    \item Improving prediction accuracy, robustness, and generalization abilities of learned latent \acrshort{mdp} models and consequently the quality of the policies and value functions learned in model-based \acrshort{drl}.
\end{enumerate}
Given these two objectives, we review how the approaches presented in Section \ref{sec:OverviewMethods} are effectively used for improving end-to-end \acrshort{drl} algorithms.

\subsection{State Representation for Sample Efficiency in Model-free RL}\label{subsec:SRL_modelfree}

This section discusses the problem of learning state representations for model-free RL. Such state representations are used by the agent to learn the value function and the policy by interacting with the environment.

\subsubsection{Learning State Representations with Principal Component Analysis}

One\marginpar{\scriptsize PCA} of the first applications of \acrshort{pca} in \acrshort{drl} was proposed by \cite{curran2015using}, where a low-dimensional state representation was learned from raw grey-scale images. Using the state representation, the  \acrshort{drl} agent was efficiently trained to solve a single level of the game Super Mario. 
Another example of the use of \acrshort{pca} for learning low-dimensional state representations was the work of \cite{parisi2017goal}, where the return-weighted \acrshort{pca}  was used to learn a state representation of a continuous grid world and the simulated robot tetherball game. The use of the return of the episodes allows to explicitly maintain information relevant to the \acrshort{drl} task.

However, in complex problems, the linear mapping learned by \acrshort{pca} from observations to states is insufficient for encoding all the relevant information, and, commonly, nonlinear and more expressive mappings are learned using for example \acrshort{ae}s (described in Section \ref{sec:OverviewMethods}).

\subsubsection{Learning State Representations with Autoencoders}

One of the first uses of \acrshort{ae}s\marginpar{\scriptsize Autoencoder} in \acrshort{drl} was introduced in \cite{lange2010deep}, where an \acrshort{ae} is used to learn a 2D representation of a grid-world. The actual state of the environment was not observable, and the agent could only use grey-scale image observations corrupted by noise. The \acrshort{ae} not only reduced the dimensionality of the observations to a 2-dimensional state vector allowing efficient \acrshort{drl}, but also acted as a denoiser of the images. Compressing the observations naturally helps removing the noise on the data. A similar example is the work of \cite{mattner2012learn}, where an \acrshort{ae} was used to learn a compact state representation from grey-scale images of a real-world inverted-pendulum in the presence of time delays. The \acrshort{drl} agent could quickly learn the optimal policy given the state representation, the previous action taken, and the delay. The last example of \acrshort{ae} for \acrshort{srl} was shown by \cite{Alvernaz2017} where the \acrshort{ae} was used for learning a compressed state representation of the game VizDoom \cite{kempka2016vizdoom}.

When\marginpar{\scriptsize Variational \hbox{Autoencoder}} uncertainties have to be considered due to the presence of noise on the data and when better generalization and regularization of the latent state space is required, e.g. when studying the problem of adaptation of the policies to different environments \cite{Higgins2017, caselles2018continual, Nair2018}, \acrshort{ae}s are often replaced by the \acrshort{vae}s (see Section \ref{sec:OverviewMethods}). In particular, a \acrshort{vae}-based state representation is used in the work of \cite{caselles2018continual} to tackle the problem of continual learning in continuously changing environments where the regularization of the state space is critical for being able to generalize to new scenarios. When even better generalization skills are required, for example, in the case of zero-shot adaptation (i.e. adaptation without further training on the new environments),  \acrshort{vae}s can be replaced by $\beta$-\acrshort{vae}. This is the case in the work of \cite{Higgins2017} and in \cite{Nair2018}, where a goal-based state representation was learned to generalize to new goal locations in navigation tasks.

Despite these successes, \acrshort{ae}s and \acrshort{vae}s have two major limitations when used in \acrshort{drl}: (i) the reconstruction loss is often independent of the relevant features for control, and (ii) only the latent states are used in \acrshort{drl}, making the decoder an unnecessary element. In the following sections, we discuss how to tackle these two challenges. 

\subsubsection{Learning State Representations with Autoencoders and Auxiliary Objectives}

The lack of discrimination between relevant and irrelevant features for control due to the reconstruction loss translates into the lack of coherence and smoothness of the latent state space. For example, two consecutive observations that should be encoded close together in the latent space may be encoded by an \acrshort{ae} far from each other if they present very different features. However, for example, from one timestep to the other, the position of a robot will not change too much due to the continuity of the world. We can use additional objectives to the reconstruction loss to enforce the learning of more informative state representations.

The first example of an auxiliary loss for an \acrshort{ae} is the Deep Spatial \acrshort{ae} introduced by \cite{finn2016deep}. Deep Spatial \acrshort{ae} was used for discovering compact state representations from \acrshort{rgb} images in the task of learning dexterous manipulation using a robot arm. To improve the quality of the representation, the authors proposed a regularization loss for keeping latent states from consecutive observations close to each other in the latent space. Similarly, a regularized \acrshort{ae} is used by \cite{yarats2021improving}. A nonlinear forward model in combination with an \acrshort{ae} is used by \cite{kim2021acceleration}.  Reward and inverse models were used by \cite{raffin2019decoupling} for improving the quality of \acrshort{ae}-based state representations and for allowing transfer learning from simulation to real-world (sim2real) in a  mobile robot navigation task. Sim2real is extremely challenging due to the differences between the simulation platform and the real world. However, the enhanced robustness of the features learned by the \acrshort{srl} methods makes it possible \cite{raffin2019decoupling, jonschkowski2014state, botteghi2021rewpriors, botteghi2021low}. The problem of imitation learning from third-person to first-person views was studied by \cite{shang2021self}. A pair of \acrshort{ae}s encoding third-person observations and respective first-person observations are regularized for disentangling view information from state space information. In particular, they used a permutation loss for matching third-person views to first-person views. 

Analogously to the \acrshort{ae} regularization using auxiliary objectives, we can do the same for \acrshort{vae}s. A \acrshort{vae} regularized by a linear latent forward model was employed in \cite{van2016stable} for learning low-dimensional representations of a pendulum and a robot arm for tactile manipulation. A stochastic forward model was instead used in \cite{lee2020stochastic}, while\acrshort{vae}s with latent forward and reward models were used by \cite{rafailov2021offline, tennenholtz2021gelato}. The latter used the forward model and the decoder to infer an approximate metric on the learned (Riemannian) manifold. The metric was used to penalize the uncertainties in latent states and latent dynamics. This penalization was added as an additional term in the reward functions to improve the training efficiency of the agent. The authors of \cite{caselles2019symmetry} enhanced a \acrshort{vae} with a latent transition model and its corresponding loss. Unlike other approaches, next to the state representation, the authors introduce an action representation network, embedding the agent's actions in a latent action space. In particular, each action is represented by a rotational matrix with four learnable parameters. This type of action embedding is effective when the actions exactly correspond to transformations (e.g. rotations, translations, roto-translations) that can be represented by matrices.

\subsubsection{Learning State Representations with MDP models}

In \acrshort{drl}, we are interested in learning task-relevant and compressed representations of the observations. The latent representations are used for learning the value functions and the policies. The observation reconstructions are rarely used in practice except for analyzing the results and for interpretability. In this section, we review approaches that do not rely on a decoder, i.e., the observation-reconstruction loss, but only on auxiliary losses derived from the latent \acrshort{mdp} models (see Section \ref{sec:OverviewMethods}).

The \marginpar{\scriptsize MDP \hbox{Homomorphism}} first important aspect is the \acrshort{mdp} homomorphism. The \acrshort{mdp} homomorphism allows us to define a metric to train the encoder without the need for the reconstruction loss, and  with the guarantee of a Markovian latent space when the loss functions approach zero. In particular, the goal is to learn a latent \acrshort{mdp} in the latent state variables, which is a homomorphic image of the original \acrshort{mdp}, i.e. with equivalent transition and reward function, but with lower dimensionality and easier to solve. In this category of approaches, we find the work of \cite{gelada2019deepmdp} using the \acrshort{mdp} homomorphism metric as a regularizer for a (distributional) Q-learning agent \cite{bellemare2017distributional}. This framework allowed the author to study the optimality of the policy and provide error bounds. In most Atari games, the approach outperformed different baselines, such as end-to-end \acrshort{drl} and \acrshort{ae}-based approaches. 

A \marginpar{\scriptsize Bisimulation \hbox{Metric}} closely-related concept to the \acrshort{mdp} homomorphism is the so-called bisimulation metric. Here we find the works of \cite{castro2020scalable, biza2019online, kemertas2021towards, Zhang2020} that combine the bisimulation metric with high-dimensional observations for learning-low dimensional representations for discrete and continuous \acrshort{mdp}s. The minimization of bisimulation and \acrshort{mdp} homomorphism metrics guarantee a Markovian latent space homomorphic to the original \acrshort{mdp} in terms of transition and reward function. However, two major challenges arise from using these two metrics: (i) preventing the collapse of the state representation to a trivial one, i.e., all the states mapped to the zero vector, in case of sparse reward functions, and (ii) learning of an accurate latent transition model.

Traditionally\marginpar{\scriptsize Contrastive Losses} in \acrshort{dl}, the collapsing of the representations is prevented using contrastive losses pushing negative embeddings apart, i.e., state representations belonging to different classes, (see Section \ref{subsec:contrastive_losses}). Contrastive losses are often used with approaches that do not rely on a decoder and  that are more prone to representation collapse, e.g., when only a transition model is used. InfoNCE loss and momentum encoders \cite{he2020momentum}  were used in combination with different \acrshort{drl} algorithms for solving continuous and discrete control problems. Such methods improve the quality of the learned representation by pushing apart non-consecutive, in terms of timesteps, latent states in \cite{stooke2021decoupling, laskin2020curl, mazoure2020deep, anand2019unsupervised, lee2020predictive}. The authors in \cite{you2022integrating} used the InfoNCE loss and the forward-model loss to improve the accuracy of the learned transition model and to add more structure to the state representation. Eventually, a return-based contrastive loss is introduced in \cite{Liu2021} to use the return, i.e., the discounted cumulative reward, as a discriminator between latent states. In contrast, a triplet contrastive loss is used in \cite{sermanet2018time} for learning state representations of robot behaviours from video demonstrations. These works show that timesteps and return information to discriminate between latent states are very effective strategies to improve performance and sample-efficiency of \acrshort{drl} algorithms.

\subsubsection{Learning State Representations with Priors} 

Instead\marginpar{\scriptsize Priors} of relying on the latent \acrshort{mdp} models which are often hard to learn or on the observation-reconstruction methods which are prone to fail in presence of irrelevant features in the observations, it is possible to incorporate known properties of the actual environment state in the loss functions used for training the encoder (see Section \ref{subsec:robotics_priors}).

The so-called robotics priors were introduced in \cite{jonschkowski2014state, jonschkowski2015learning}. These loss functions were inspired and shaped accordingly to simple physical laws in the context of robotics. In \cite{jonschkowski2017pves}, a new set of priors was introduced. Differently from the original set, the authors proposed priors regularizing the positions of the latent states and their velocities, i.e., variations between two consecutive (in time) latent states. Specifically, these priors better exploit the underlying smoothness of the physical world by promoting the latent states and their velocities to be close to each other in consecutive timesteps. In \cite{botteghi2021prior}, the authors introduced a novel set of robotics priors, which exploits action structures in continuous action spaces, a common scenario in robotics, while in \cite{hofer2016unsupervised} the set of original priors from \cite{jonschkowski2014state} were extended to multi-task learning with the addition of a task-coherence prior pushing apart latent states from different tasks. In \cite{lesort2019deep}, the robustness of the original priors was tested in the presence of visual distractors and domain randomization. The authors introduced the so-called reference-point prior compensating for the degrading performance due to the rapid changes in the environment background. The reference-point prior uses true state samples, which makes the reference-point prior makes the method weakly supervised, as references for patching together the different latent states from different trajectories. Because of their structure, the robotics priors can properly match latent states from the same trajectory, but struggle in enforcing coherence of embeddings from different trajectories. This problem becomes severe in the case of domain randomization. The robotics priors have been extended to deal with POMDP by \cite{morik2019state}. To do that, the authors introduced the so-call landmark prior that extends the reference prior from \cite{lesort2019deep} to multiple reference points. Using multiple reference points is crucial for the coherence of the different trajectories in the latent space. 

A limiting factor of all these methods is that the robotics priors do not use any of the task-specific information  contained in the rewards. One could argue that the state representation has to incorporate information about the goal, especially in the presence of irrelevant features in the observation (see the tutorial example in Section \ref{subsec:tutorial}). On this line of thought, the authors of  \cite{botteghi2021rewpriors} introduced a new set of robotics priors enforcing similarities and dissimilarities of state pairs using the rewards collected. These reward-shape priors are a natural way to incorporate task-specific knowledge in the state representation. 

 Results suggest that, while the robotics priors are a natural way of enforcing known physical properties into the latent state space and allow the learning of compact low-dimensional representations suitable for control and transfer learning from simulation to the world, they tend to struggle in adaptation and generalization to new environments and in the presence of irrelevant and randomly-moving features in the observations. These limitations can be reduced by incorporating true state values as anchors for the representations. However, true state values may only be available sometimes. The need for true state values limits the usage of these priors to general problems. Surprisingly, no one has attempted to combine the priors with additional objectives (except the causality and the temporal coherence loss), making it an unexplored area of research.

\subsubsection{Learning State Representation with Multiple Objectives}

Instead\marginpar{\scriptsize Multiple \hbox{Objectives}} of relying only on reconstruction, forward, reward, or contrastive losses, many authors proposed different combinations of multiple objectives in addition to the \acrshort{drl} algorithm objective of maximizing the cumulative reward. One of the earliest approach utilizing multiple objectives is \cite{jaderberg2016reinforcement}, which combines the \acrshort{drl} algorithm losses with a reward-model loss and a pixel-control loss for learning a Q-function capable of understanding which actions can change the value of each pixel maximally. In the context of robot navigation, in \cite{mirowski2016learning} additional objectives are used for inferring depth maps from \acrshort{rgb} images and detecting loop closure. In \cite{guo2020bootstrap}, the authors combine a recurrent encoder, a forward model - predicting the latent states from the history of observations -, and a reverse model - a model predicting the history of observations from the latent states. Recurrent models for estimating the belief in \acrshort{pomdp}s were proposed in \cite{gregor2019shaping, zintgraf2019varibad},  while the work in \cite{Agrawal2016} combined forward and inverse models for learning to poke with the Baxter robot in the real world. Reconstruction loss, reward loss, temporal coherence loss, causality loss, and forward and inverse model losses were used in \cite{de2018integrating}. 

Multiple objectives were also introduced in \cite{Thomas2017}, where a \textit{selectivity} loss for specifically learning K controllable features using K independent policies was combined with the losses of \acrshort{drl} agent. The selectivity of the features was defined as the expected difference in the representation of the current state and the next state. Selectivity can be interpreted as the reward for the control problem. The \acrshort{ae} loss and the selectivity of the policy are jointly optimized to learn a navigation task in a grid world. Despite the novelty, this approach does not scale well with the size of the problem, as K different policies have to be trained jointly. Unfortunately, K can vastly grow in complex scenarios. The authors in \cite{kim2021goal} used a goal-aware cross-entropy loss for discriminating goals in the context of multi-target \acrshort{drl}, and the authors in \cite{Allen} used an inverse model loss combined with a mutual information loss to learn a provably-Markovian state abstraction without the need for forward and reward models. In the context of multi-agent \acrshort{drl} in partially observable settings, the authors of  \cite{papoudakis2021agent} proposed a recurrent \acrshort{ae} for learning the relations between the trajectory of the controlled agent and the trajectory of the modelled agent. The recurrent \acrshort{ae} encoded sequences of observations and actions and produced a latent code that was used together with the raw observations as inputs to the agent.

Despite their different flavours, additional objectives tend to help \acrshort{drl} algorithms to improve the learned state representations and consequently improve the learning of the value function and policy. However, there is no structured procedure to know a priori the best auxiliary losses to use because the choice of objectives is very problem-dependent, as the plethora of approaches from the literature suggests. While adding many objectives together may seem a valid approach, one has to consider that each additional objective requires careful balancing with respect to the others, introducing additional complexity and computational cost. Identifying the auxiliary loss for each problem is an important open question of the \acrshort{srl} research.

\subsubsection{Learning State Representation without Auxiliary Models}

Improving the quality of the state representation can be done without auxiliary models\marginpar{\scriptsize No \hbox{Auxiliary} \hbox{Models}}. An example is the work of \cite{merckling2020state} where the Behavioral Cloning  \cite{torabi2018behavioral} loss was used for training the encoder. Two consecutive latent states were concatenated to form the vector that was fed to the \acrshort{drl} networks. A similar approach was used by \cite{shang2021reinforcement} where consecutive  latent states from a stack of frames were concatenated together with their latent flows, i.e. the difference of pairs of consecutive latent states, to form the augmented state vector that was used by the \acrshort{drl} algorithm. The authors of \cite{xu2020cocoi} proposed a framework for learning a context representation using the history of image stacks and the readings from a force sensor to learn how to push objects using a robot arm in continuous and discrete action spaces. The context representation was concatenated with the encoder features and fed to the \acrshort{drl} algorithm. Eventually, the authors of \cite{parisi2022unsurprising} show that pre-training the encoder on rich computer vision datasets is sufficient for learning good features for control.

\subsubsection{Summary}

In this section, we aim to summarize a key aspect of each methods briefly:
\begin{itemize}
    \item Most of the reviewed contributions rely on the \acrshort{ae} or on the \acrshort{cl} frameworks. \acrshort{ae}s and \acrshort{cl} are essential elements for \acrshort{srl} in \acrshort{drl} by providing the foundation for learning informative latent state variables and, consequently, the different latent \acrshort{mdp} models. \acrshort{mdp} models, which are prone to representation collapse when used alone. 
    \item The non-\acrshort{srl} objectives that we named as "Other Objectives" in the tables are often included when learning the representations. These loss function are often the \acrshort{drl} algorithms objectives, such as minimization of the temportal-difference error or maximization of the reward. Despite being done in practice often, the policy objective was empirically proven to be detrimental to the learning of the state representation \cite{yarats2021improving}. 
    \item One could also combine many methods to obtain the best representation, e.g., in \cite{de2018integrating}. However, training the encoder using multiple loss functions needs additional care in balancing the different terms. An unbalanced total loss may cancel some of the gradients of single losses and hinder the learning of the task-relevant features.
\end{itemize}

In Table \ref{tab:contribution-methods-modelfree}, we summarize the works presented in this section and we highlight the methods employed by each work. 
\begin{table*}[h!]
    \caption{Which SRL methods are used by each contribution? PCA=Principal Components Analysis, AE=Autoencoder, VAE=Variational Autoencoder, FW=Forward Model, RW=Reward Model, IN=Inverse Model, MDP H=MDP Homomorphism, BM=Bisimulation Metric, CL=Contrastive Learning.}
    \label{tab:contribution-methods-modelfree}
    \centering
    \tiny
    \begin{tabular}{|p{0.09\linewidth} | p{0.05\linewidth}|p{0.05\linewidth}|p{0.05\linewidth}|p{0.05\linewidth}|p{0.05\linewidth}|p{0.05\linewidth}|p{0.05\linewidth}|p{0.05\linewidth}|p{0.05\linewidth}|p{0.05\linewidth}|p{0.05\linewidth}|}
     \hline
    \hline
      Method ($\rightarrow$) Work ($\downarrow$)    & PCA & AE & VAE & FW & RW & IN & MDP H & BM & PR & CL & Other \hbox{Objectives}   \\
     \hline
    \hline
    \cite{curran2015using} & \faCheck & & & & & & & & & & \\
    \hline 
    \cite{parisi2017goal}  & \faCheck & & & & & & & & & & \\
    \hline
    \cite{lange2010deep}  & & \faCheck & & & & & & & & & \\
    \hline
    \cite{mattner2012learn}  & & \faCheck & & & & & & & & & \\
    \hline
    \cite{Alvernaz2017}  & & \faCheck & & & & & & & & & \\
    \hline
     \cite{kempka2016vizdoom}  & & \faCheck & & & & & & & & & \\
    \hline
    \cite{Higgins2017}  & &  & \faCheck & & & & & & & & \\
    \hline
    \cite{vincent2008extracting}   & &  & \faCheck & & & & & & & & \\
    \hline
    \cite{caselles2018continual}   & &  & \faCheck & & & & & & & & \\
    \hline
    \cite{Nair2018}   & &  & \faCheck & & & & & & & & \\
    \hline
    \cite{finn2016deep} & & \faCheck & & & & & & & \faCheck & & \\
    \hline
     \cite{caselles2019symmetry} & & & \faCheck & \faCheck & & & & & & & \\
    \hline
     \cite{raffin2019decoupling} & & \faCheck & &  & \faCheck & \faCheck & & & & & \\
    \hline
     \cite{van2016stable} & & & \faCheck & \faCheck & & & & & & & \\
    \hline
      \cite{kim2021acceleration} & & \faCheck & & \faCheck & & & & & & & \\
    \hline
    \cite{rafailov2021offline} & & & \faCheck & \faCheck & \faCheck & & & & & & \\
    \hline
     \cite{tennenholtz2021gelato} & & & \faCheck & \faCheck & \faCheck & & & & & & \\
    \hline
     \cite{shang2021self} & & \faCheck & & & & & & & & & \faCheck \\
    \hline
     \cite{de2018integrating} & & \faCheck & & \faCheck & \faCheck & \faCheck & & & \faCheck & \faCheck & \faCheck \\
    \hline
     \cite{yarats2021improving} & & \faCheck & & & & & & & & & \faCheck \\
    \hline
    \cite{lee2020stochastic}  & & & \faCheck & \faCheck & & & & & & & \faCheck \\
    \hline
    \cite{papoudakis2021agent} & & \faCheck & & & & & & & & & \\
    \hline
    \cite{Thomas2017} & & \faCheck & & & & & & & & & \faCheck\\
    \hline
    \cite{gelada2019deepmdp} & &  & & & & &  \faCheck & & & & \\
    \hline
    \cite{munk2016learning} & &  & &  \faCheck & \faCheck & & & & & & \\
    \hline
     \cite{castro2020scalable} & &  & & & & & & \faCheck & & & \\
         \hline
    \cite{biza2019online}  & &  & & & & & & \faCheck & & & \\
    \hline
     \cite{kemertas2021towards} & &  & & & & & & \faCheck & & & \\
    \hline
    \cite{Zhang2020} & &  & & & & & & \faCheck & & & \\
    \hline
    \cite{stooke2021decoupling} & &  & & & & & & & & \faCheck & \faCheck \\
    \hline
     \cite{laskin2020curl} & &  & & & & & & & & \faCheck & \faCheck \\
    \hline
    \cite{mazoure2020deep} & &  & & & & & & &  & \faCheck & \faCheck \\
    \hline
    \cite{anand2019unsupervised} & &  & & & & & & &  & \faCheck & \faCheck \\
    \hline
    \cite{lee2020predictive} & &  & & & & & & &  & \faCheck & \faCheck \\
    \hline
    \cite{you2022integrating} & &  & & \faCheck & & & & &  & \faCheck & \faCheck \\
    \hline
    \cite{Liu2021} & &  & & & & & & & & \faCheck & \faCheck \\
    \hline
   \cite{sermanet2018time}  & &  & & & & & & & & \faCheck & \faCheck \\
    \hline
    \cite{jaderberg2016reinforcement} & &  & & & \faCheck & & & & & & \faCheck \\
    \hline
    \cite{mirowski2016learning} & &  & & & & & & & & & \faCheck\\
    \hline
    \cite{guo2020bootstrap} & &  & & \faCheck & & & & & & & \faCheck \\
    \hline
    \cite{Agrawal2016} & &  & & \faCheck & & \faCheck & & & & & \faCheck \\
    \hline
    \cite{kim2021goal} & &  & & & & & & & & & \faCheck \\
    \hline
    \cite{Allen} & &  & & & &\faCheck & & & & & \faCheck \\
    \hline
      \cite{jonschkowski2014state} & &  & & & & & & &\faCheck & \faCheck & \\
    \hline
     \cite{jonschkowski2015learning} & &  & & & & & & &\faCheck & \faCheck & \\
    \hline
    \cite{jonschkowski2017pves} & &  & & & & & & &\faCheck & \faCheck & \\
    \hline
    \cite{botteghi2021prior} & &  & & & & & & &\faCheck & \faCheck  & \\
    \hline
    \cite{hofer2016unsupervised} & &  & & & & & & &\faCheck & \faCheck & \\
    \hline
     \cite{lesort2019deep} & &  & & & & & & &\faCheck & \faCheck & \\
    \hline
    \cite{morik2019state} & &  & & & & & & &\faCheck & \faCheck & \\
    \hline
    \cite{botteghi2021rewpriors}  & &  & & & & & & &\faCheck & \faCheck & \\
    \hline
    \cite{merckling2020state} & &  & & & & & & & & & \faCheck \\
    \hline
    \cite{shang2021reinforcement} & &  & & & & & & & & & \faCheck \\
    \hline
     \cite{xu2020cocoi} & &  & & & & & & & & & \faCheck \\
    \hline
    \cite{kalashnikov2018scalable} & &  & & & & & & & & & \faCheck \\
    \hline
    \hline
    \end{tabular}

\end{table*}

\clearpage

\subsection{State Representation for Accurate Latent Models in Model-based RL}\label{subsec:SRL_modelbased}

In this section, we review the most relevant model-based \acrshort{drl} methods for learning state representations through the latent forward models in Section \ref{subsubsec:mbrl_fwm} or through the latent forward and reward models in Section \ref{subsubsec:mbrl_frwm}. In the first case, the goal is to learn the state representation that allows learning the forward latent dynamics by only relying on the tuple $(\bm{o}_t, \bm{a}_t, \bm{o}_{t+1})$, while in the second case, the goal is to learn the state representation that encodes information about the forward dynamics but also of the reward function by relying on  $(\bm{o}_t, \bm{a}_t, \bm{o}_{t+1}, r_t)$.

\subsubsection{Learning State Representation with Latent Forward Models}\label{subsubsec:mbrl_fwm}

Embed to Control (\acrshort{e2c}) \cite{watter2015embed} used a \acrshort{vae} for learning a low-dimensional latent state space that was used as input to a linear latent forward model for obtaining a linear reduced-order model of a system from high-dimensional observations. The linear forward model forced the encoder to encode features that linearize the latent state space dynamics. The \acrshort{vae} and the latent forward model are jointly trained together. After learning the reduced-order model, the control was performed via model-based techniques relying on the learned model, such as iterative linear quadratic regulation and approximate inference control. The authors of \cite{wahlstrom2015pixels} proposed a similar approach to \acrshort{e2c}. However, a nonlinear latent forward model was combined with an \acrshort{ae}. Another example is the work of \cite{Assael}, where again, an \acrshort{ae} with nonlinear forward dynamics was employed with the addition of a latent-space consistency loss. The consistency loss enforces that the latent states are close to the embedding of their next observation, providing further regularization to the latent state space -this is similar to what the temporal coherence prior (in Equation \eqref{eq:tempcohexample}) does. In both cases, the learned latent forward model was used to generate potential trajectories of the systems that can be optimized by any trajectory-optimization control system, e.g. Model Predictive Control (\acrshort{mpc}) \cite{rawlings2000tutorial},  to maximize the reward. 

The State Space Models (\acrshort{ssm}s) \cite{chua2018deep, buesing2018learning} introduced an approach for learning deterministic and stochastic state representations for state estimation in the context of model-based RL. The deterministic \acrshort{ssm}s used an \acrshort{ae} or a \acrshort{vae} in combination with a deterministic forward model, while the stochastic \acrshort{ssm}s used a \acrshort{vae} in combination with a probabilistic forward model. Differently from many works in the literature, the stochastic forward model did not directly approximate $\bm{z}_{t+1} \sim p(\bm{z}_{t+1}|\bm{z}_t,\bm{a}_t)$, but made use of a latent variable $\bm{\eta_{t+1}} \sim p(\bm{\eta}_{t+1}|\bm{z}_{t},\bm{a}_{t},\bm{o}_{t+1})$ to predict the next latent state $\bm{z}_{t+1}=T(\bm{z}_t,\bm{a}_t,\bm{\eta}_{t+1})$. Similar approaches to stochastic \acrshort{ssm}s, combining \acrshort{vae}s and probabilistic forward models, were introduced by \cite{karl2016deep, doerr2018probabilistic,krishnan2015deep,fraccaro2017disentangled, ha2018recurrent, ha2018world}  for state estimation \cite{karl2016deep, doerr2018probabilistic,krishnan2015deep,fraccaro2017disentangled}, and state estimation with control \cite{ha2018recurrent, ha2018world}. 

All the previously-introduced approaches relied on a vector representation of the latent states and a \acrshort{fcnn} for approximating the latent forward models. However, encoding observations as vectors is not the only option. One could think of exploiting inductive biases, i.e., prior knowledge, and encoding the observations as graph nodes. The Contrastive Structured World Model \cite{kipf2019contrastive} encoded the latent states and actions as an element of a graph (nodes and edges respectively) and used a nonlinear forward model approximated by a Message Passing Graph Neural Network \cite{gilmer2017neural, battaglia2018relational}. Another approach embedding observations into a graph is Plan2Vec \cite{Yang2020}, where the graph was constructed by learning the distance between the nodes, i.e. the latent states and a search algorithm was used to learn the shortest path on the graph from the current state to a goal state.

\subsubsection{Learning State Representation with Latent Forward and Reward Models}\label{subsubsec:mbrl_frwm}

The problems of continuous control, state representation learning, and latent models learning were tackled by PlaNet  \cite{hafner2019learning}, and Dreamer \cite{hafner2019dream}, where a  convolutional and recurrent \acrshort{vae} with probabilistic latent forward and reward model was learned. The policy was learned via \acrshort{mpc} by PlaNet and via actor-critic \acrshort{drl} by Dreamer. Control via \acrshort{mpc} was employed in \cite{van2021deepkoco} after learning a task-agnostic state representation using an \acrshort{ae}, a linear forward model relying on the Koopman operator \cite{brunton2022data}, and a reward model. Discrete control for learning to play the Atari games in a model-based fashion was done in \cite{kaiser2019model} via predictive autoencoding, i.e., an \acrshort{ae} predicting the next frame given the current one and the action, and a reward model. The learning of a plannable \acrshort{mdp} homomorphism was proposed by \cite{van2020plannable}, using a similar approach to \cite{gelada2019deepmdp} but in the context of model-based \acrshort{drl}. Additionally, the authors of \cite{van2020plannable} learned an action representation and used a contrastive hinge loss (see Section \ref{sec:OverviewMethods}) to prevent the state representation from collapsing due to sparse rewards and the lack of a decoder. After learning the state representation, latent forward and reward models, they discretized the learned state and action space and used them for planning and learning the policy without further real-world interaction via Value Iteration \cite{Sutton1998}. 

A combination of model-free and model-based DRL was proposed by \cite{franccois2019combined}, where the state representation was learned using multiple objectives: latent forward model, discount factor prediction, contrastive loss, and  \acrshort{drl} loss. Moreover, they introduce a loss for improving the interpretability of the learned representations. Multiple objectives were also used by the authors of \cite{zhao2021consciousness} where a state representation was learned via the \acrshort{drl} loss, the forward, reward, and terminal condition losses - using the terminal-condition model to predict the end of the episodes. However, instead of relying on vector representations, the authors used a set-based representation, learned through a set-based encoder \cite{zaheer2017deep}. A set can be viewed as a particular case of a graph with no edges \cite{bronstein2021geometric}. We will discuss Geometric Deep Learning in Section \ref{sec:discussion}. Moreover, they proposed the so-called consciousness bottleneck using a multi-layer self-attention \cite{vaswani2017attention} to enforce sparsity of the features and representations that the forward, reward, and terminal condition model could learn. The set representation and the attention mechanism allowed outstanding generalization properties of the agent when placed in unseen environments.

\subsubsection{Summary}

In this section, we aim to summarize a key aspect of each category of methods briefly:
\begin{itemize}
    \item When latent models are available, classic model-based techniques for trajectory optimization, e.g. \acrshort{mpc}, are viable options next to \acrshort{drl}.
    \item In model-based \acrshort{drl}, \acrshort{ae}s and \acrshort{vae}s tend to be preferred to \acrshort{cl}. Methods relying on a decoder often outperform \acrshort{cl}-based ones. An empirical comparison was provided in \cite{hafner2019dream}.
    \item Probabilistic models are more often employed than deterministic ones. However, the quantification of the uncertainties is often under-investigated. In some cases, all the system's uncertainties are quantified by the variance of the encoder neural network, e.g. \cite{hafner2019learning, hafner2019dream}, while in others, even when the variances of the latent models are learned, no analysis on the quantified uncertainties is done, e.g. \cite{karl2016deep, doerr2018probabilistic, krishnan2015deep, fraccaro2017disentangled}.
\end{itemize}

In Table \ref{tab:contribution-methods-modelbased}, we summarize the most prominent \acrshort{srl} methods in the context of model-based \acrshort{drl}.
\begin{table*}[h!]
    \caption{Which SRL methods are used by each work? PCA=Principal Components Analysis, AE=Autoencoder, VAE=Variational Autoencoder, FW=Forward Model, RW=Reward Model, IN=Inverse Model, MDP H=MDP Homomorphism, BM=Bisimulation Metric, CL=Contrastive Learning.}
    \label{tab:contribution-methods-modelbased}
    \centering
    \tiny
    \begin{tabular}{|p{0.09\linewidth} | p{0.05\linewidth}|p{0.05\linewidth}|p{0.05\linewidth}|p{0.05\linewidth}|p{0.05\linewidth}|p{0.05\linewidth}|p{0.05\linewidth}|p{0.05\linewidth}|p{0.05\linewidth}|p{0.05\linewidth}|p{0.05\linewidth}|}
     \hline
    \hline
      Method ($\rightarrow$) Work ($\downarrow$)    & PCA & AE & VAE & FW & RW & IN & MDP H & BM & PR & CL & Other \hbox{Objectives}   \\
     \hline
    \hline
    \cite{watter2015embed} & & & \faCheck & \faCheck & & & & & & & \\
    \hline
     \cite{wahlstrom2015pixels} & & \faCheck & & \faCheck & & & & & & & \\
     \hline
     \cite{Assael} & & \faCheck & & \faCheck & & & & & \faCheck & & \\
     \hline
     \cite{buesing2018learning} & & \faCheck & \faCheck & \faCheck & & & & & & & \\
     \hline
     \cite{karl2016deep} & & & \faCheck & \faCheck & & & & & & & \\
     \hline
      \cite{doerr2018probabilistic} & & & \faCheck & \faCheck & & & & & & & \\
     \hline
     \cite{krishnan2015deep} & & & \faCheck & \faCheck & & & & & & & \\
     \hline
     \cite{fraccaro2017disentangled} & & & \faCheck & \faCheck & & & & & & & \\
     \hline
     \cite{kipf2019contrastive} & & & & \faCheck & & & & & & \faCheck & \\
     \hline
      \cite{Yang2020} & & & & \faCheck & & & & & & \faCheck & \faCheck \\
     \hline
     \cite{hafner2019learning} & & & \faCheck & \faCheck & \faCheck & & & & & & \\
     \hline
     \cite{hafner2019dream} & & & \faCheck & \faCheck & \faCheck & & & & & & \\
     \hline
     \cite{van2021deepkoco} & & \faCheck & & \faCheck & \faCheck & & & & & & \\
     \hline
     \cite{kaiser2019model} & & \faCheck & & \faCheck & \faCheck & & & & & & \\
     \hline
     \cite{van2020plannable} & & & & & & & \faCheck & & &\faCheck & \\
     \hline
     \cite{franccois2019combined} & & & & \faCheck & \faCheck & & & & & \faCheck & \faCheck \\
     \hline
    \cite{zhao2021consciousness} & & & & \faCheck & \faCheck & & & & & & \faCheck \\
     \hline
    \hline
    \end{tabular}

\end{table*}

%% file: StateRepresentationLearning/5-AdvancedMethods.tex
\begin{figure}[h!]
    \centering
         \includegraphics[width=1.0\textwidth]{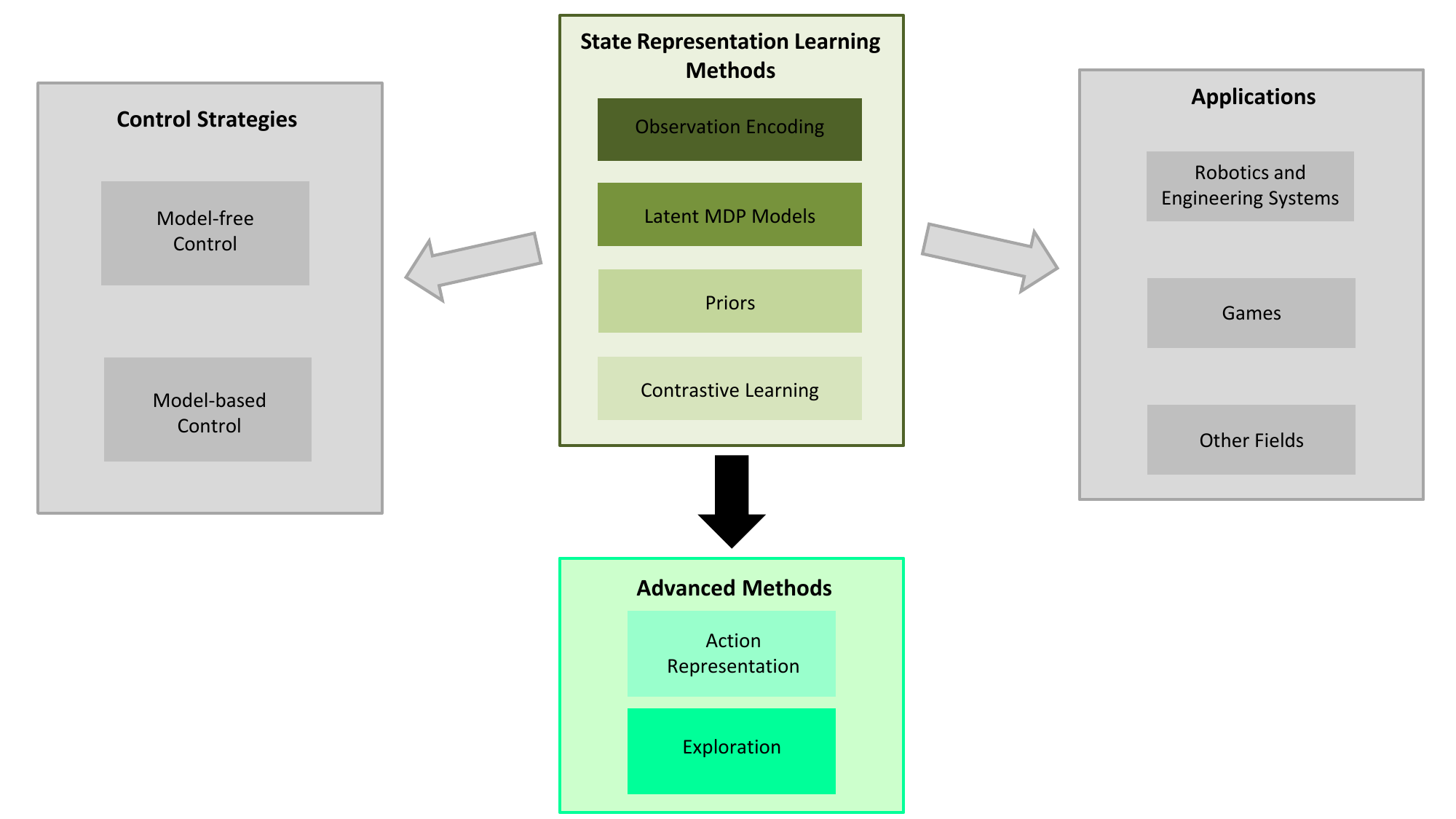}
        \caption{Overview of advanced methods.}
        \label{fig:bigpicture3}
\end{figure}

In this section (see Figure \ref{fig:bigpicture3}), we provide a different dimension to the \acrshort{srl} methods we discussed in Section \ref{sec:unsupervisedlearningDRL}. We focus on advanced derivations of the \acrshort{srl} methods that solve not only the \acrshort{srl} problem but also two broader class of challenges of \acrshort{drl}, namely (i) the action representation and (ii) the exploration-exploitation trade-off. 
In particular, we enhance the review by studying two crucial challenges for scaling \acrshort{drl} to complex applications such as:
\begin{itemize}
    \item transforming large action spaces in lower-dimensional latent action spaces for easier policy learning and
    \item enhancing exploration skills of the \acrshort{drl} agents by using the latent representations to generate intrinsic rewards.
\end{itemize}

\subsection{Learning Action Representations}\label{subsec:SRL_actionrepresentation}

Reducing the dimensionality of the observations by encoding them in lower-dimensional latent state space is a powerful and important tool for sample-efficient \acrshort{drl}. In this section, we argue that the same mechanism of encoding observations can be used for the agent's actions. While many works have only focused on learning state representations, only a few have focused on learning action representations or jointly learning state and action representations. This section reviews the methods relying on action (and state-action) representations.

One of the first examples is the work of \cite{he2015deep} in the context of Natural Language Processing \cite{uc2022survey}. The goal was to train an agent that, by looking at a sentence, i.e. the state, could predict the sentence that would follow. State and actions are recurrently and independently encoded, and their embeddings, i.e. the state and the action representations, were fed to the Q-network to estimate the Q-value. Learning the Q-value in the latent state-action space proved to be the critical ingredient for outperforming other approaches in the literature. Large and discrete action spaces are traditionally a challenge for \acrshort{drl} algorithms as it is hard to sufficiently explore the action space. However, one could think of embedding the action space in a low-dimensional latent space to exploit similarities among actions. Analogously to the state representation problem, mapping actions with similar effects on the states close together in the latent action space makes learning the value function and the policy easier. In this context, we find the work of \cite{dulac2015deep} that, given a large, discrete action space, first clustered the actions  via K-Nearest Neighbours (\acrshort{knn}), and then evaluated the Q-value of those actions only, drastically reducing their number and the difficulty of evaluating their Q-value. This approach efficiently saved evaluations of the Q-value, making the \acrshort{drl} algorithm efficient for large action spaces. Another way to learn action embeddings was proposed by \cite{chandak2019learning}. The authors used an inverse model with a bottleneck. Two consecutive states were used as inputs to an inverse model to predict the latent action between two states. Then, the latent action was fed to an action-decoder network that maps the latent action to the original action space. A continuous latent policy was efficiently trained to map states to latent actions. 

State representation and action representations were combined by \cite{pritz2020joint} with similar principles to \cite{chandak2019learning} using a forward model instead of an inverse one. Again, embedding states and actions proved to be a powerful way to achieve sample efficiency in \acrshort{drl}, especially in the case of large discrete action spaces where embedding the actions added structures and exploited their similarities. In the same line of thought, the authors of \cite{botteghi2021low} proposed a framework for jointly learning state and action representations from high-dimensional observations by relying on the \acrshort{mdp} homomorphism metrics. The latter guarantees optimality and equivalence of the learned latent policy that maps latent states to latent actions and the overall policy from observations to actions. A similar architecture to \cite{pritz2020joint} was proposed by \cite{whitney2019dynamics}, but, similarly to \cite{he2015deep}, the action representation was learned from a sequence of actions. Because sequences of actions were mapped to a latent action vector, decoding this vector back to a sequence of actions was an ill-posed inverse problem requiring regularization via an additional loss. 

\subsection{State Representation for Efficient Exploration in Complex Environments}\label{subsec:SRL_exploration}

The exploration-exploitation dilemma is one of the most studied problems in \acrshort{drl}. Efficient exploration of the unknown world while exploiting known high rewards is especially problematic in large and complex environments where random exploration or heuristics cannot guarantee sufficient coverage of the underlying state space. This aspect is even more severe when the state of the environment is not available to the agent and only high-dimensional observations are received by the agent. Herein, we show how \acrshort{srl} can be crucial in the context of better exploration of the world.

Curiosity-driven exploration \cite{schmidhuber1991curious, houthooft2016vime} builds its success on the concept of rewarding the agent for exploring unexplored areas of the world. Unexplored areas are areas without or with limited samples. Therefore, the dynamics of these areas of the state space is hard to predict using, for example, forward models. By rewarding the agent with a scalar value proportional to the prediction error, we promote visiting uncertain states more often and enhance the agent's exploration skills. In this section, we focus on a curiosity-driven exploration through learning state representations to highlight the versatility of \acrshort{srl} further and its benefits in many aspects of \acrshort{drl}. Again \acrshort{srl} and the process of encoding observations to a low-dimensional space while maintaining the relevant information for exploration is a crucial element. For a complete review on curiosity-driven exploration, we refer the reader to \cite{yang2021exploration}. 

One of the first works combining learned state representations and latent \acrshort{mdp} models to generate curiosity rewards is \cite{pathak2017curiosity}, where an encoder trained by an inverse model and a forward model was used to reduce the dimensionality of the observations. The inverse model is used to focus the learning of features that are controllable by the agent and, therefore, more interesting for exploration. The latent forward model and the embeddings of the observations are used to compute the prediction error that was properly scaled and added to the environment reward for promoting exploration. However, this method presented several limitations that were highlighted in \cite{burda2018large}. The major one arises when the environment dynamics is stochastic. As it is, the curiosity-driven agent could not distinguish prediction uncertainties due to the lack of data from prediction uncertainties due to the stochasticity of the environment. Therefore, the agents were subjected to learning policies that exploited the stochasticity of the environment without any exploratory meaning. Improvements to curiosity-driven exploration to limit the problem of the stochasticity of the environment were proposed by\cite{pathak2019self}, where an ensemble of latent forward models was used to make N independent prediction-error terms. The curiosity bonus was then computed by looking at the variance of the error terms to exploit the disagreement of the models. Exploration can also be promoted if the agent learns to avoid terminal states that the actions cannot reverse. With this idea in mind, the authors in \cite{grinsztajn2021there} developed an approach for intelligent exploration that penalized visiting irreversible states. Differently, a randomly-initialized \acrshort{ae} was used to generate exploration bonuses for aiding the visit of different locations of the latent state space \cite{seo2021state}.

The novelty of a state can be defined not only by prediction error but also in terms of its distance from other visited states. This idea was exploited by \cite{savinov2018episodic} with the method named episodic curiosity. First, observations were embedded in the latent state space. Then, each embedding was compared with the embeddings of novel states, stored in the so-called reachability buffer to generate a score depending on how far the embedding was from the others in the buffer. If the score was above a certain threshold, i.e. the observation was deemed novel, the observation was added to the reachability buffer. The score was used as a curiosity bonus, rewarding the agent for visiting states far from the set of known ones. A similar concept is used by \cite{tao2020novelty}, where the curiosity bonus was defined by looking at the latent states' \acrshort{knn} to reward visiting states with high \acrshort{knn} scores. States with high \acrshort{knn} scores belonged to areas with more exploratory potential. Closely related, we found the work of \cite{badia2020never}, where an inverse model aided the learning of the latent states used for computing the \acrshort{knn} score to focus on controllable features. The work of \cite{liu2021behavior}  used a contrastive loss instead of the inverse model, and the work of  \cite{Yarats2021} that, differently from the other approaches, simultaneously learned a state representation that was directly used by the policy and to generate the exploration bonus. 

Eventually, we find the methods that keep track of the visiting count of each state and rewarding exploration strategies that achieve uniform coverage if the state space \cite{martin2017count,tang2017exploration,machado2020count}. However, these methods struggle in large discrete and continuous state spaces where complicated and expensive density models have to be learned and updated during the exploration process.

\subsection{Summary}

In this section, we aim to summarize a key aspect of each category of methods:
\begin{itemize}
    \item Combining the learning of state and action representations from high-dimensional observations has been proposed only in \cite{whitney2019dynamics, botteghi2021low}. These approaches may open the door to more sample-efficient \acrshort{drl} algorithms combined with temporal abstractions of actions and states.
    \item In particular, while additional objectives are very often used in this context, we notice that, especially in exploration, forward and inverse models are used to extract the set of epistemically uncertain yet controllable features. 
    \item Similarly to model-based \acrshort{drl}, the exploration via forward models requires precise quantification and discrimination of aleatoric and epistemic uncertainties. Hence, for further progress in the field, it is important to deepen the connection between uncertainty quantification, \acrshort{srl}, and learning latent dynamical models.
\end{itemize}

In Table \ref{tab:contribution-methods-advancemethods}, we summarize how the reviewed works that utilize the \acrshort{srl} methods.
\begin{table*}[h!]
    \caption{Which \acrshort{srl} methods are used by each work? PCA=Principal Components Analysis, AE=Autoencoder, VAE=Variational Autoencoder, FW=Forward Model, RW=Reward Model, IN=Inverse Model, MDP H=MDP Homomorphism, BM=Bisimulation Metric, CL=Contrastive Learning.}
    \label{tab:contribution-methods-advancemethods}
    \centering
    \tiny
    \begin{tabular}{|p{0.09\linewidth} | p{0.05\linewidth}|p{0.05\linewidth}|p{0.05\linewidth}|p{0.05\linewidth}|p{0.05\linewidth}|p{0.05\linewidth}|p{0.05\linewidth}|p{0.05\linewidth}|p{0.05\linewidth}|p{0.05\linewidth}|p{0.05\linewidth}|}
     \hline
    \hline
      Method ($\rightarrow$) Work ($\downarrow$)    & PCA & AE & VAE & FW & RW & IN & MDP H & BM & PR & CL & Other \hbox{Objectives}   \\
     \hline
    \hline
     \cite{pathak2017curiosity} & & & & \faCheck & & \faCheck & & & & & \\
    \hline
    \cite{burda2018large} & & & & \faCheck & & \faCheck & & & & & \\
    \hline
    \cite{pathak2019self} & & & & \faCheck & & & & & & & \\
    \hline
    \cite{grinsztajn2021there} & & & & & & & & & & & \faCheck \\
    \hline
     \cite{savinov2018episodic} & & & & & & & & & & & \faCheck \\
    \hline
    \cite{tao2020novelty} & & & & \faCheck & \faCheck & & & & & \faCheck & \faCheck \\
    \hline
    \cite{badia2020never} & & & & & & \faCheck & & & & & \faCheck\\
    \hline
     \cite{liu2021behavior} & & & & & & & & & & \faCheck & \faCheck \\
    \hline
    \cite{Yarats2021} & & & & & & & & & & & \faCheck \\
    \hline
    \cite{seo2021state} & & & & & & & & & & & \faCheck \\
    \hline
    \cite{martin2017count} & & & & & & & & & & & \faCheck \\
    \hline
    \cite{tang2017exploration}& & \faCheck & & & & & & & & & \faCheck \\
    \hline
     \cite{machado2020count} & & \faCheck & & \faCheck & & & & & & & \faCheck \\
    \hline
    \cite{he2015deep} & & & & & & & & & & & \faCheck \\
    \hline
    \cite{dulac2015deep} & & & & & & & & & & & \faCheck \\
    \hline
    \cite{chandak2019learning} & & & & & & \faCheck & & & & & \\
    \hline
    \cite{pritz2020joint}  & & & & \faCheck & & & & & & & \\
    \hline
     \cite{whitney2019dynamics} & & & \faCheck & \faCheck & & & & & & & \\
    \hline
    \cite{botteghi2021low} & & & & & & & \faCheck & & & \faCheck & \faCheck \\
    \hline
     \hline
    \hline
    \end{tabular}

\end{table*}

In Figure \ref{fig:bigpicture3}, we show the use of the \acrshort{srl} methods in exploration and action representation. In both cases, the latent \acrshort{mdp} models are the most often employed methods and the robotics priors are never used.

%% file: StateRepresentationLearning/6-Applications.tex
\begin{figure}[h!]
    \centering
         \includegraphics[width=1.0\textwidth]{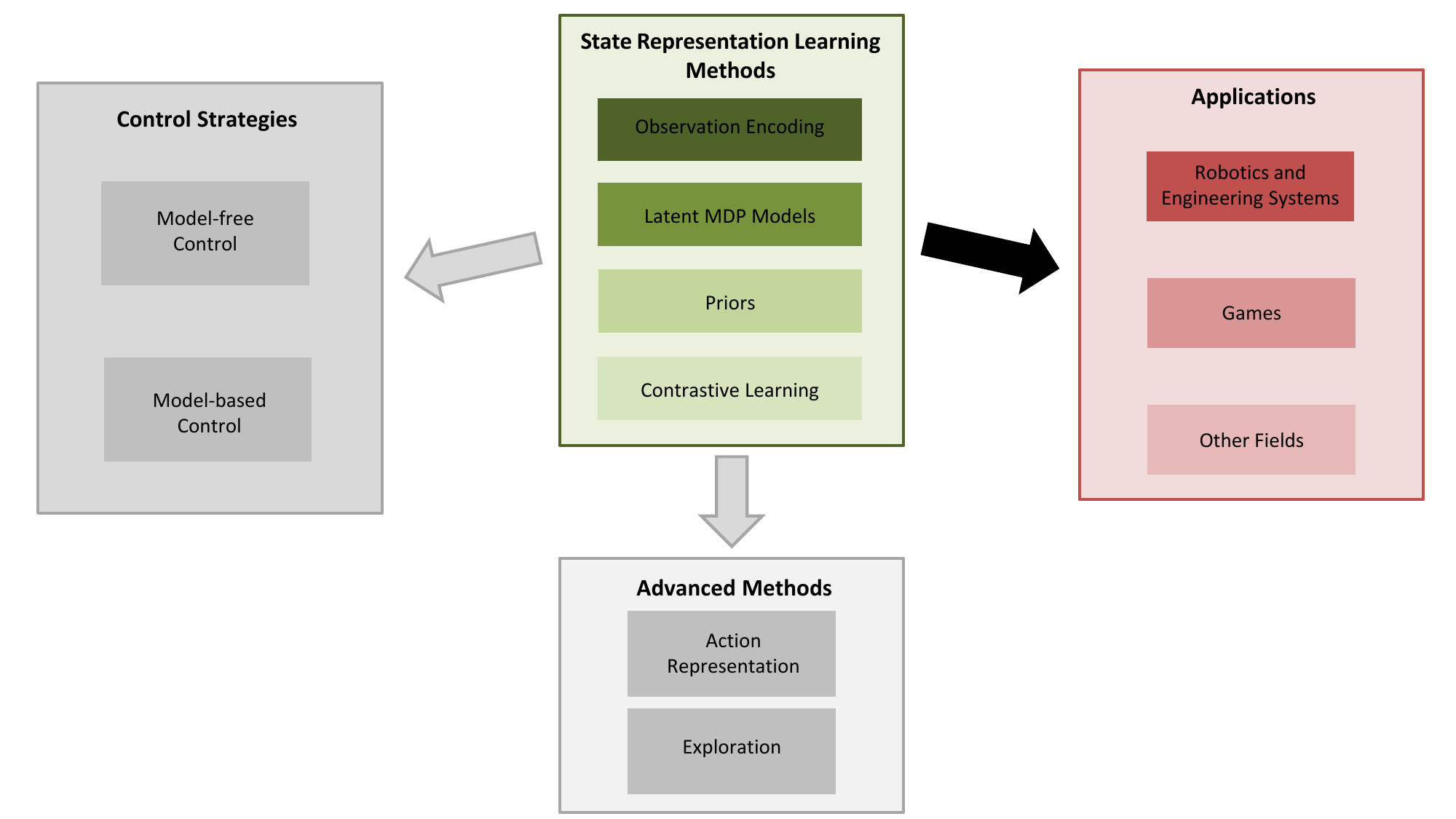}
        \caption{Overview of the applications.}
        \label{fig:bigpicture4}
\end{figure}

After an in-depth presentation of the basic approaches, methods and advanced methods, we focus on the most relevant applications of \acrshort{srl} (see Figure \ref{fig:bigpicture4}). State estimation is one of the most well-studied problems in dynamical systems evolving over time. Predicting such systems' natural and controlled evolution is crucial when we talk about physical systems, such as robotics and fluid dynamics, games, e.g. chess or Atari, and other fields such as health and finance.

\subsection{Robotics and Engineering Systems}

Physical systems often present a crucial challenge: real-world interaction. The interaction with the real world, and the consequent need for learning from limited data, is possibly the most significant obstacle \acrshort{drl}-based control systems must overcome. In the context of physical systems, robotics is one of the most researched application domains due to the potential societal impacts of autonomous robots.

Learning directly in the real world is often too expensive and dangerous for the hardware. Thus, simulators are often used as a proxy for the real world. However, the simulation-to-reality gap often hinders the direct transfer of the learned behaviors to the real world \cite{kober2013reinforcement}. \acrshort{srl} plays an important role here as well. Compared to \acrshort{drl} algorithms learning directly from high-dimensional measurements, the policies learned using the low-dimensional state representations are more robust to noise, to visual distractors, and to discrepancies between the simulator and the real world. Moreover, due to the higher sample efficiency of the \acrshort{drl} algorithm using the learned state representations, it is possible to fine-tune the performance with a few real-world samples or, in some cases, even directly transfer the learned policies in simulated environments to the real world. Alternatively, one could think of training the algorithms directly in the real world. While model-free approaches are predominant with simulators, model-based approaches become more appealing when learning directly in the real world. Therefore, learning accurate and reduced-order latent models from measurements is crucial. Again, sample efficiency and robustness are essential concepts that must be considered.

The literature proposed various robotics applications, spanning from robot navigation, to dexterous manipulation with robot arms, continuous control, and autonomous driving. In particular, in the context of (wheeled-)robot navigation, we found many approaches relying on pixel observations for learning target-reaching navigation policies in unknown environments in the presence of obstacles. Many approaches studied this problem in simulated environments \cite{jonschkowski2014state, botteghi2021low, hofer2016unsupervised, stooke2021decoupling, laskin2020curl, morik2019state, jaderberg2016reinforcement, mirowski2016learning, guo2020bootstrap, gregor2019shaping, Higgins2017} (see Section \ref{subsec:benchmakrs}), while only a few studied the problem of transferring the policies learned in the simulated environment to the real-world \cite{raffin2019decoupling, botteghi2021rewpriors, botteghi2021low} or the learning directly in the real-world \cite{jonschkowski2015learning}. Another popular application for \acrshort{srl} is dexterous manipulation using robot arms. Examples of tasks are pick-and-place, poking, pushing, and manipulation of deformable objects. Similarly to the case of robot navigation, we found approaches focusing on simulated environments \cite{parisi2017goal, kim2021acceleration, raffin2019decoupling, shang2021self, lesort2019deep, Agrawal2016, kim2021goal,merckling2020state, Yang2020}, on transfer learning from simulation to the real world \cite{Higgins2017, xu2020cocoi}, and on real-world learning \cite{Nair2018, finn2016deep, van2016stable, rafailov2021offline, sermanet2018time, doerr2018probabilistic}. \acrshort{drl} algorithms for continuous control are tested on traditional control baselines such as pendulum, cart-pole, double pendulum, and more advanced ones such as cheetah, hopper, and humanoid \cite{yarats2021improving, lee2020stochastic, rafailov2021offline, tennenholtz2021gelato, kemertas2021towards, lee2020predictive, you2022integrating, Liu2021, Allen, shang2021reinforcement, watter2015embed, Assael, hafner2019learning, hafner2019dream, van2020plannable}. The stabilization of the pendulum around an unstable equilibrium is one of the most studied baselines \cite{mattner2012learn, van2016stable, kemertas2021towards, jonschkowski2017pves, watter2015embed, wahlstrom2015pixels, Assael, karl2016deep, fraccaro2017disentangled, van2021deepkoco}. Finally, car simulators are used in \cite{Zhang2020, de2018integrating, ha2018world, ha2018recurrent}.

In the context of continuous control with \acrshort{drl}, especially in recent years, fluid dynamics has gathered more attention \cite{garnier2021review}. Traditionally, fluid dynamics has focused only on model-order reduction and state estimation. \acrshort{srl} principles have yet to be exploited in this field, but they can be crucial for  controlling these high-dimensional systems. Another recent topic of interest is Digital Twins \cite{batty2018digital, tao2019make, el2018digital}. Digital Twins are virtual and precise copies of engineering systems that can be used to predict and monitor the system's evolution \cite{tao2019make}. Digital Twins need to function in real-time, making dimensionality reduction and \acrshort{srl} a critical aspect of their construction.

\subsection{Games}

Games such as chess, Go, Atari and Montezuma's Revenge are other successful applications of \acrshort{drl}. While sample efficiency might not appear critical due to the lack of real-world interaction, solving games with long-term dependencies and multiple difficulty levels is extremely challenging. Solving such games requires (often) unobtainable amounts of data due to the need for complex exploration policies and learning to collaborate with multiple players. Therefore, learning good and general state representations and latent models can allow the development of new approaches and more advanced solutions to, for example, efficient exploration or credit assignment. Unlike physical systems, where continuous action spaces are often preferred for the smoothness of the control laws, games present discrete action spaces spanning from simple north-south-west-east movements on a grid to more complex actions such as grabbing objects, opening doors and moving different chess pieces.

In particular, one of the most studied examples is the grid-world and its variations \cite{parisi2017goal, lange2010deep, caselles2018continual, botteghi2021low, caselles2019symmetry, gelada2019deepmdp, castro2020scalable, biza2019online, jaderberg2016reinforcement, Thomas2017, papoudakis2021agent, kipf2019contrastive, Yang2020, van2020plannable, franccois2019combined, zhao2021consciousness}. These examples mostly focused on learning navigation policies to traverse the grid world and reach one or more different targets. Another popular baselines are the Atari games \cite{gelada2019deepmdp, castro2020scalable, stooke2021decoupling, laskin2020curl, anand2019unsupervised, Liu2021, jaderberg2016reinforcement, shang2021reinforcement, buesing2018learning, kipf2019contrastive, kaiser2019model}  comprising a variety of different games with various level of difficulties (see Section \ref{subsec:benchmakrs}). Other games used for testing \acrshort{srl} methods are Vizdoom \cite{kempka2016vizdoom, Alvernaz2017, kim2021goal, ha2018world, ha2018recurrent}, Ms Pacman \cite{mazoure2020deep, buesing2018learning}, Minecraft \cite{shang2021self}, and Super Mario \cite{parisi2017goal}.

\subsection{Other fields}

\acrshort{drl} has become increasingly popular in many applications in different fields. However, \acrshort{srl} methods remain understudied. Here, we show how the problem-specific challenges of each field can enormously benefit from \acrshort{srl} methods.

The first example is economics. Decision-making problems such as the stock market, portfolio management, capital allocation, online services, and recommendation systems can all be tackled with \acrshort{drl} \cite{mosavi2020comprehensive}. Improving information compression, quantifying uncertainties in the measured variables over wide windows of time, and providing interpretable results are some problems the \acrshort{srl} methods could potentially solve. 

Secondly, the health domain \cite{yu2021reinforcement} has recently shifted its focus towards dynamic treatment regimes with long-term effects for chronic disease, automated medical diagnoses, and health resource scheduling and allocation of time, nursing resources, and diagnostic devices. Many challenges arise in this field, such as the sample efficiency due to the limited data, state-action space engineering due to multiple high-dimensional measurements over long time windows and potentially coming from different devices, interpretability of the decisions and the learned latent variables, development of patient-specific models, dealing with uncertainties, exploration of new treatment regimes, and the integration of prior knowledge. These challenges perfectly fit the scope of \acrshort{srl} methods, and \acrshort{srl} can open new frontiers in health research.

\acrshort{drl} has also invaded domains such as the Internet of Things and communication networks \cite{xiong2019deep}. For instance, \acrshort{drl} methods are used for solving power control and power management, reducing latency and traffic load problems, cloud computing, smart grid, and resource allocation. With many users involved, the key challenges include dealing with multi-agent systems and encoding partial information over time.

Natural Language Processing presents problems that \acrshort{drl} has started to solve, such as language understanding, text generation, machine translation and conversational systems \cite{uc2022survey}. The proper encoding and representation of sentences and sounds, and the generalization to new speakers are crucial for decision-making. \acrshort{srl} may be again an essential element for further progress.

Eventually, we find the field of Explainable AI (\acrshort{exai}) \cite{wells2021explainable, nauta2022anecdotal}. \acrshort{exai} aims to open the black box and, in \acrshort{drl}, understand the agent's behavior. \acrshort{srl} can be seen as a way to simplify the problem of decision-making by separating feature extraction and policy learning by identifying the low-dimensional set of important variables for control. These \acrshort{srl} goals align with \acrshort{exai}, and further research should be conducted to strengthen this connection.

\subsection{Summary}

In Figure \ref{fig:bigpicture4}, we show the use of the \acrshort{srl} methods for the different groups of applications. In particular, we highlight:
\begin{itemize}
    \item latent models, and priors play a crucial role for physical systems where prediction of the dynamics of the world is often as important as the learning of the policy. When dealing with physical interaction with the world, model-based \acrshort{drl} is preferable to model-free.
    \item model-free \acrshort{drl} is often preferred for games, and contrastive losses are often employed next to the \acrshort{drl} algorithm losses (e.g. reward maximization) to help the learning of good state representations.
    \item in other fields, \acrshort{srl} principles have yet to be studied and exploited. Therefore, each of the four \acrshort{srl} methods has the potential to bring important contributions to the different fields.
\end{itemize}

%% file: StateRepresentationLearning/7-EvaluationandBenchmark.tex
\subsection{Evaluation Metrics}\label{subsec:EvalMetrics}

Evaluating the learned representations is crucial for assessing and comparing new approaches, especially in \acrshort{drl}, where many results come from empirical evaluations. We can identify three major ways of evaluating state representations:
\begin{itemize}
    \item reward over training and testing,
    \item structural metrics, and
    \item qualitative metrics.
\end{itemize}

\subsubsection{Reward Over Training and Testing}

The reward over training and testing is the most used of the metric. The quality of the different learned representations is indirectly compared using the training speed of the learned policy and its robustness when tested. The policy can be learned jointly with the representation or after the representation is entirely learned. The performance of the agent is ultimately the critical metric in \acrshort{drl}. However, it is usually computational expensive to train multiple agents. Due to the intrinsic stochasticity of \acrshort{drl} experiments, runs with multiple random seeds (5 to 20 nowadays) have to be performed (a complete discussion on the topic is provided in \cite{agarwal2021deep}) and the performance is heavily dependent on the choice of the \acrshort{drl} algorithm and its hyperparameters, making it difficult to disentangle the benefits of the learned representation from the overall increment of the cumulative rewards.

\subsubsection{Structural Metrics}

An alternative and appealing way of assessing the learned representations' quality is using the so-called structural metrics. Structural metrics comprise all the metrics evaluating the properties of the learned representations independently of the performance of the \acrshort{drl} agent. An example is the correspondence between the learned states and the actual states of the environment. The state of the environment is still considered non-observable over training, but in most cases, e.g. games and physics simulators, it is available for evaluation. Common structural metrics are:
\begin{itemize}
    \item Ground Truth Correlation (\acrshort{gtc}) \cite{raffin2018s} measuring the correlation between the learned states and the true environment states.
    \item \acrshort{knn} \cite{raffin2018s} measuring the distance between K neighbours of latent states and  K neighbours of true states.
    \item Linear Reconstruction Error \acrshort{lre} \cite{chopra2005learning} measuring the disentanglement of the state features by linearly mapping them to their respective true state value.
    \item Validation Decoder (\acrshort{vd}) \cite{morik2019state} measuring if the latent states can be used to reconstruct the observations without disambiguities. The gradients of the validation decoder are not propagated through the encoder as we aim at validating a fixed state representation.
    \item Ground Truth Error \cite{Henderson2018} measuring the error to the ground truth with Euclidean or geodesic distance.
    \item Normalization Independent Embedding Quality Assessment \acrshort{nieqa} \cite{zhang2012new} measures the quality of the latent manifold in terms of preservation of the true underlying structure.
    \item Feature predictiveness \cite{finn2016deep} measuring how many of the features learned by the encoder are used by the forward model for predicting the future states.
    \item Hits at rank K \cite{kipf2019contrastive} measuring if the predicted latent state belongs to the \acrshort{knn} of the true encoded observation.
\end{itemize}

\subsubsection{Qualitative Metrics}

Eventually, qualitative evaluations are often provided by the authors of the papers, especially when the representation is attributable to a lower-dimensional domain, where the features can be directly related to physical quantities, such as positions and velocities in robot control. Visualization of latent representations is crucial for interpretability. In most cases, the dimension of the latent representations is chosen higher than the dimension of actual state space to improve the training of the encoder and the disentangling of the features. Therefore, usually \acrshort{pca} or t-distributed Stochastic Neighbour Embedding (\acrshort{tsne}) \cite{van2008visualizing} are used to project, linearly and non-linearly, respectively, the latent state variables into visualizable spaces, either 2D or 3D, that can be plotted and interpreted a posteriori. However, state representations of complex problems can not be reduced to 2D or 3D. Therefore, their interpretability remains an open problem in the literature.
 
\subsubsection{From Metrics to Properties}
In Section \ref{subsec:propertieslearningstaterepresentation}, we introduced four main objectives of \acrshort{srl} methods for \acrshort{drl}, namely sample efficiency, robustness, generalization, and interpretability, and six desired properties we would like to encode into the state representation, namely smoothness, low dimensionality, simple dependencies, Markovianity, temporal coherence, and sufficient expressiveness. Herein, we relate these features of \acrshort{srl} to the metrics introduced in the sections above.

In particular, achieving better sample efficiency, robustness, and generalization translates into higher rewards over training and testing. Thus, the reward metric can be used to compare different methods directly. Interpretability, on the other side, is less straightforward. Metrics such as \acrshort{gtc}, \acrshort{knn}, or \acrshort{nieqa} can be used for quantitatively assessing interpretability, while \acrshort{pca}, and \acrshort{tsne} can be used for qualitative assessment.

Regarding the properties of the state representation, one could think of using \acrfull{knn}, \acrshort{pca}, \acrshort{tsne} to assess the smoothness, low dimensionality, and temporal coherence of the state representations. Furthermore,\acrshort{vd} or \acrshort{lre} can be used to assess the complexity of the dependencies. The representation's Markovianity and expressiveness can again be measured in terms of training and testing rewards.

\subsection{Benchmarks}\label{subsec:benchmakrs}

SRL algorithms are usually evaluated using the same environments built for \acrshort{drl}. Their domains span from the Atari games to continuous control tasks such as pendulum and cheetah. Below, we list the most used platforms:
\begin{itemize}
    \item Atari Learning Environment \cite{bellemare2013arcade} providing a \acrshort{drl} interface to the Atari games with the possibility of playing them from pixels. The Atari games offer the possibility of testing discrete state and action space algorithms and present several challenges to the algorithms, such as partial observability, long-term exploration, and reward sparsity.
    \item DeepMind Control Suite \cite{tassa2018deepmind} including a vast range of continuous control tasks such as pendulum, cheetah, and humanoid. Continuous control tasks require the algorithms to deal with continuous state and action spaces, multiple degrees of freedom, and partial observability when played from pixels. The physics is simulated by Mujoco \cite{todorov2012mujoco}. 
    \item DeepMind Lab \cite{beattie2016deepmind} introducing a first-person 3D environment for testing the navigation, exploration, and generalization skills of the agents.
    \item SRL-Toolbox \cite{raffin2018s} offering a complete package with many of the early \acrshort{srl} approaches from the literature, with evaluation functions, and with a set of testing environments.
    \item OpenAI-Gym \cite{brockman2016openai} collecting a wide variety of \acrshort{drl} environments, spanning from the Atari games to continuous control and robotics problems.
\end{itemize}

%% file: StateRepresentationLearning/8-Discussion.tex
Eventually, we discuss two final aspects: open challenges and future directions.

\subsection{Open Challenges and Future Directions}

\acrshort{drl} and \acrshort{srl} are two quickly-evolving fields, and here we summarize the most exciting future directions.
From our review of the literature, we have identified several aspects that are currently understudied in the field:
\begin{itemize}
    \item the incorporation of inductive biases and prior knowledge in the neural network architectures using the tools provided by Geometric Deep Learning \cite{bronstein2021geometric},
    \item the proper quantification of uncertainties in the latent variables, observations, and \acrshort{mdp} dynamics,
    \item the learning of general representations using the Meta-Learning paradigm \cite{vilalta2002perspective}, 
    \item the evaluation and comparison of the approaches,
    \item and the interpretability and explainability of the latent representations. 
\end{itemize}

\subsubsection{Incorporation of Inductive Biases}

In the last decade, \acrshort{dl} has solved many high-dimensional problems considered beyond reach by developing increasingly complex architectures with a continuously-growing number of parameters and requiring more data.

When scaling to larger problems, simply increasing the model complexity and the size of the training set do not seem viable options due to the need for specialized hardware architectures with high power consumption and, consequently, emissions. However, many complex problems present low-dimensional structures, symmetries, and geometry deriving from the physical world. Incorporating this knowledge into the neural network architectures and the learned representations is essential for tackling the curse of dimensionality, reducing the number of parameters needed by each model, and improving generalization to unseen data \cite{bronstein2021geometric}. For example, when learning a latent forward model for a mobile robot, we would like the model to learn \textit{invariant} features to background changes or to noise in the measurements.

The \acrshort{srl} methods described in this review have prepared the ground for incorporating geometry, group theory and symmetries in \acrshort{drl} methods. Exploiting symmetries, structures, and prior knowledge are three critical ingredients for sample efficient \acrshort{srl} and \acrshort{drl} that are yet under-investigated but have shown outstanding potential for future research \cite{van2020mdp, mondal2020group}.

\subsubsection{Uncertainty Quantification in Unsupervised Representation Learning}

Uncertainties may arise from different sources, e.g. the measurement noise or the chaotic dynamics of the system at hand. Quantifying and disentangling these data uncertainties is crucial to the generalization of state representations and latent models. While this problem is primarily studied in supervised learning \cite{abdar2021review}, it is even more complex in unsupervised representation learning in dynamical systems. The goal of understanding and discovering the low-dimensional set of variables describing the evolution of the systems cannot be achieved without proper uncertainty quantification. 

In particular, we highlight Deep Kernel Learning \cite{wilson2016deep, wilson2016stochastic, botteghi2022deep} as a computationally-efficient approach, scalable to high-dimensional data, for combining the expressive power of neural networks with the uncertainty quantification abilities of Gaussian processes and kernel methods \cite{rasmussen2003gaussian}.

\subsubsection{Meta Learning General Representations}

\acrshort{drl} agents have faced an uncountable amount of different tasks, and this trend is only destined to grow in the future. Learning to adjust the behavior to new problems from limited data is still an open question in \acrshort{drl} research. While Meta Learning has been studied in \acrshort{drl} for learning general-purpose policies \cite{finn2017model, pang2018meta, li2018learning}, this paradigm has yet to be exploited for learning general-purpose state and action representations and \acrshort{mdp} models. In contrast, with the expensive Meta \acrshort{drl}, which directly uses high-dimensional observations, Meta Learning general representations may provide new ways to tackle generalization and transfer learning of the learned policy to new tasks and to new environments.

\subsubsection{Evaluation and Interpretability of the Latent Representations}

In Section \ref{sec:evaluationandbenchmark}, we discussed different metrics for assessing the quality of the representations, e.g. through the reward over training or through the structural and qualitative metrics.

However, here we argue that none of these metrics provides the ultimate answer to which method is the best. While the reward over training is the most popular metric, it is subjected to the intrinsic stochasticity of the \acrshort{drl} process, and the results need to be evaluated over multiple runs. The need for multiple runs makes the process extremely computationally and resource-intensive for complex environments and high-dimensional observations. On the other side, structural metrics provide a good way of comparing approaches without high-computational demands. However, high metric scores do not necessarily translate into optimal policy results and the highest reward over training. The same holds for qualitative metrics. High human-interpretability of the representations does not necessarily mean optimal behaviors of the agents. 

Therefore, we believe that further research must be done to understand what the optimal state representation is, how to learn it, and how to interpret and explain representations and agents' behaviors.

%% file: StateRepresentationLearning/9-Conclusion.tex
This paper reviewed the most important and newest research trends on unsupervised \acrshort{srl} in \acrshort{drl}. From the literature, we identified four major classes of \acrshort{srl} methods: observation encoding, learning of latent \acrshort{mdp} models, using priors, and contrastive learning. Moreover, we study how these methods are combined to develop more complex approaches. Eventually, we extended the review to include advanced methods and applications of \acrshort{srl}.

Unsupervised representation learning has proven to be a crucial element of the best-performing \acrshort{drl} algorithms for improving sample efficiency, robustness, generalization, and interpretability. Despite the progress in the field, many issues remain open for tackling complex real-world problems. 

For example, the incorporation of prior knowledge and geometry, the exploitation and learning of symmetries, the quantification of the uncertainties, the generalization to different tasks, and the evaluation and interpretation of the learned representations are important problems to address in the future.

%% file: StateRepresentationLearning/10-Appendix.tex
\subsection{Pendulum Model}

For our experiments, we consider the pendulum described by the following equation:
\begin{equation}
    \ddot{\psi}(t) = - \frac{1}{ml}(mg \sin{\psi}(t)+a(t))\,,
    \label{eq:pendulum_dynaimcs}
\end{equation}
where $\psi$ is the angle of the pendulum, $\ddot{\psi}$ is the angular acceleration, $m$ is the mass, $l$ is the length, and $g$ denotes the gravity acceleration. 

We assume no access to $\psi$ or its derivatives, and the measurements are RGB images of size $84 \times 84 \times 3$ obtained through an RGB camera and we utilize frame stacking to allow the network to infer velocity information, making the observations of size $84 \times 84 \times 6$. The observation are collected by following a random policy for 200 steps with different initial conditions. The training set is composed of 10000 data tuples $(\mathbf{o}_t, a_t, \mathbf{o}_{t+1}, r_t)$, while the test set is composed of 2000 data tuples. Different random seeds are used for collecting training and test sets. 

The pendulum environment used for collecting the data tuples is the {\fontfamily{cmtt}\selectfont Pendulum-v1} from Open-Gym \cite{brockman2016openai}.

\subsection{Encoder Architecture}
The input observations are two $84\times84\times 3$ RGB images stacked together. The (deterministic) encoder $\phi_{\text{enc}}$ is composed of 4 convolutional layers with 32 filters per layer. The convolutional filters are of size $(3\times3)$ and shifted across the images with stride 1 (only the first convolutional layer has stride 2 to quickly reduce the input dimensionality). Batch normalization is also used after the 2nd and 4th convolutional layers. A similar convolutional architecture is used in PlaNet\cite{hafner2019learning} and Dreamer\cite{hafner2019dream}. The output features of the last convolutional layer are flattened and fed to two final fully connected layers of dimensions 256 and 50, respectively, compressing the features to a 50-dimensional latent state vector. Each layer has ELU activations, except the last fully-connected layer with a linear activation. 

Analogously, the VAE (stochastic) encoder shares the same architecture with the deterministic encoders but instead of learning the latent state vectors $\mathbf{z}_t$, the encoder learns the mean and the covariance of a Gaussian distribution representing the distribution $p(\bm{z}_t|\bm{o}_t) = \mathcal{N}(\bm{\mu}_{t}, \boldsymbol{
\Sigma}_{t})$. Samples $\bm{z}_t$ from $ p(\bm{z}_t|\bm{o}_t)$ are obtained by reparametrization trick \cite{kingma2014stochastic}.

\subsection{Decoder Architecture}

 The decoder $\phi_{\text{dec}}$ is parameterized by an NN composed of a linear fully-connected layer and 4 transpose convolutional layers with 32 filters each. The convolutional filters are of size $(3 \times 3)$ and shifted across the images with stride 1 (again, the last convolutional layer has stride 2). Batch normalization is used after the 2nd and 4th convolutional layers, and ELU activations are employed for all the layers except the last one. The output is the observation reconstruction $\hat{\bm{o}}_t$. 
 
\subsection{Latent Models Architecture}
The latent models implemented are:
\begin{itemize}
    \item forward model with input $\bm{z}_t$ and $\bm{a}_t$ and output $\bm{z}_{t+1}$,
    \item reward with input  $\bm{z}_t$ and $\bm{a}_t$ or $\bm{z}_t$ and output the predicted reward $\hat{r}_t$,
    \item inverse with input  $\bm{z}_t$ and  $\bm{z}_{t+1}$ and output the predicted action $\hat{\bm{a}}_t$, and 
    \item action encoder with input $\bm{z}_t$ and $\bm{a}_t$ and output $\bar{\bm{a}}_{t}$
\end{itemize}
All these models are composed of an input layer of size dependent on the input data, an hidden layer of size 256, and an output layer again model-dependent. 

\subsection{State Decoder Architecture}
The linear and nonlinear state decoders are composed of three \acrshort{fcnn} layers with input dimension equal to the latent state dimension, output dimension equal to the the true state dimension, and hidden layer of dimension 256. For the nonlinear state decoder we add two eLU activations after the first and second layer.

\subsection{Hyperparameters of the Experiments}
The hyperparameters used in our experiments are reported in Table \ref{tab1app}, where for fairness of comparison they are shared among the different methods. 
\begin{table}[h!]
\caption{Hyperparameters used in the experiments.}
\centering
\begin{tabular}{|c | c|} 
\hline
 \hline
 Hyperparameter & Value \\ [0.5ex] 
 \hline\hline
learning rate of NN  & $3\mathrm{e}-4$  \\ 
 \hline
$L^2$ regularization coefficient & $1\mathrm{e}{-3}$ \\
\hline
 $\beta$ (VAE) & $0.1$  \\
  \hline
   $\gamma$ (BM) & $0.99$  \\
   \hline
  number of epochs & $50$ \\
  \hline
    true state dimension & $3$ \\
  \hline
  latent state dimension & $50$ \\
 \hline
  latent action dimension & $50$ \\
 \hline
  image dimension & $84\times84\times3$ \\
  \hline
  observation dimension & $84\times84\times6$ \\
  \hline
  action dimension & 1 \\
 \hline
 mass of the pendulum & 1 \\
 \hline
 length of the pendulum & 1 \\
 \hline
 \hline
\end{tabular}
 \label{tab1app}
\end{table}